\documentclass[12pt,twoside,a4paper,openright]{report}
\usepackage{xcolor}
\usepackage{qutthesis}
\usepackage{amsmath}
\usepackage{amssymb}
\usepackage{verbatim}
\usepackage{longtable}
\usepackage{hyperref}

\usepackage[sort&compress,square]{natbib}
\bibpunct{[}{]}{,}{a}{,}{,}

\usepackage{pdfpages}
\usepackage{graphicx}
\usepackage{siunitx}
\usepackage{enumitem}
\usepackage{float}
\usepackage{xcolor}
\usepackage[caption=false, font=footnotesize]{subfig} 
\usepackage{titlesec}
\usepackage{gensymb}
\usepackage{tabularx,booktabs}
\usepackage[font=footnotesize]{caption}
\usepackage{adjustbox}
\newcommand{\w}{6}
\newcommand{\h}{3.65}

\usepackage{todonotes}

\newcolumntype{Y}{>{\centering\arraybackslash}X}
\newcolumntype{x}[1]{>{\centering\let\newline\\\arraybackslash\hspace{0pt}}m{#1}}
\usepackage{array}
\newcolumntype{P}[1]{>{\centering\arraybackslash}p{#1}}
\newcolumntype{M}[1]{>{\centering\arraybackslash}m{#1}}
\usepackage{graphbox}
\tablespagetrue
\begin{document}


\title{Learning Visuo-Motor Behaviours for 
\\ Robot Locomotion Over Difficult Terrain}

\author{Brendan Tidd \\ BEng(Mech)(Hons)}

\authoremail{brendan.tidd@hdr.qut.edu.au}


\thesistype{Doctor of Philosophy}    
\university{The Queensland University of Technology} 
\faculty{Faculty of Engineering}   
\school{School of Electrical Engineering and Robotics}   
\universitylogo{yes}{1.0}{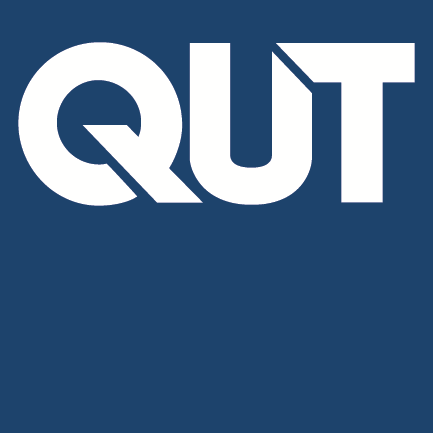}  
\submissiondate{2022}
\copyrightyear{2022}
\informationcutoffdate{March 2022} 

\maketitle	   


\setcounter{page}{1} 



\begin{dedication}
For Bree, Elliott and Darcy.
\end{dedication}

\begin{abstract}

As mobile robots become useful performing everyday tasks in complex real-world environments, they must be able to traverse a range of difficult terrain types such as stairs, stepping stones, gaps, jumps and narrow passages. This work investigated traversing these types of environments with a bipedal robot (simulation experiments), and a tracked robot (real world). Developing a traditional monolithic controller for traversing all terrain types is challenging, and for large physical robots realistic test facilities are required and safety must be ensured. An alternative is a suite of simple behaviour controllers that can be composed to achieve complex tasks.

The contribution of this thesis is behaviour-based locomotion over challenging terrain, which requires designing controllers for traversing complex terrain, and understanding how to transition between behaviours. Visuo-motor behaviours were developed for specific obstacles, representing separate subsets of terrain the robot may encounter. In the first contribution, the paradigm of curriculum learning (CL) was explored and an efficient three-stage process was developed for training behaviours, resulting in robust bipedal locomotion in simulation experiments over curved paths, stairs, stepping stones, gaps and hurdles. 

For the second contribution, behaviours developed in simulation were applied to a real-world scenario where a tracked robot was required to pass through a narrow doorway. A gap behaviour trained in simulation was transferred to a real robot, achieving a 73\% success rate traversing small gaps where traditional path planning methods failed. The importance of state overlap was demonstrated in simulation experiments with the third contribution by learning when to switch controllers, enabling the biped to successfully traverse a randomised sequence of obstacles and achieve a success rate of 71.4\%, compared to 0.7\% for policies trained without an explicit state overlap. Policies were developed in the fourth and final contribution that guide the biped towards a switch state for the next behaviour, improving the reliability of switching. These setup policies achieved an 82\% success rate on a difficult jump task, compared to 51.3\% with the best method for training transition policies.

This work efficiently trained complex behaviours to enable mobile robots to traverse difficult terrain. By minimising retraining as new behaviours became available, robots were able to traverse increasingly complex terrain sets, leading toward the development of scalable behaviour libraries.







\end{abstract}


\begin{ack}

To my supervisor Dr Juxi Leitner, you started me on this journey, thank you for the insightful discussions, guidance, and opportunities you have provided for me over the years. Thank you to Professor Peter Corke for having the vision that has allowed me and many others to pursue research in robotics, technology that will shape the world of the future. Thank you to Dr Nicolas Hudson for opening big doors for me at CSIRO and through the DARPA Subterranean Challenge. It was your faith that led me to command a fleet of autonomous robots on the world's stage, exhilarating experiences that I will remember forever. Thank you to Dr Akansel Cosgun for your dedication to writing a good paper, you have taught me many lessons for which I am grateful. Thank you to Dr Niko Suenderhauf and Dr Jason Williams for being on my final seminar panel and providing useful insights. 

I am thankful for all of the team at CSIRO, they have been a great source of inspiration. In particular I would like to thank Tom Hines, Tom Molnar, Alex Pitt, and Fletcher Talbot for their assistance in getting my code to work on real robots, and Ryan Steindl for the inevitable repair. I would also like to thank all of the people I have interacted with over the years at the robotics lab at QUT, and in particular the reinforcement learning group, Jake Bruce, Riordan Callil, Vibhavari Dasagi, Jordan Erskine, William Hooper, Robert Lee, Krishan Rana, and Fangyi Zhang, for always providing insightful ideas and feedback. 

Mostly, I would like to thank my wife Bree, and acknowledge her endless patience and support. Her efforts throughout this process deserve many more accolades than my own. To my family and friends that I have long neglected, thank you for forever being supportive. Finally, my pursuit of contributing to technology is driven in part to provide inspiration for my sons, Elliott and Darcy. I hope that one day you find your way to where no one has ever been.
\newpage
This research was supported by an Australian Government Research Training Program Scholarship, and an Australian Research Council Centre of Excellence for Robotic Vision top-up scholarship.






\end{ack}



\listnomenclatureatfront{yes}{./nomenclature.tex} 



\afterpreface

\chapter[Introduction]{Introduction}
\label{cha:Introduction}





Robots are making their way into an increasing number of roles in our society. As the expectations of autonomous platforms increase, locomotion over complex terrain becomes a fundamental limitation. For example, a delivery robot that cannot cross gaps in the pavement or traverse stairs is severely limited to where it can deliver, yet this is trivial for a human. Designing controllers for these scenarios is challenging in part due to the complexity of perception and action coupling (known as visuo-motor coupling) where extracting salient features from information-rich sensors like vision and costmaps is difficult. For example, bipedal locomotion over complex terrain requires footstep and contact planning as the robot interacts with the terrain. Practical limitations, such as the need for testing facilities, safety harnesses and human supervisors for controller development and tuning, restricts controller design to a small subset of what may be encountered in the real world. Furthermore, new terrain conditions are likely to be identified during operation, where each inclusion poses difficult and diverse visuo-motor challenges. For these reasons, the development of a single monolithic controller to perform all expected maneuvers is challenging. Controller design must allow for the development of specialist behaviours, and integration with existing behaviours by understanding when each can be activated, where a behaviour is defined as a mode of control for performing a set motion.

In recent years, Deep Reinforcement Learning (DRL) methods have demonstrated impressive results for a diverse range of robotics and artificial intelligence applications. For robot locomotion, DRL enables the development of behaviours that are otherwise extremely difficult to engineer, particularly where perception must be deeply integrated with control. However, DRL policies are typically inefficient to train. Training efficiency refers to the duration of interaction with the environment needed to produce a suitable behaviour. DRL polices typically require a large number of training steps, often converging to a suboptimal behaviour with standard state exploration strategies and reward design (as discussed in Chapter~\ref{cha:ch3a}). Furthermore, to accommodate combinations of terrain types, learning methods usually require access to all variations in terrain when training. These challenges can be overcome by training behaviours separately for specific conditions, but introduces the complexity of integrating policies with other controllers. Switching controllers when the robot is not in a compatible state (robot position and configuration relative to the terrain) can result in undesired behaviour or instability. The set of states, from which a controller will converge to its designated behaviour over time is known as the region of attraction (RoA) of the controller. Behaviour switching must occur when the robot is in the RoA of the subsequent controller, however, it may be difficult to determine the RoA for a complex behaviour based on learned policies. These limitations impede the application of DRL policies in real-world problems. This thesis investigates how to efficiently develop complex behaviours, and how to integrate separate controllers to create robust solutions for broader conditions. 





Real-world scenarios require robots to traverse diverse terrain conditions and this was highlighted in the recent Defense Advanced Research Projects Agency (DARPA) Subterranean Challenge (SubT) where robots autonomously explored mine tunnels, urban underground and cave networks providing situational awareness for first responders [\cite{darpa_darpa_2021}]. The robots faced many complex terrain artifacts such as stairs, platforms, train tracks, and narrow passages, which posed significant challenges for mobile robots of various morphology and complexity. For example, a large tracked robot was required to traverse narrow doorways, traditional path planning algorithms failed due to kinematic constraints of the robot and low resolution costmaps [\cite{hudson_heterogeneous_2021}]. For urban and mine environments, bipedal robots offer the possibility of traversing a similar range of terrains as does a human. However, bipedal robots are currently highly complex and expensive, so switching between behaviours must occur with care to prevent damage to the robot or the environment. This thesis describes experiments with a dynamic biped in simulation, and a large tracked platform in a real-world scenario. 

The focus of this thesis is on learning behaviours for autonomous locomotion over complex terrain. Each behaviour investigated in this work requires perception and action coupling, and integration into a larger control suite. The aim of this work is to have an extensible behaviour library, where new controllers can be developed efficiently, and incorporated without retraining existing behaviours. This will be key to developing autonomous mobile platforms that operate at scale.

\section{Research Questions}

In this section, the research questions investigated throughout this thesis are introduced. 

\begin{itemize}[label={}] 

    \item \textbf{Research Question 1: \textit{How can complex visuo-motor locomotion behaviours be learned efficiently?}}

    Developing behaviours for robot locomotion over difficult terrain is challenging and complexity arises from the robot (many degrees of freedom (DoF), dynamics, balancing, underactuation), terrain difficulty, or from information-dense sensors such as vision. DRL methods have been used for developing complex visuo-motor policies, however, training can be inefficient using exploration strategies that rely on random control signals for policy improvement. These na\"ive exploration methods require extensive interaction between the robot and the training environment, and local minima may prevent the robot from discovering the necessary behaviour. Efficiency can be improved by affording additional information to the agent, for example, by providing expert demonstrations [\cite{peng_deepmimic_2018}], however, skilled demonstrations are difficult to acquire. This thesis investigates how complex behaviours can be learned efficiently using simple controllers for guidance, reducing the number of samples needed for training while achieving high success rates traversing difficult terrain types. 

    



    \item \textbf{Research Question 2: \textit{How can a safe switch state be determined to facilitate the reliable switching of behaviours?}}

    In real-world scenarios, robots are required to traverse many obstacles, each requiring a separate behaviour. For safe and reliable behaviour switching to occur, a behaviour must be activated when the robot is within that behaviour's RoA. Determining when a robot is in a safe state for switching can be challenging due to the complexity of the robot, terrain, controller, and their various interaction. This thesis investigates reliable behaviour switching considering these challenges.


  \item \textbf{Research Question 3: \textit{How can an agent prepare for an upcoming behaviour such that safe switching can occur?}}


    While research question 2 investigates identifying states where behaviours can be safely switched, a robot however, may not always be moving toward such a state. The final investigation of this thesis is how can a robot learn to move towards the RoA of an upcoming behaviour such that safe switching can occur. While there has been work in this area for learning transition policies for simpler locomotion agents [\cite{lee_composing_2019}], solving this problem for robots with many DoF performing complex maneuvers remains challenging. 
    

\end{itemize}
\newpage

\section{Summary of Contributions}

A summary of primary contributions presented in this work are as follows:
\begin{itemize}
    \item A novel curriculum learning approach reduced the training time for learning visuo-motor behaviours that make progress over complex terrain. Policies were trained efficiently for a dynamic biped in simulation using a set of joint and body positions from a simple walking trajectory, guiding learning for behaviours that traverse a diverse set of difficult terrain \textbf{[RQ1]}.
    
    \item A visuo-motor policy was trained efficiently for traversing narrow gaps with a tracked robot using a simple waypoint controller to guide the robot towards the gap \textbf{[RQ1]}. A selection policy was trained for autonomously switching between the narrow gap controller and a set of traditional controllers \textbf{[RQ2]}. The effectiveness of these behaviours developed in simulation were demonstrated in a real-world scenario. 
    
    
    
    \item Switch estimator networks were trained for each behaviour that predict the outcome of switching from a given state. These networks improved the success rate for traversing a sequence of terrain types with a simulated biped compared to several alternative methods for predicting when to switch \textbf{[RQ2]}.
    

    \item Setup policies were trained to transition between behaviours, moving a robot into the RoA of a target behaviour. A significant improvement in success rate was demonstrated using setup policies with a simulated biped preparing for a difficult jump behaviour \textbf{[RQ3]}. These policies were trained using a novel reward signal designed to guide the robot towards a behaviour required to traverse the terrain.
    

\end{itemize}

\newpage

\section{Document Outline}

The outline of this thesis is as follows. Chapter \ref{cha:LiteratureReview} provides a review of literature for behaviour based robotics locomotion and control. Chapters \ref{cha:ch3a}-\ref{cha:ch5} presents the research papers created during candidature. 

Chapter \ref{cha:ch3a} presents the paper published at ACRA 2020 [\cite{tidd_guided_2020}] that uses a three-stage curriculum learning method for learning complex behaviours, demonstrated with a dynamic biped walking over several complex terrain obstacles (curved paths, stairs, gaps, hurdles, and stepping stones). This work was conducted in simulation with torque controlled actuation, and perception supplied by a depth camera mounted to the robot. 

Chapter \ref{cha:ch3b} presents the paper published at IROS 2021 [\cite{tidd_passing_2021}], where a tracked robot learned to traverse narrow doorways using an occupancy map as input. A goal-dependent behaviour selection policy learned when to activate the gap behaviour or when to utilise a set of traditional controllers. The effectiveness of these policies was demonstrated in a real scenario with a large tracked robot passing through small gaps.

Chapter \ref{cha:ch4} introduces switch estimator policies from the paper published at IROS 2021 [\cite{tidd_learning_2021}]. Switch estimators were trained to predict when a simulated biped should switch from a stair, gap, or hurdle behaviour for traversing a sequence of terrain types. 




Chapter \ref{cha:ch5} presents the paper published in the IEEE Robotics and Automation Letters [\cite{tidd_learning_2021-1}]. Setup policies were developed for preparing the robot for the subsequent behaviour, greatly improving switching between behaviours without a reliable RoA overlap. 

Chapter \ref{cha:conclusion} provides a summary, discussion of our findings, and several directions for improvements and future research. 


\chapter[Literature Review]{Literature Review}
\label{cha:LiteratureReview}

This chapter presents a survey of literature related to robot locomotion over challenging obstacles, summarising the existing solutions for these topics and identifying areas of interest for further investigation. For decades classical methods enabled mobile robots to move over difficult terrains, however, as robot and task complexity increases, the requirement of manual tuning becomes a limitation of these approaches. In recent years, learning methods have been used for developing visuo-motor policies for complex locomotion behaviours.

Existing methods for bipedal locomotion are summarised, including a review of curriculum learning (CL) for legged platforms. Navigation with mobile platforms has been researched extensively, works related to passing through small gaps are examined as a focus of this thesis. Literature is presented on behaviour composition, and the current methods for combining separate controllers, considering traditional and learning approaches. Finally, relevant literature on learning transition policies for connecting separate behaviours is introduced.

\section{Walking Over Difficult Obstacles}

Walking is a challenging control problem. Once stable walking is achieved, the average infant falls 17 times per hour [\cite{adolph_how_2012}]. Designing controllers for dynamic bipedal walking robots is difficult, particularly when operating over complex terrain [\cite{atkeson_what_2016}]. In this section, literature is presented from classical and deep reinforcement learning approaches for bipedal walking.

\begin{figure}[h!tb]
\centering
\subfloat[A cart-table model]{\includegraphics[width=0.4\columnwidth]{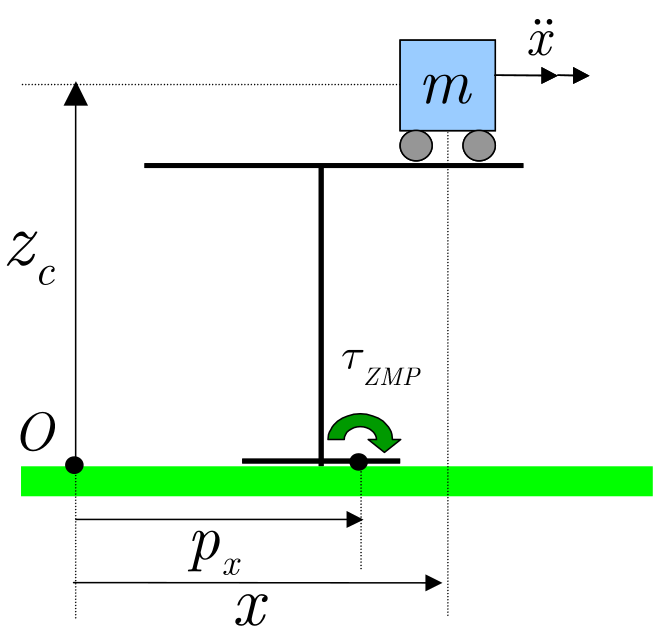}}
\subfloat[A pendulum under constraint]{\includegraphics[width=0.4\columnwidth]{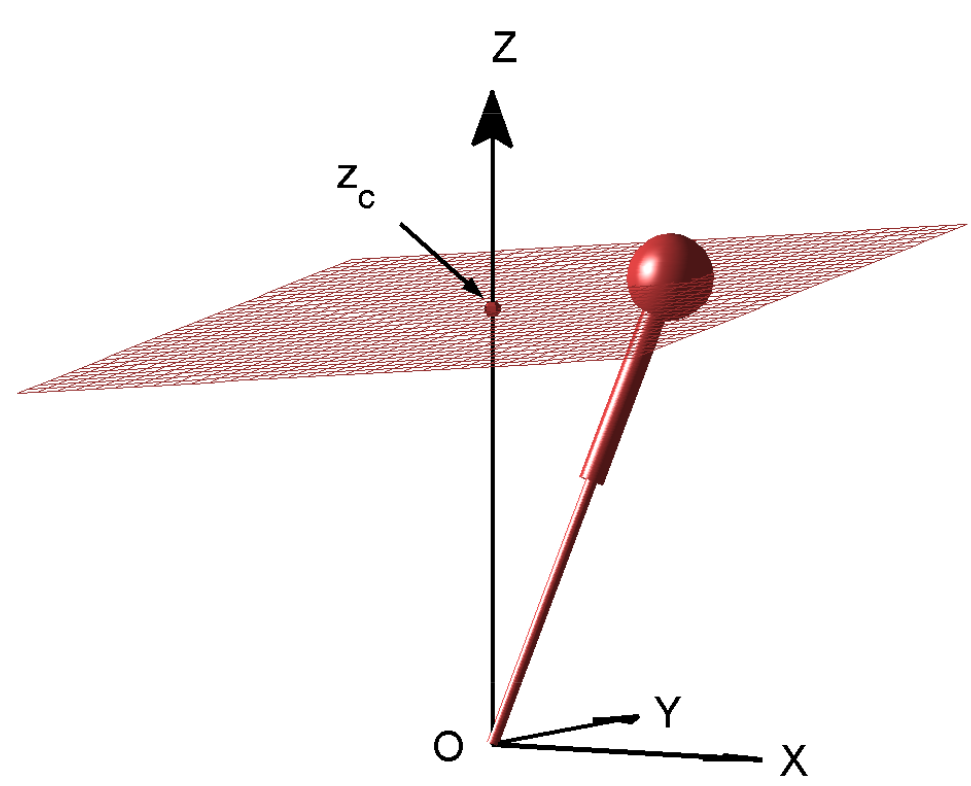}}

\caption{Zero moment point (ZMP) model using an inverted pendulum under constraint[\cite{kajita_biped_2003}].}
\label{fig:zmp}
\end{figure}
\begin{figure}[h!tb]
\centering
\subfloat[The capture point (CP) is located within the capture region, and defines a point the robot can step to come to a complete stop.]
{\includegraphics[width=0.5\columnwidth]{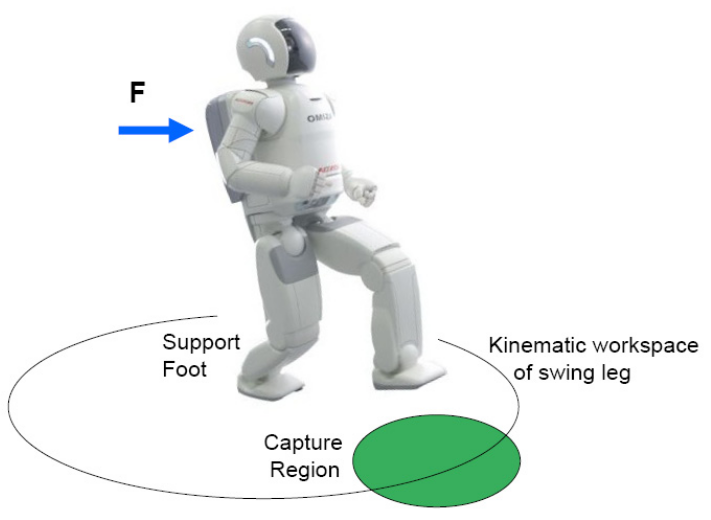}}
\hspace{0.5cm}
\subfloat[The flywheel inverted pendulum considers the inertia of the center of mass (CoM).]
{\includegraphics[width=0.4\columnwidth]{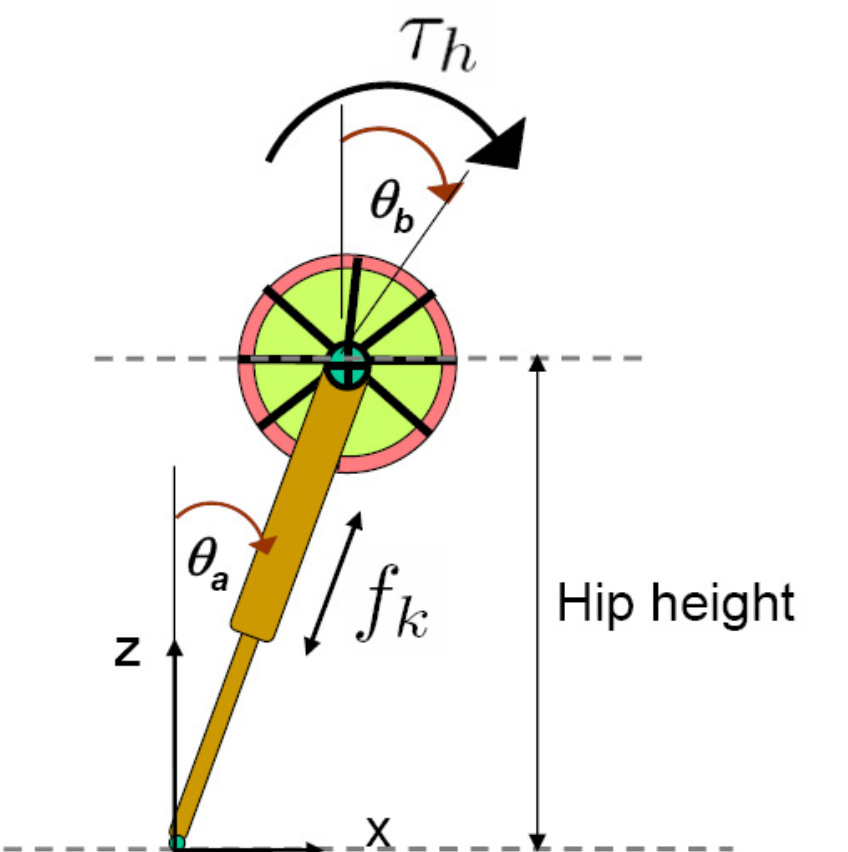}}
\caption{Capture point (CP) controller using a flywheel inverted pendulum model[\cite{pratt_capture_2006}].}
\label{fig:capture_point}
\end{figure}

Classical control methods have been used for biped walking for decades. Commonly, ideas such as Zero Moment Point (ZMP) [\cite{kajita_biped_2003}] and Capture Point (CP) [\cite{pratt_capture_2006}] utilise simplified dynamical models such as the spring loaded inverted pendulum (SLIP) to generate stable trajectories [\cite{geyer_compliant_2006}]. ZMP refers to the point on the ground where the horizontal moments sum to zero. This location is the same as the centre of pressure if both feet are on the same plane (centre of pressure is linked to contact forces, ZMP to inertial and gravity forces). While this point is contained within the support polygon created by the stance feet, the robot will remain stable. Figure \ref{fig:zmp} shows the ZMP idea with a simple cart-table model, along with the inverted pendulum model used by \cite{kajita_biped_2003} to plan walking trajectories for a biped. Capture point (CP, also referred to as Divergent Component of Motion (DCM)) refers to a point that can bring the robot to a complete stop if that point is located within the support polygon created by the stance feet. Considering the inverted pendulum with flywheel model (Figure \ref{fig:capture_point}.b), the CP is not a unique point. A capture region defines a collection of CP's that exist within reach of the swing foot (Figure \ref{fig:capture_point}.a). Using the CP idea, it is not necessary to have perfect foot placement, but good foot placement within a capture region is important [\cite{pratt_capture_2006}].

Classical methods for controller design have allowed bipedal robots ascend and descend stairs [\cite{ching-long_shih_ascending_1999}], balance on a Segway [\cite{gong_feedback_2019}], and perform jumping behaviours [\cite{xiong_bipedal_2018}]. Humanoid robots often employ a set of control primitives, each individually developed and tuned. Dynamic Movement Primitives (DMP) consisting of discrete and rhythmic controllers have enabled a humanoid robot to play the drums and swing a tennis racket [\cite{schaal_dynamic_2006}]. \cite{siciliano_using_2008} used primitives to place and remove a foot from contact, enabling a robot to walk on uneven terrain, up a step, and climb a ladder. Primitives developed from motion capture can be stitched together with Hidden Markov Models (HMMs) to create sequences of motions [\cite{kulic_incremental_2012}]. While classical methods can achieve complex behaviours, they typically require extensive expert tuning. Furthermore, behaviours that rely on perception, for example walking up and down stairs, require footstep locations to be resolved explicitly from high dimensional sensors. For these reasons, classical methods are limited in their extension to more challenging obstacles.

\begin{figure*}[htb!]
\centering
\includegraphics[width=0.8\columnwidth]{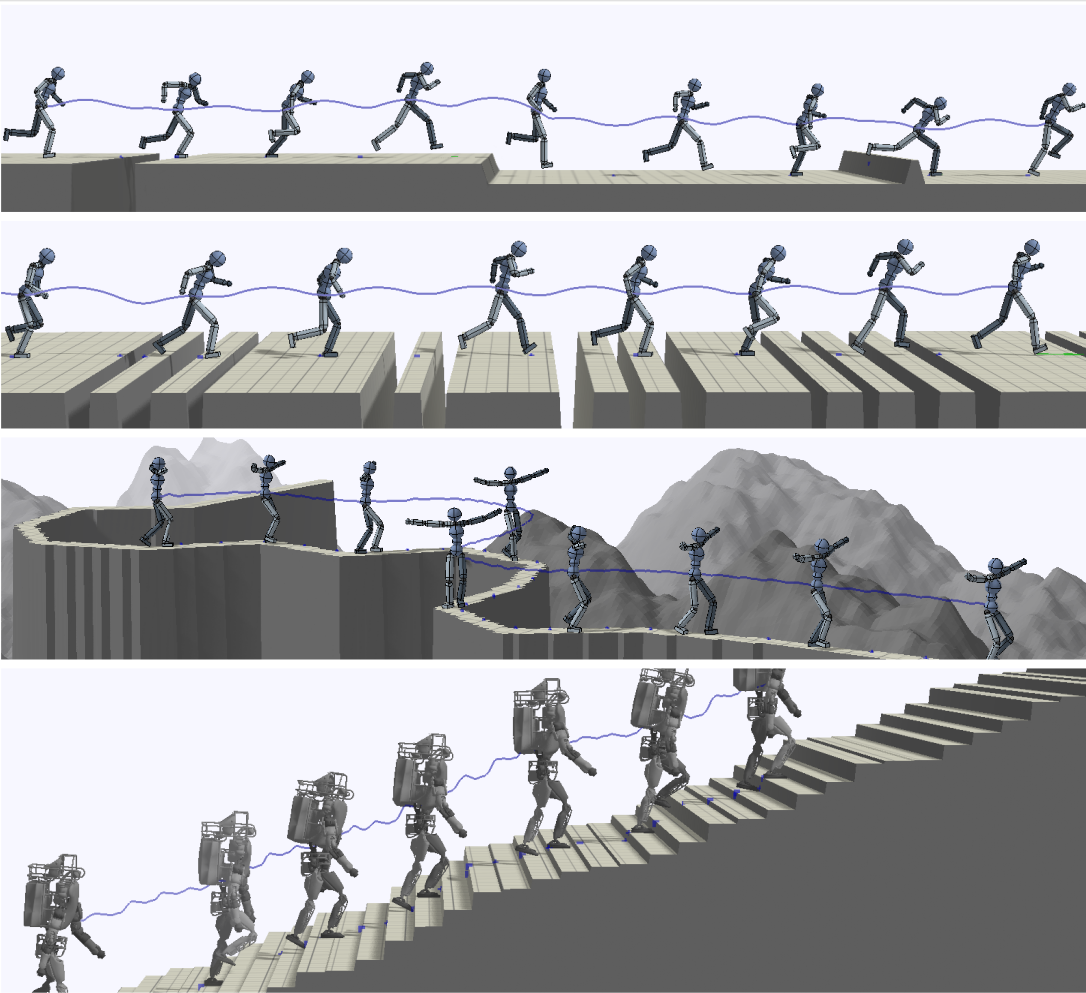}
\caption{Characters traversing randomly generated terrains [\cite{peng_deepmimic_2018}].}
\label{fig:ch2_deep_mimic1}
\vspace{-3mm}
\end{figure*}

Learning methods have recently demonstrated behaviours for bipedal walking over obstacles in simulation. \cite{peng_deepmimic_2018} learn complex maneuvers for high degree of freedom (DoF) characters by following motion capture data. In this work, the agent is first trained on flat terrain, then policies are augmented with a heightmap, enabling the robot to walk up stairs. Example behaviours are shown in Figure \ref{fig:ch2_deep_mimic1}. This paper also introduces methods for learning which behaviour to select from a one-hot encoded behaviour selector policy and by constructing composite behaviours that combine learned skills. \cite{song_recurrent_2018} develop a single recurrent neural network policy to demonstrate locomotion with a 2D biped walking over several obstacles including gaps, stairs, and hurdles using low dimensional observations of the terrain.


Learning approaches have also been used in real-world scenarios with bipeds walking over difficult terrain. \cite{da_supervised_2017} use supervised learning to train a control policy for a biped from several manually derived controllers that perform periodic and transitional gaits on flat and sloped ground. This method uses controllers developed with a full dynamic model, virtual constraints, and parameter optimization that are instilled into a neural network policy. The output of the policy is a set of B\'ezier coefficients to control the walking motion of a real biped on an undulating grass field. \cite{siekmann_sim--real_2021} use a periodic reward to train an adjustable walking policy for a real-world Cassie robot, performing complex behaviours such as jumping, skipping, and stepping over difficult obstacles such as stairs.

\subsection{Deep Reinforcement Learning}

\begin{figure*}[htb!]
\centering
\includegraphics[width=0.8\columnwidth]{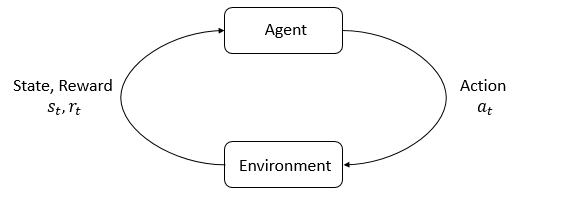}
\caption{Agent-environment interaction loop.}
\label{fig:ch2_rl_diagram}
\vspace{-3mm}
\end{figure*}

This section introduces deep reinforcement learning (DRL), with a focus on continuous control. DRL is an alternative to classical control methods for developing complex behaviours. In place of extensive manual design, robots instead learn behaviours by interacting with the environment. Figure \ref{fig:ch2_rl_diagram} [\cite{SpinningUp2018}] shows the agent-environment interaction loop, where an agent performs an action, receives a reward, and observes the next state.

An agent acting in an environment is considered a Markov Decision Process $\text{MDP}$, defined by tuple $\{\mathcal{S},\mathcal{A}, \mathcal{R}, \mathcal{P}, \gamma\}$ where $s_t \in \mathcal{S}$, $a_t \in \mathcal{A}$, $r_t \in \mathcal{R}$ are state, action and reward observed at time $t$, $\mathcal{P}$ is an unknown transition probability from $s_t$ to $s_{t+1}$ taking action $a_t$, and $\gamma \in [0,1)$ is a discount factor. A system that obeys the Markov property requires only the most recent state and action, and no prior history, to make decisions about future actions. For continuous control policies, actions are sampled from a deep neural network policy $a_t\sim\pi_\theta(s_t)$, where $a_t$ is is a continuous control signal, for example, the torque applied to each joint at time $t$. The policy is parameterised by weights $\theta$. Neural networks used in DRL are typically small in comparison to those used for supervised learning tasks, usually with only two or three fully connected layers [\cite{schulman_proximal_2017, heess_emergence_2017}], and two or three preceding convolutional layers for tasks that utilise images [\cite{mnih_human-level_2015, peng_deepmimic_2018}]. Smaller networks allow for faster inference, reducing the time to train DRL policies that require millions of interaction steps with the environment.


The goal of reinforcement learning is to maximise the sum of future rewards following trajectory $\tau$:

\begin{equation}
R(\tau) = \sum_{t=0}^{\infty}\gamma^tr_t
\end{equation}

where $r_t$ is a scalar reward provided by the environment at time $t$. 

\noindent\textbf{Proximal Policy Optimisation (PPO):}
PPO was first introduced by \cite{schulman_proximal_2017} and has become a widely used DRL algorithm for continuous control tasks. Notable applications have demonstrated complex control in simulation environments[\cite{heess_emergence_2017, peng_deepmimic_2018, yu_learning_2018, lee_composing_2019, xie_allsteps_2020}], and on robots in the real world [\cite{tan_sim--real_2018, xie_learning_2020, rudin_learning_2021}]. PPO is an on-policy algorithm, meaning it collects experience from the policy being optimised. On-policy algorithms are particularly suited for evolving training regimes such as curriculum learning (CL).

PPO belongs to a family of DRL algorithms known as actor-critic methods [\cite{sutton_reinforcement_1998}] where a neural network policy $\pi$ (the actor), determines the actions an agent should take in the environment, and a second neural network (the critic), evaluates the performance of the policy through a value function. The value function $V^\pi(s_t)$, predicts the discounted sum of future rewards given the current state, and running policy $\pi$:


\begin{equation}
    V^\pi(s_t) = \mathbb{E}_{\tau \sim \pi}[R(\tau) | s_0 = s_t]
\end{equation}

The advantage is a zero centered indication of how much better or worse the current policy is performing compared to what was expected. The advantage is calculated from $A^\pi(s_t) = Q^\pi(s_t) - V^\pi(s_t)$, where $Q^\pi(s_t) = \mathbb{E}_{\tau \sim \pi}[R(\tau) | s_0 = s_t, a_0 = a_t] $ provides an estimate of the sum of future rewards after taking $a_t$. The approximated advantage $\hat{A}^\pi$ can be determined by the temporal difference (TD) error: 

\begin{equation}
    \hat{A}^\pi(s_t) = r_t + \gamma V^\pi(s_{t+1}) - V^\pi(s_t) 
\end{equation}

In practice, PPO uses general advantage estimation (GAE) with Monte-Carlo returns (non-truncated episode reward) [\cite{schulman_gae_2016}], to provide a low variance estimate of the advantage. 

PPO is a trust region method, ensuring that policy updates do not cause the new policy to perform very different actions to the policy that was used to collect the data. The parameters of the policy $\pi_\theta$ are updated based on the following clipped loss function:

\begin{equation}
    L^{CLIP}(\theta) = \hat{\mathbb{E}}_t[\text{min}(\hat{r}_t(\theta)\hat{A}_t, \text{clip}(\hat{r}_t(\theta), 1 - \epsilon, 1 + \epsilon)\hat{A}_t)]
\end{equation}

where $\epsilon$ is a tuned clipping parameter, and $\hat{r}_t(\theta)$ is a ratio of the probability of action $a_t$ with the current policy $\pi_\theta(s_t)$, and the policy from the previous update $\pi_{\theta_{old}}(s_t)$:
\begin{equation}
    \hat{r}_t(\theta) = \frac{\pi_\theta(a_t | s_t)}{\pi_{\theta_{old}}(a_t | s_t)} 
\end{equation}

\begin{figure*}[htb!]
\centering
\includegraphics[width=0.8\columnwidth]{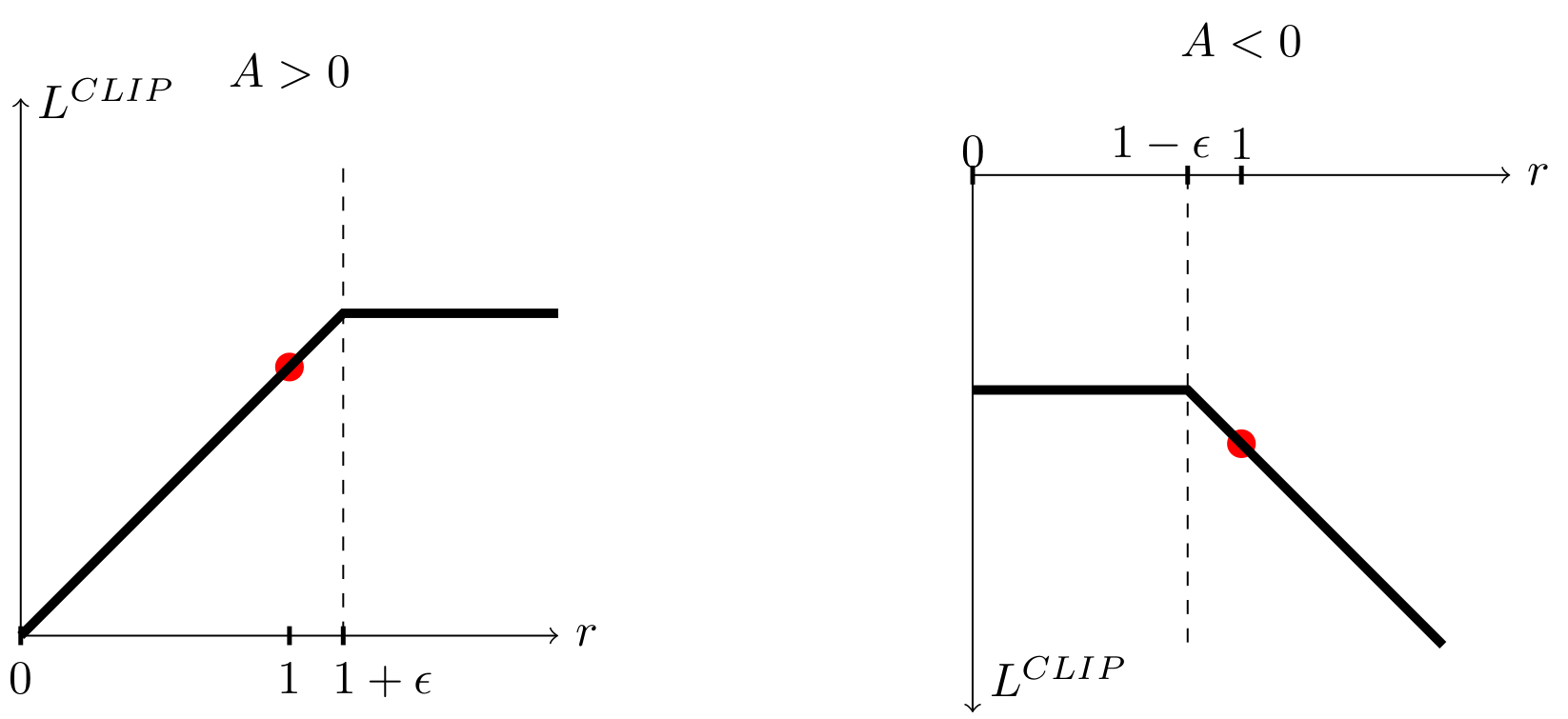}
\caption{Plots showing the behaviour of the $L^{CLIP}$ loss function of PPO, when the policy performed better than expected $A>0$, and worse than expected $A<0$.}
\label{fig:ch2_ppo}
\vspace{-3mm}
\end{figure*}

The \textit{min} in the loss function ensures that the clipping is one sided depending on the sign of the advantage (shown in Figure \ref{fig:ch2_ppo}). Intuitively, if the robot did better than expected, the loss function prevents too large of a step in the positive direction, but if the robot did worse than expected, and the action is now more probable with the new policy, the update can be completely reversed. Therefore, policy updates remain close to the policy that generated the data.

While there have been improvements in the sample efficiency of reinforcement learning algorithms in recent years [\cite{haarnoja_soft_2018, fujimoto_td3_2018, Song2020V-MPO, chen_randomized_2021}], PPO is still commonly used for its simplicity, ease of distributed training, and the benefits of on-policy learning. For example, \cite{rudin_learning_2021} recently used PPO to train a walking policy for a quadruped from just 20 minutes of training using parallel workers on a desktop GPU. The training process included a game-inspired curriculum of increasing terrain complexity, and the resulting policy was able to walk up and down stairs with an ANYmal-C quadruped in real-world experiments. \cite{xie_feedback_2018} used PPO to learn a policy that takes robot state and reference motion as input, and outputs a delta added to the reference motion. This method was able to control a simulated version of the biped Cassie over undulating terrain, and was then applied to the real biped, displaying reduced recovery time after stepping on an unexpected obstacle [\cite{xie_iterative_2019}].

\subsection{Curriculum Learning}

Curriculum learning is the concept of learning from a simpler form of the task, then slowly increasing the complexity \cite{elman_learning_1993}. Many examples of curriculum learning (CL) are present throughout our lives, from progressing through school, playing sport, or learning a musical instrument [\cite{narvekar_curriculum_2020}]. CL has been applied to a wide range of machine learning domains by providing training data to the learning algorithm in order of increasing difficulty [\cite{yu_learning_2018}]. These ideas were applied to learning grammatical structure by starting with a simple subset of data and gradually introducing more difficult samples [\cite{elman_learning_1993}]. \cite{bengio_curriculum_2009} highlight that for complex loss functions, CL can guide training towards better regions, helping find more suitable local minima. This results in a faster training time and better generalisation, demonstrated on a shape recognition task, and a natural language task. A summary of curriculum learning methods for deep reinforcement learning (DRL) can be found in \cite{narvekar_curriculum_2020}, and the blog post by \cite{weng_curriculum_2020}.

CL has been applied in continuous control domains such as robotics. \cite{sanger_neural_1994} applied Trajectory Extension Learning, where the desired trajectory for a two joint robot arm slowly moves the robot from a region of solved dynamics to a region where the dynamics are unsolved. The work by \cite{mendoza_curriculum_2017} shows that progressively increasing the number of controllable joints, incrementally moving a robot further from a target, and reducing joint velocities improves learning on a Jaco robot arm.

\begin{figure*}[t!]
\centering
\includegraphics[width=0.8\columnwidth]{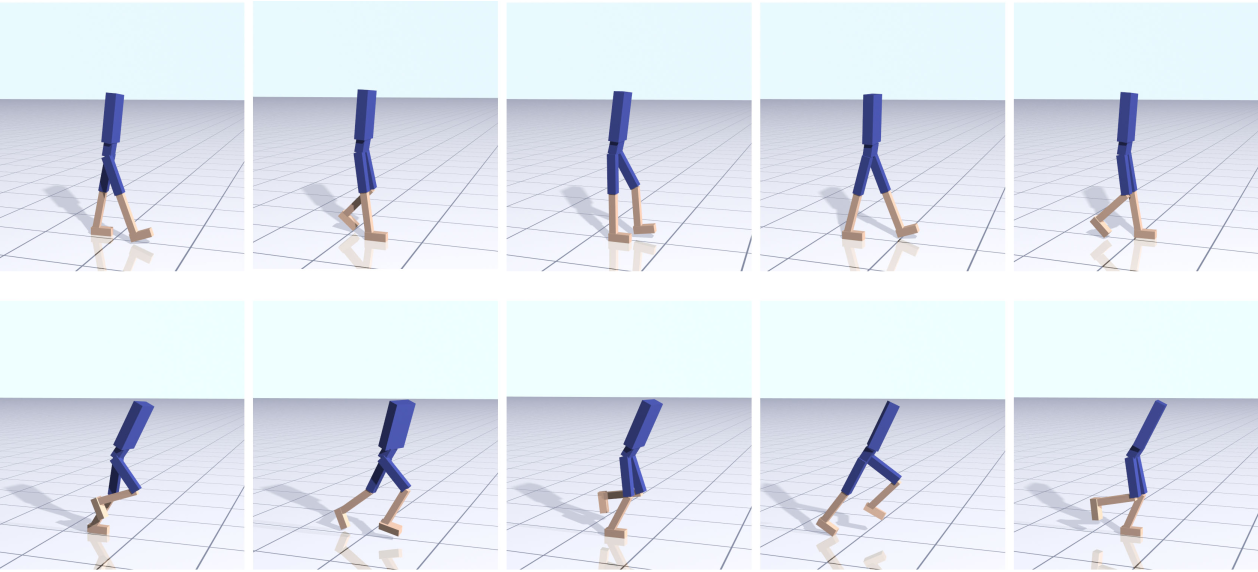}
\caption{Simplified biped walking (top) and running (bottom) [\cite{yu_learning_2018}]. 
}
\label{fig:ch2_yu_learning1}
\vspace{-3mm}
\end{figure*}

\begin{figure}[h!tb]
\centering
\subfloat[]{\includegraphics[width=0.35\columnwidth]{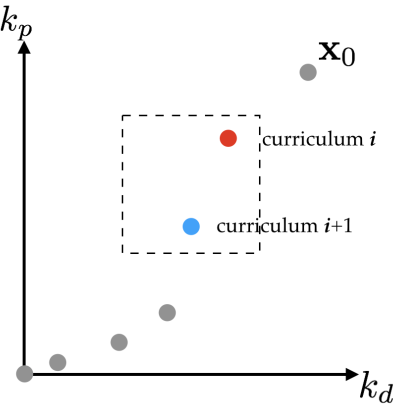}}
\hspace{0.5cm}
\subfloat[]{\includegraphics[width=0.35\columnwidth]{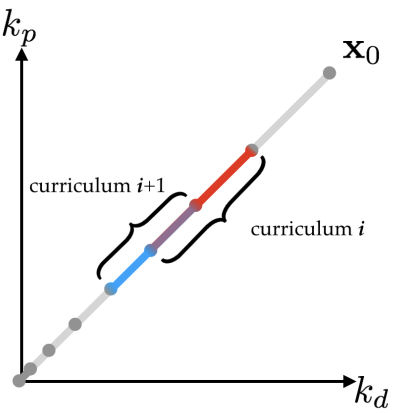}}
\caption{(a) The learner-centered curriculum determines the lessons adaptively based on the current skill level of the agent. (b) Environment-centered curriculum follows a series of predefined lessons [\cite{yu_learning_2018}].}
\label{fig:ch2_yu_learning2}
\end{figure}

For agents that have an unconstrained base, such as bipeds, several methods employ base stabilisation forces [\cite{van_de_panne_controller_1992, wu_terrain-adaptive_2010, yu_learning_2018}]. \cite{yu_learning_2018} demonstrate that employing a virtual assistant that applies external forces to stablise the robot base and encourage forward motion reduces training time and increases the asymptotic return of walking and running gaits for several simulated bipedal actors (Figure \ref{fig:ch2_yu_learning1}). The force applied by the virtual assistant is reduced as a success criteria is reached, for example, if the robot has not fallen after a nominal number of seconds. This work discusses different options for curriculum design, including a learner centered curriculum, and an environment-centered curriculum (Figure \ref{fig:ch2_yu_learning2}). CL was demonstrated to allow for a higher energy penalty on the magnitude of joint torques, without being detrimental to learning.

Recent work has seen the adoption of curriculum learning for increasing terrain difficulty [\cite{xie_allsteps_2020, lee_learning_quad_2020, rudin_learning_2021}], as well as a curriculum on reward parameters [\cite{hwangbo_learning_2019}]. \cite{hwangbo_learning_2019} first allow a quadrupedal robot to learn any movements that result in forward motion. During training the reward evolves to encourage actions that produce a desirable walking gait. \cite{lee_learning_quad_2020} use teacher-student learning to first train a policy with privileged information about the robot state, then train a student using restricted state input through imitation learning. This work applies an adaptive curriculum to the terrain to learn locomotion policies for a real quadruped over stairs, steps and hills.

\section{Real-World Applications}

This section considers literature around robot navigation and passing through small gaps. The recently concluded DARPA Subterranean Challenge [\cite{darpa_darpa_2021}] highlighted the need for agents exploring underground environments to autonomously pass through small gaps like doorways. Costmaps are used by path planners to construct trajectories to traverse difficult terrain while considering the kinematic limitations of the robot [\cite{hudson_heterogeneous_2021}]. However, the task becomes challenging when gaps that are marginally wider than the robot do not appear as traversable through common perception modalities, such as low resolution occupancy maps. Distinguishing when a gap is suitable to traverse, and designing a behaviour to pass through is an open challenge for mobile platforms.

Classical methods for autonomous navigation struggle to traverse small gaps. Artificial potential fields (APF) [\cite{khatib_real-time_1986}] is a widely used algorithm for navigation, where the robot is attracted to goal locations, and repelled from obstacles. Vector Field Histograms (VHF) generate the steering angle based on the polar density of surrounding obstacles [\cite{borenstein_vector_1991}]. Bug algorithms have also been used extensively [\cite{mcguire_comparative_2019}], using the idea that the robot moves toward a goal unless an obstacle is encountered, then the robot moves along the obstacle boundary until it can once again move toward the goal. Small gaps are a limitation for each of these methods. Local minima with narrow gaps cause oscillations when the attractive and repelling forces are balanced [\cite{koren_potential_1991}]. 

\begin{figure}[t!]
\centering
\subfloat[]{\includegraphics[width=0.3\columnwidth]{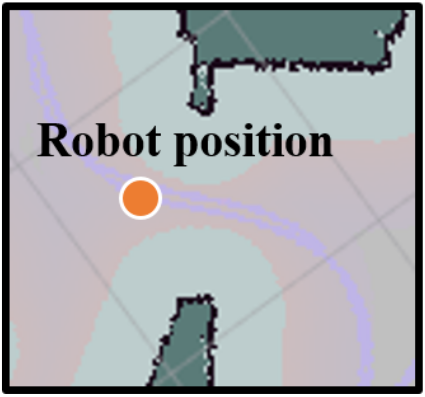}}
\subfloat[]{\includegraphics[width=0.3\columnwidth]{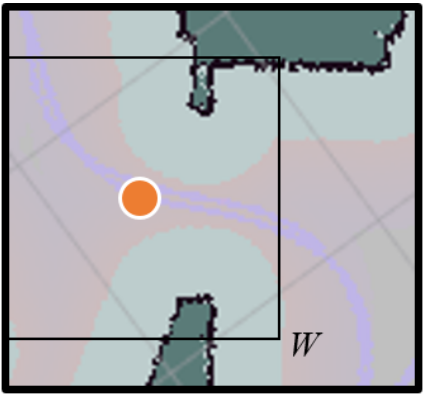}}
\subfloat[]{\includegraphics[width=0.3\columnwidth]{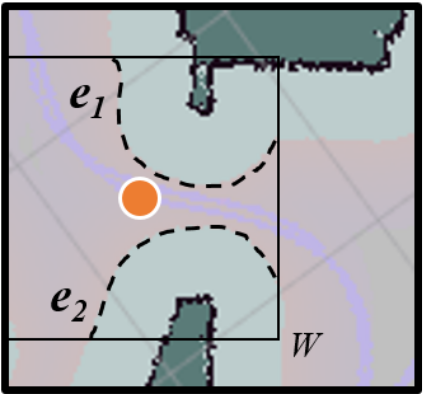}}
\hfill
\subfloat[]{\includegraphics[width=0.3\columnwidth]{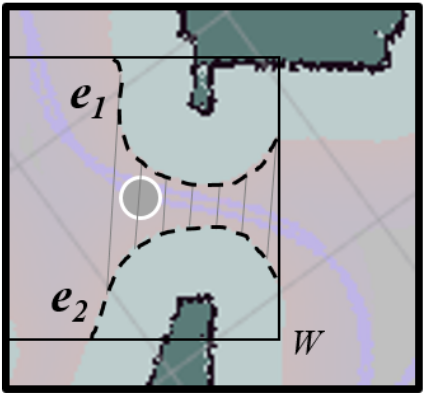}}
\subfloat[]{\includegraphics[width=0.3\columnwidth]{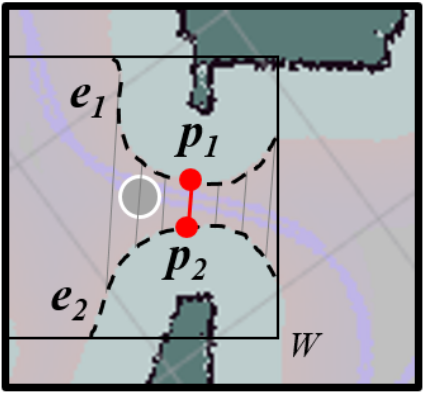}}
\subfloat[]{\includegraphics[width=0.3\columnwidth]{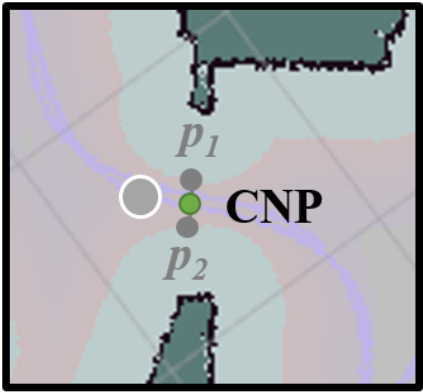}}
\caption{Detection of a critical navigation point (CNP) in a costmap. All distances between two near edges are computed, and the points yielding the minimum distance are selected. The CNP is finally placed
at the midpoint. [\cite{moreno_automatic_2020}].}
\label{fig:cha2_moreno}
\end{figure}

Many solutions have been proposed for the narrow gaps problem. \cite{zheng_sun_narrow_2005} introduce a bridge test to detect narrow passages for use with probabilistic roadmap (PRM) planning. Planners like PRM and rapidly-exploring random trees (RRTs) rely on verifying path queries with a robot footprint, and therefore also have difficulty with narrow spaces perceived as smaller than the robot. \cite{mujahed_admissible_2018} proposed an admissibility gap: a virtual gap that satisfies the kinematic constraints of the vehicle to plan through tight spaces. In urban environments, \cite{rusu_laser-based_2009} designed a geometric door detection method using 3D point cloud information, provided the door was within a set of specifications. With this method, a PR2 robot was able to detect a door and its handles [\cite{meeussen_autonomous_2010}], as well as negotiate the open or closed configuration to get through [\cite{chitta_planning_2010}]. \cite{cosgun_context_2018} detect doors from door signs, and direct a robot through with pointing gestures. \cite{moreno_automatic_2020} generate auxiliary waypoint locations (critical navigation points (CNP)) at problematic narrow regions, a depiction is provided in Figure \ref{fig:cha2_moreno}. Sensory errors and low resolution perception maps contribute to the failure of these methods 


Classical navigation methods often rely on extensive manual tuning of parameters [\cite{khatib_real-time_1986, hines_virtual_2021, mcguire_comparative_2019}]. \cite{hines_virtual_2021} use the concept of a virtual surface to slowly approach negative obstacle candidates in the case of perception uncertainty. This work also introduces a behaviour stack, where each behaviour has a priority, and an admissibility criteria. A behaviour becomes active if it has a higher priority than the current behaviour, and is admissible. The primary path follow method uses Hybrid A* [\cite{dolgov_path_2010}] to plan a path through a costmap to a goal, considering the vehicle constraints. Admissibility criteria, and the behaviours themselves, are manually designed and require considerable tuning. 

Designing an autonomous navigation system typically requires extensive manual tuning to handle difficult terrain scenarios. Learning methods used for navigation have shown promising results for mobile robots in human-centered domains [\cite{rana_multiplicative_2020,bansal_combining_2019,kumar_learning_2019,gupta_cognitive_2019}], and in field applications [\cite{kahn_badgr_2020}]. \cite{rana_multiplicative_2020} learn a navigation policy using a prior controller to guide exploration, and decreasing the dependence on the prior as training progresses. This method efficiently utilises an APF controller while using an ensemble of policies to determine a confidence for running the learned controller. With learning methods it is possible to use raw RGB images to navigate on various surfaces [\cite{kahn_badgr_2020}], and through narrow gaps [\cite{bansal_combining_2019}]. BADGR by \cite{kahn_badgr_2020} is a method that trains a navigation policy directly from images with a self supervised approach that not only learns to navigate to a waypoint, but also to favour travelling on a concrete path over grass. \cite{bansal_combining_2019} learn to navigate through narrow doorways to get to a specified goal location using a monocular camera mounted to a robot. \cite{gupta_cognitive_2019} use end-to-end reinforcement learning to learn an egocentric map and how to plan to a goal. \cite{kumar_learning_2019} learn to plan at a high level of abstraction (turn left, turn right) using video data of expert trajectories. While learning methods reduce some of the manual tuning seen with traditional control methods, learned behaviours are typically challenging to integrate with other controllers. 

\subsection{Sim-to-real Transfer of Locomotion Policies}

Due to training inefficiencies, learning methods are predominately trained in simulation and transferred to the real world. For scenarios with low sim-to-real disparity, policies trained in simulation have been directly transferred to real-world platforms [\cite{tai_virtual--real_2017, xie_learning_2018, rana_residual_2020}], while more complex systems, for example dynamic legged platforms, require system identification [\cite{tan_sim--real_2018}], modelling with real world data [\cite{hwangbo_learning_2019, peng_learning_2020, rudin_learning_2021, miki_learning_2022}], and heavy dynamics randomisation [\cite{peng_learning_2020}] during training to allow for policies to jump the reality gap. Laser base sensors are less subjected to sim-to-real disparity than cameras and other sensors [\cite{rana_multiplicative_2020}]. \cite{tai_virtual--real_2017} train a policy that learns to avoid obstacles and move to a target using a mobile robot with a laser depth sensor. \cite{xie_learning_2018} learn to switch between a simple controller and a trained policy for reaching a target in the robot frame. This work introduced Assisted DPPG (AsDDPG), an extension of deep deterministic policy gradient (DDPG [\cite{Lillicrap2016ContinuousCW}]), where the simple controller provides an exploration policy for efficient learning. \cite{rana_residual_2020} train a navigation policy that sums with an APF prior controller to improve the performance of the prior controller. For each of these examples, policies train in simulation transfer directly to the real robot with no fine tuning. However, each example used laser based perception to operate on a flat surface with a simple differential drive robot (Turtlebot [\cite{tai_virtual--real_2017}], Pioneer [\cite{xie_learning_2018}], PatrolBot mobile [\cite{rana_residual_2020}]). For complex scenarios, direct sim-to-real transfer is likely to result in a drop in performance when deployed on a real platform.

\section{Behaviour Composition}

Often it is impractical to design a single controller for all expected conditions. Instead, an arbiter is typically used to determine which individual controller to activate. Importantly, controller switching can only occur when the robot is in a suitable state for the upcoming behaviour. This review considers literature on behaviour composition from classical methods, learning methods, and hierarchical reinforcement learning.

\subsection{Sequential Composition}

\begin{figure*}[!t]
\centering
\includegraphics[width=0.5\columnwidth]{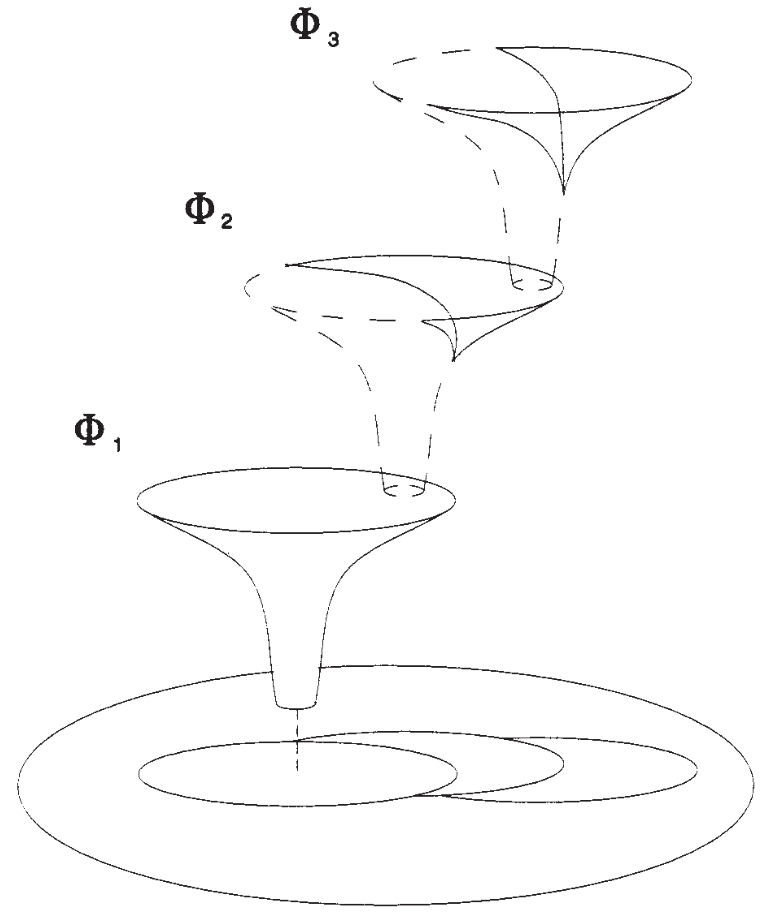}
\caption{The sequential composition of controllers [\cite{burridge_sequential_1999}]. Each
controller is only active in the part of its domain that is not
already covered by those nearer the goal. Here, $\Phi_3 \succeq \Phi_2 \succeq \Phi_1$ , and $\Phi_1$ is the goal controller.}
\label{fig:cha2_sequential}
\vspace{-3mm}
\end{figure*}

Understanding when a behaviour should be activated is an important property of behaviour-based robotics [\cite{arkin_behavior-based_1998}]. Classical methods for behaviour selection rely on understanding the model of the system, and knowing the set of states for which a given controller will operate and safely converge to a target behaviour. A key component is determining the domain of attraction (DoA) or region of attraction (RoA) of each controller, defined as the set of states, from which activating a specified controller will converge to a designated behaviour defined for that controller. Formally, if each controller $\Phi_i$ has a goal set (or invariant set) $\mathcal{G}(\Phi_i) = \{s_i^*\}$ that defines all of the states the controller operates nominally, then the RoA for $\Phi_i$ is given by:

\begin{equation}
    \mathcal{R}(\Phi_i) = \{s_0 \in \mathcal{S} : \lim_{t\to\infty} s(t,s_0) = \mathcal{G}(\Phi_i)\}
\label{eq:roa}
\end{equation}

Controllers can be composed in such a way that the goal set for one controller exists in the RoA of another, thus activating controllers in the correct sequence will drive the robot toward a desired goal behaviour. This sequencing of controllers is known as sequential composition [\cite{burridge_sequential_1999}]. In this work, composing controllers in sequence was demonstrated on a juggling robot. The sequential composition framework is illustrated in Figure \ref{fig:cha2_sequential}.

\begin{figure*}[!t]
\centering
\includegraphics[width=0.6\columnwidth]{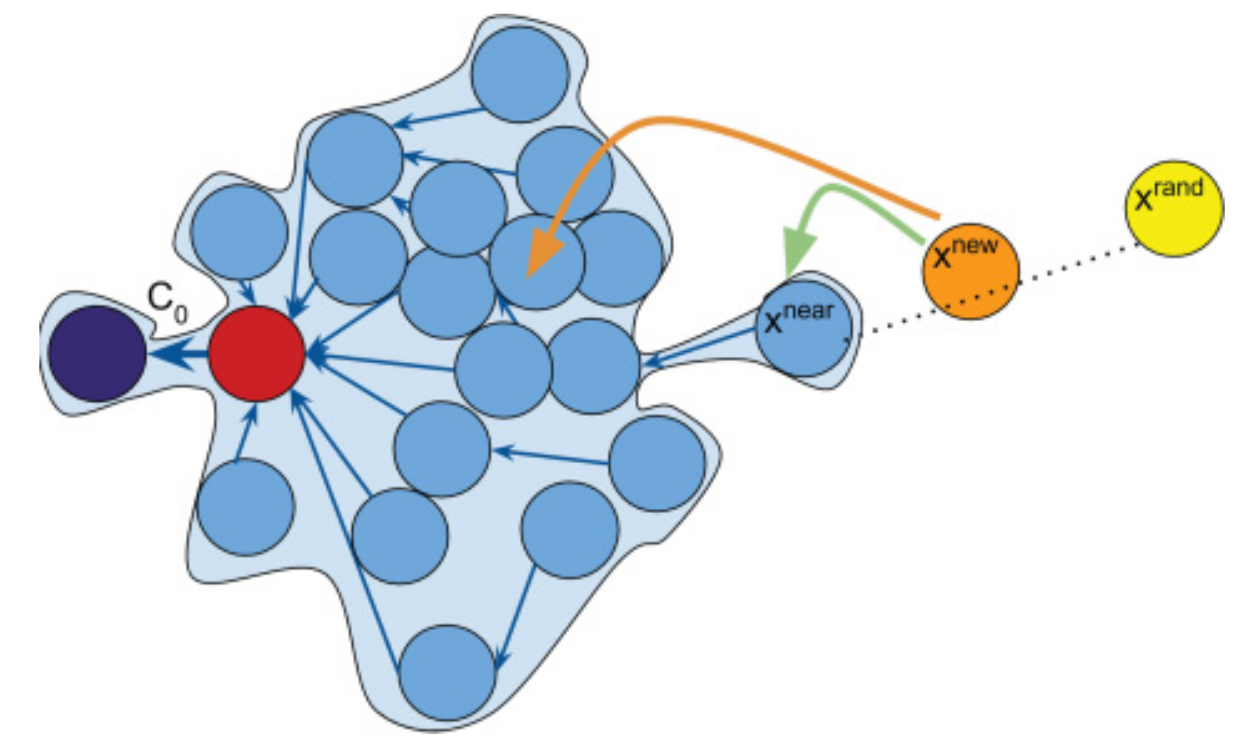}
\caption{RoA expansion by sampling with an RRT [\cite{borno_domain_2017}].
}
\label{fig:ch2_borno}
\vspace{-3mm}
\end{figure*}

Estimating the RoA overlap between controllers can be difficult for robots with high state spaces. The problem can be simplified by defining a set of pre and post conditions for each controller [\cite{faloutsos_composable_2001}], defining fixed length control sequences [\cite{peng_deepmimic_2018}], or by providing a rule-based bound on parameters like heading angle and switching frequency [\cite{gregg_control_2012}]. \cite{manchester_2011_regions} estimate the RoA for a walking robot by decomposing the dynamics to find the regions of orbital stability for a compass gait system. \cite{motahar_composing_2016} switch between straight, left turn, and right turn controllers operating a 3D simulated biped using a reduced dimension hybrid zero dynamics control law. These methods result in combinations of complex primitives, though they require a mathematically defined RoA, or a hand designed switching criteria for each primitive. RoA expansion can provide a greater overlap between controllers. \cite{borno_domain_2017} estimate the RoA for a simulated humanoid using multiple forward passes of a dynamics model simulation. Using rapidly exploring random trees (RRTs), the closest state within the RoA of a given controller is found, and the new trajectory included in the RoA of the controller (Figure \ref{fig:ch2_borno}). As task complexity increasing, understanding the RoA for a controller becomes difficult. Current methods that rely on simplified representations limit the capabilities of the controller and reduce the functionality of the robot in difficult dynamic scenarios.

\subsection{Learning Behaviour Composition}

Behaviour composition has also been demonstrated for learned behaviours. Combining DRL primitives usually involves training a deep Q network (or a similar discrete switching network) that selects which primitive to use [\cite{liu_learning_2017,merel_hierarchical_2019,lee_robust_2019}], or learning a complex combination of primitives [\cite{peng_mcp_2019}]. \cite{liu_learning_2017} combine control fragments with a Q-learning scheduler that enables highly complex simulated characters to perform balancing tasks. \cite{merel_hierarchical_2019} use motion capture to train primitives for a simulated humanoid. Each primitive is spliced into small fragments of known length, and a control fragment selector utilises perception to determine which controller to use at each interval. \cite{lee_robust_2019} train a switching policy for several complex behaviours for a quadruped to recover from a fall, stand, and continue walking. \cite{peng_mcp_2019} learn how to combine pre-trained motion primitives for a simulated humanoid using multiplicative compositional policies (MCP). Combining several pre-trained policies using a complex combination of actions results in the smooth transition between behaviours and the emergence of new behaviours. \cite{yang_multi_2020} train a gating neural network to blend several separate expert neural network policies to perform trotting, steering, and fall recovery on a real quadruped. This framework can acquire more specialised skills by fusing network parameters, generating behaviours that are adaptable to unseen scenarios. \cite{lee_learning_2020} learn primitive skills for individual subsets of a complex task, then train a meta policy to compose behaviours to complete the task with a second agent. This work was demonstrated in simulation with two robot arms performing manipulation with multiple end-effectors, and with two quadrupeds pushing a block into position. In each of these examples, access to all expected environment conditions are required during training, and adding new primitives requires retraining of the selection policy, limiting the scalability of these methods in the real world.


\subsection{Hierarchical Reinforcement Learning}

Hierarchical reinforcement learning (HRL) separates tasks into subtasks using multiple levels of abstraction. Each subtask solves a subgoal, offering flexibility through architecture. It is common to train all segments of the hierarchy concurrently [\cite{sutton_between_1999, bacon_option-critic_2018, frans_meta_2017,peng_deeploco_2017,levy_learning_2019}], or to train parts of the hierarchy separately [\cite{merel_hierarchical_2019, lee_composing_2019}]. When trained together, HRL can improve task level outcomes, such as steering and object tracking [\cite{peng_deeploco_2017}], or learning efficiency by reusing low level skills across multiple high level tasks [\cite{frans_meta_2017}] or by utilising hindsight experience from explored trajectories [\cite{levy_learning_2019}]. When trained separately, HRL can break up a large difficult problem into smaller solvable subproblems [\cite{schaal_dynamic_2006}]. Low level controllers can be refined efficiently using prior knowledge, for example by utilising motion capture data [\cite{peng_deepmimic_2018, merel_hierarchical_2019, peng_mcp_2019}] or behavioural cloning [\cite{strudel_learning_2020}]. For robotics applications it may be difficult to develop controllers for multiple behaviours simultaneously. Policy sketches introduce a hierarchical method that uses task specific policies, with each task performed in sequence [\cite{andreas_modular_2017}]. CompILE uses soft boundaries between task segments [\cite{kipf_compile_2019}]. Work by \cite{peng_terrain-adaptive_2016} trains several actor-critic control policies, modulated by the highest critic value in an given state. \cite{sharma_learning_2020} use reinforcement learning to select a hierarchical subset of object-centric controllers for complex manipulation tasks such as turning a hex screw with a robot arm. For these examples there must be a reasonable state overlap between adjacent controllers.


Hierarchical methods are often useful for legged robots. \cite{da_learning_2020} develop a hierarchical method that uses a model-based controller to combine several low-level primitives with a high-level behaviour selector trained with reinforcement learning. Body pose, and relative foot position was provided as input to the selector policy, trained to minimise energy expenditure. This method was developed in simulation and deployed on a real robot with no randomisation or adaption scheme required. \cite{jain_pixels_2020} concurrently learn high and low level policies in a hierarchical framework for locomotion and navigation with a quadruped. The high level processes first-person camera images, and feeds a latent command to the low level controller to navigate curved cliffs and a maze. 

With hierarchical methods, policies can be integrated in complex ways, combining pre-trained behaviours by learning latent representations of skills [\cite{pertsch_accelerating_2020}] or primitives [\cite{ha_distilling_2020}] and through interpolation in the latent space. From an offline dataset of experience, \cite{pertsch_accelerating_2020} were able to combine low level controllers for manipulation tasks and a locomotion task for a multi-legged agent. \cite{ha_distilling_2020} utilise motion capture to learn latent representations of primitives, then use model predictive control for the high level navigation of a high dimensional humanoid. The FeUdal approach learns a master policy that modulates low-level policies using a learned goal signal [\cite{vezhnevets_feudal_2017}]. Interpolation between behaviours yields natural transitions [\cite{xu_hierarchical_2020}], however in each of these approaches, experience from all behaviours must be available during training. 

Hierarchical methods can be used to break a difficult task into solvable subtasks, similar in concept to CL where tasks are presented in order of complexity. CL and HRL can be combined to achieve complex behaviours, and improve learning efficiency. \cite{karpathy_curriculum_2012} train a dynamic articulated figure called an ``Acrobot" with a curriculum of increasingly difficult maneuvers. In this work, a low level curriculum learns a specific skill, and a high level curriculum specifies which combination of skills are required for a given task. 

\begin{figure}[t!]
\centering
\subfloat[RoA and goal sets of a sequential
composition controller, where $x_0 \in \mathcal{D}(\Phi_1), x_d \in \mathcal{D}_2(\Phi_2)$, but controller $\Phi_1$ does not prepare controller $\Phi_2$.]{\includegraphics[width=0.5\columnwidth]{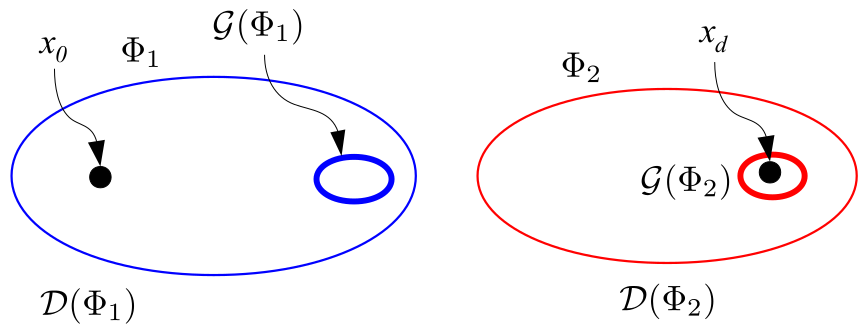}}
\hspace{0.5cm}
\subfloat[RoA goal sets of the controllers after training (the green ellipse is the trained behaviour).]{\includegraphics[width=0.45\columnwidth]{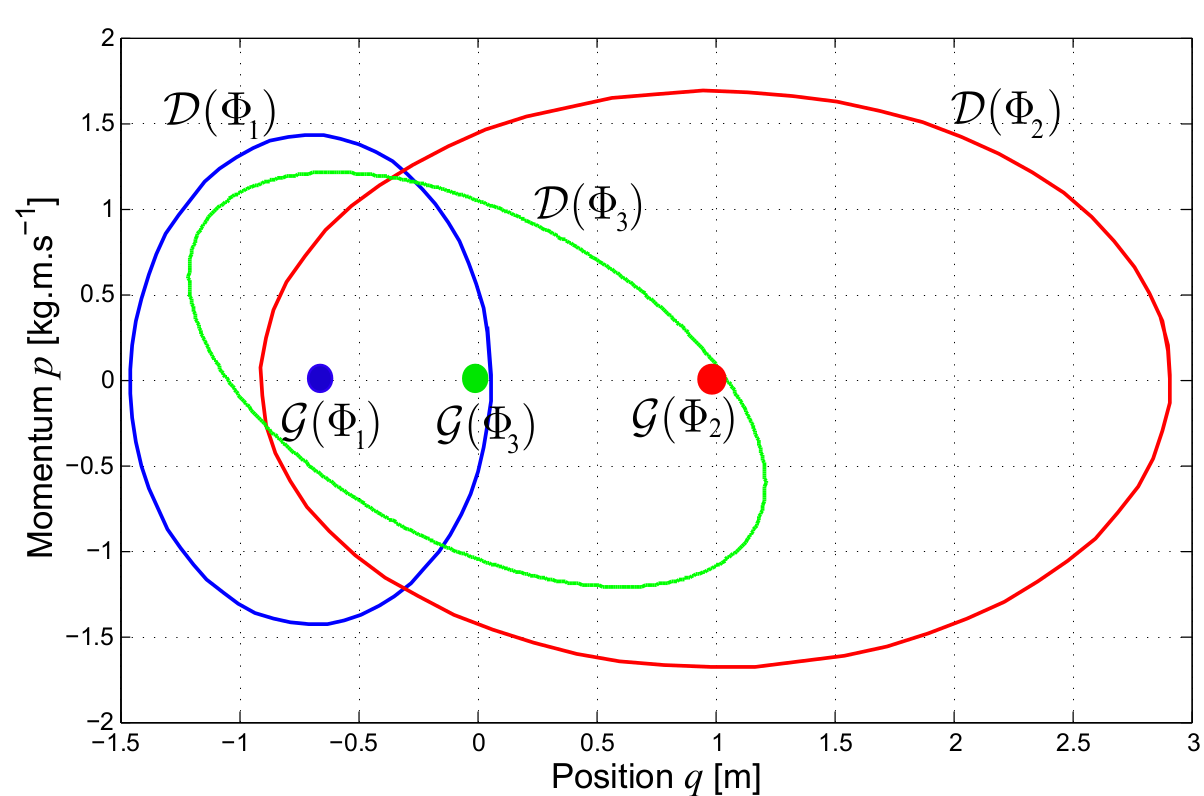}}
\caption{Learning sequential composition [\cite{najafi_learning_2016}].}
\label{fig:cha2_learning_sequential}
\end{figure}

\section{Learning Transition Behaviours}

For behaviours without a RoA overlap, transition policies can bridge sequential controllers so they can be combined. Transition policies have been integrated with classical methods to move a robot through a state of difficult to model dynamics. Estimating the RoA of a policy is possible for systems with a relatively small state space, such as the single [\cite{berkenkamp_safe_2017, najafi_learning_2016}], or double inverted pendulum [\cite{randlov_combining_2000}], however for more complex systems the RoA is difficult to determine. Where it is possible to differentiate between unmodeled regions of the state from those that are well behaved, DRL can guide an agent to where a classical controller can take over [\cite{najafi_learning_2016}], [\cite{randlov_combining_2000}]. \cite{najafi_learning_2016} train a policy to swing an inverted pendulum to a state where the upright controller can take over. Figure \ref{fig:cha2_learning_sequential}.a) shows a depiction of the upright, and downward controllers ($\Phi_1$ and $\Phi_2$), without an overlap. In this work, a separate controller was trained with DRL to create an overlap between the existing controllers ($\Phi_3$, shown in green in Figure \ref{fig:cha2_learning_sequential}.b). The new behaviour $\Phi_3$, enables the pendulum to move from the downward position ($\Phi_2$), through the goal set of controller $\Phi_3$, to the RoA of controller $\Phi_1$, and finally the upward position (the goal state of controller $\Phi_1$).

\begin{figure}[!t]
\centering
\subfloat[\cite{lee_composing_2019} design transition policies. A meta-policy chooses a primitive policy of index c, the corresponding transition policy helps initiate the chosen primitive policy, the primitive policy executes the skill, and a success or failure signal for the primitive skill is produced.]{\includegraphics[width=0.95\columnwidth]{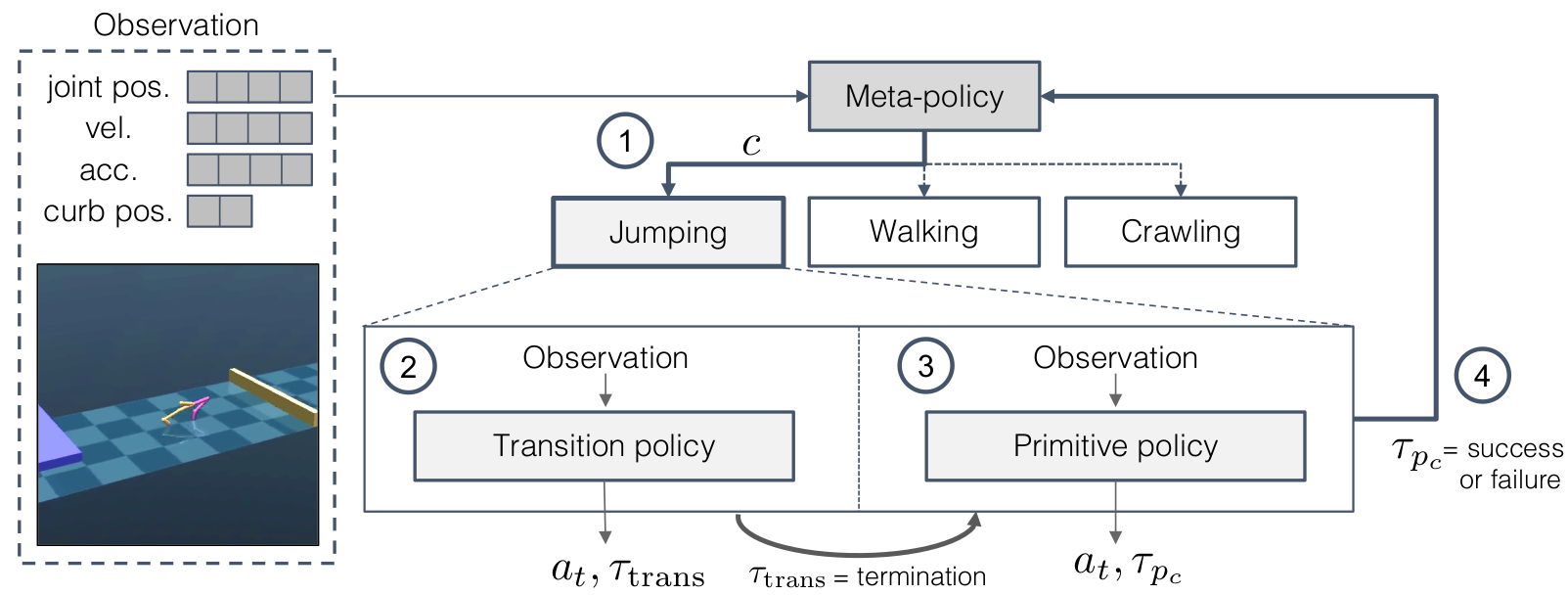}}
\hfill
\subfloat[A proximity predictor is trained on states sampled from the two buffers to output the proximity to the initiation set. The predicted proximity serves as a reward to encourage the transition policy to move toward good initial states for the corresponding primitive policy.]{\includegraphics[width=0.95\columnwidth]{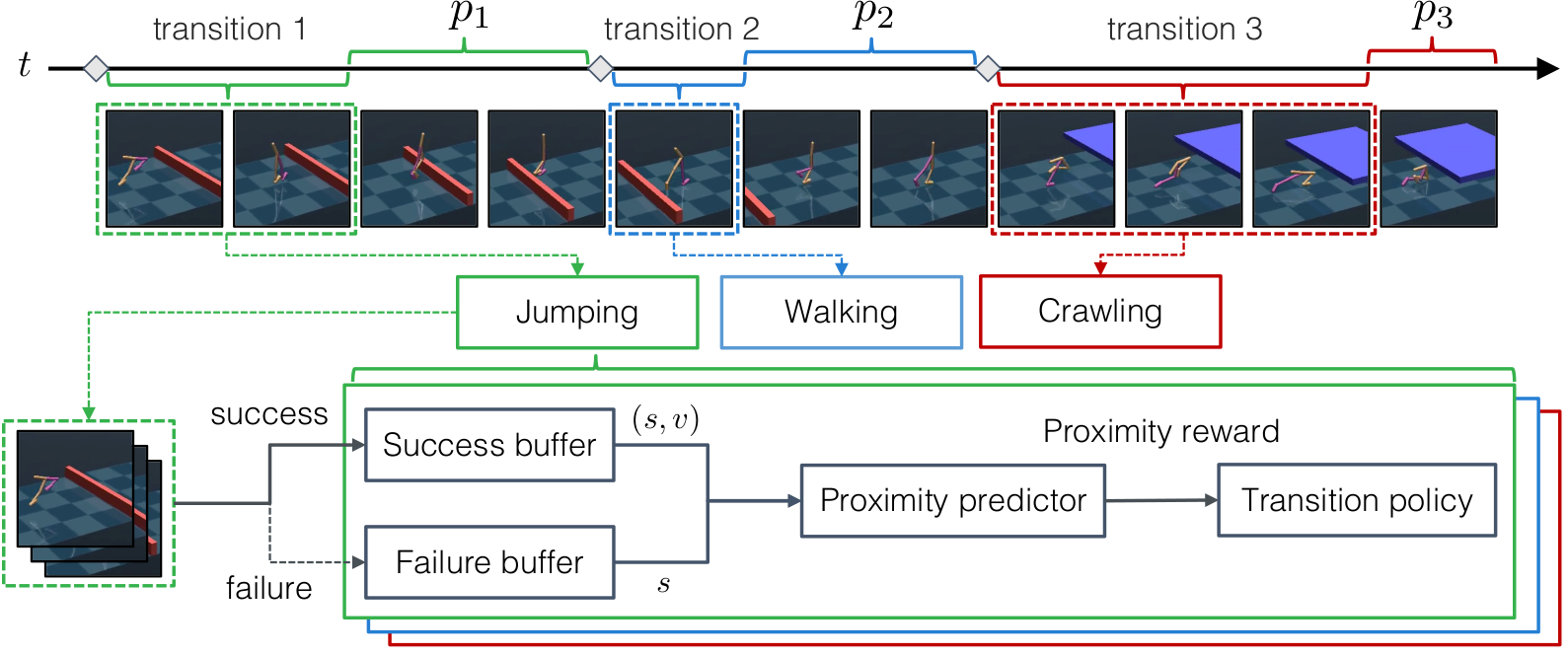}}
\caption{Learning transition policies [\cite{lee_composing_2019}].}
\label{fig:cha2_lee_composing}
\end{figure}

Transition policies can improve the stability of switching between complex behaviours where the RoA is not well-understood. \cite{lee_composing_2019} learn a proximity predictor to train a policy to transition a 2D biped to the initial state required by jumping, walking, and crawling controllers. Figure \ref{fig:cha2_lee_composing}a) outlines the scenario where transition policies are used, and Figure \ref{fig:cha2_lee_composing}b) shows the training procedure. Each policy has a transition policy, trained by collecting data of successful transitions. These buffers are used to train a proximity predictor function $P_\omega(s)$ that predicts how close the robot is from a transition state that would result in the successful traversal of the obstacle. Success is a defined parameter relating to the task, for example, if the robot has past the obstacle by 1.5m it was deemed successful. Equation \ref{eq:cha2_lee_composing1} shows the loss function for the proximity predictor:

\begin{equation}
    L_P(\omega,\mathcal{B}^S, \mathcal{B}^F) = \frac{1}{2}\mathbb{E}_{(s,v) \sim \mathcal{B}^S}[(P_\omega(s) - v)^2] + \frac{1}{2}\mathbb{E}_{(s,v) \sim \mathcal{B}^F}[P_\omega(s)^2] 
\label{eq:cha2_lee_composing1}
\end{equation}

Data is sampled from buffers of success and failures ($\mathcal{B}^S, \mathcal{B}^F$), $v$ is a linearly discounted function $v = \delta^{step}$. $\delta$ is a decay parameter, and $step$ refers to the number of timesteps required to reach the transition state. Using this prediction as a reward, transition policies were trained to maximise the reward function ($\gamma$ is a discount factor):

\begin{equation}
    R_{\text{trans}}(\phi) = \mathbb{E}_{(s_0,s_1,...,s_T) \sim \pi_\phi}[\gamma^TP_\omega(s_T) + \sum^{T-1}_{t=0}\gamma^t(P_\omega(s_{t+1}) - P_\omega(s_t))] 
\label{eq:cha2_lee_composing2}
\end{equation}

Learning to transition between behaviours allows separate controllers to be combined, particularly those without a reliable RoA overlap. The current methods are promising for composing sequential controllers, however, more investigation is required for robots with a high DoF performing complex behaviours.

\section{Summary}

Deep reinforcement learning (DRL) has been applied as a suitable alternative to traditional controller design for developing complex behaviours, however, sample inefficiencies and local minima limits the extensive application of DRL methods in real-world scenarios. Learning from expert demonstrations, for example from motion capture, can improve the outcomes of learning methods [\cite{peng_deepmimic_2018}], however, expert demonstrations can be costly to collect. Curriculum learning (CL) has recently been used for learning complex behaviours, particularly when training on challenging terrain. Through research question 1, this thesis investigates improving the sample efficiency of learning complex behaviours by exploring how simple trajectory guidance and CL can be applied for traversing difficult terrain.

The challenges of deploying policies trained in simulation in the real world have been investigated for many locomotion robots, however, further work is needed to improve the development of complex navigation behaviours. For example, passing through narrow gaps is a challenging problem when contact with the environment may be unavoidable or even necessary. Hand crafted behaviours are difficult to design in these cases. Furthermore, traditional path planning algorithms are suitable for many situations, learning methods need to be incorporated with these methods. Few works were found in the literature that consider these practical scenarios. This thesis investigates the challenge of traversing a narrow doorway with a real robot and integrating learned policies with traditional path planning methods.


In many examples from the literature, separate behaviours were required to perform complex tasks. Behaviours were composed to increase the capabilities of the robot from one task to many. Behaviour composition requires the RoA or switch conditions of a controller be defined explicitly, however, for learned policies it is difficult to know when a behaviour can be switched. Hierarchical learning is often used to determine which primitive to activate from a set, however, hierarchical methods require training with all behaviours, limiting the application to small sets of behaviours, and requiring retraining as new behaviours are introduced. Research question 2 investigates behaviour switching by learning switch conditions for complex policies and minimising retraining for new behaviours.

Complex behaviours may operate in very different areas of the state and have narrow or no RoA overlap with each other. Transition policies were found in the literature that encourage RoA overlap, however, transition behaviours for complex robots with dense perception modalities are yet to be developed. Research question 3 explores methods for learning visuo-motor transition behaviours that enable a complex robot to traverse a sequence of difficult terrains. 


\chapter[Guided Curriculum Learning]{Guided Curriculum Learning for Walking Over Complex Terrain}

\label{cha:ch3a}

\includepdf[pages=-,pagecommand={},width=\textwidth]{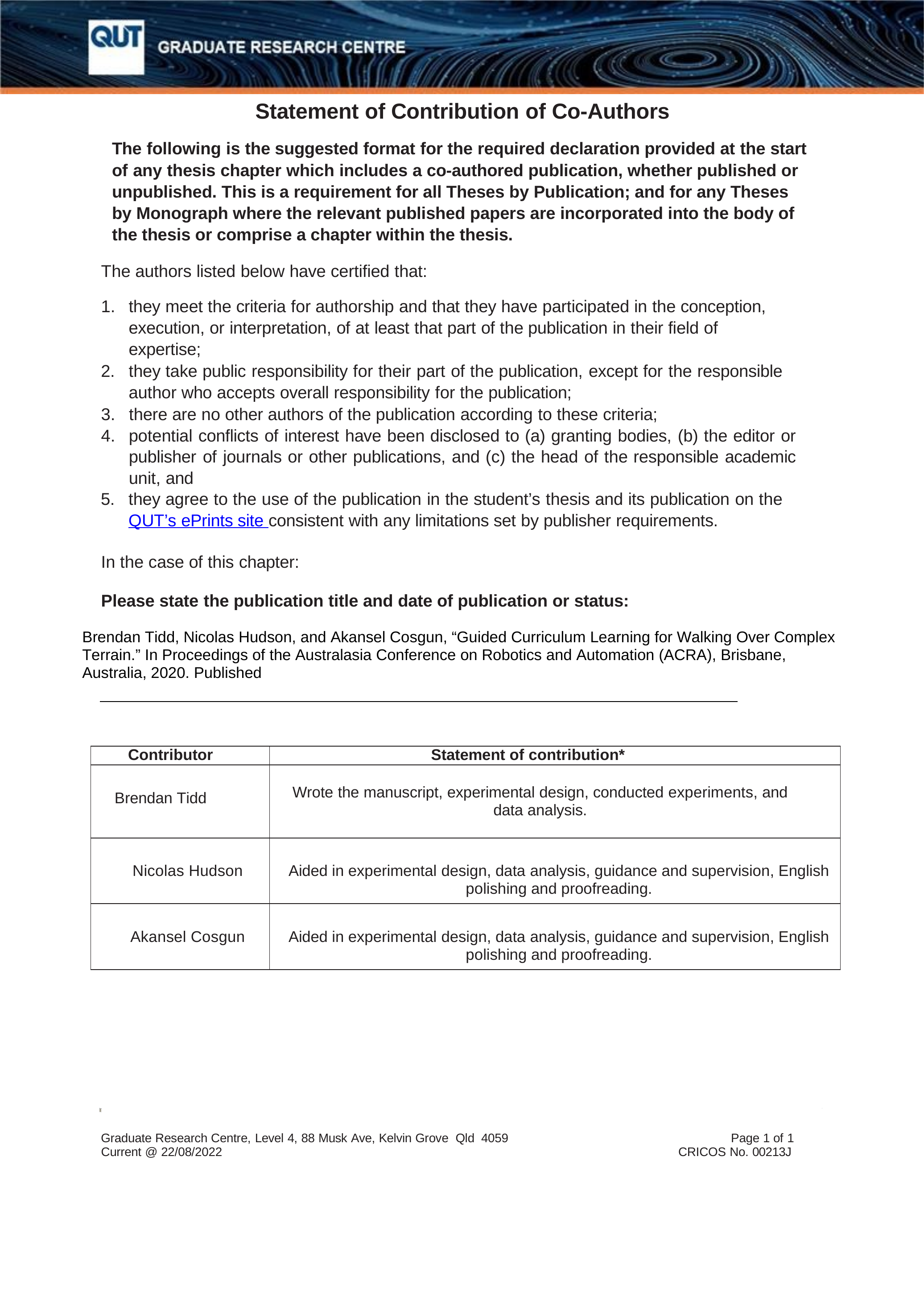}


This chapter presents the first investigation of research question 1: \textbf{\textit{how can complex visuo-motor locomotion behaviours be learned efficiently?}} Current learning methods can be inefficient to train, where efficiency refers to the duration of interaction between the agent and the environment to produce a suitable behaviour. To improve efficiency, methods utilise expensive expert demonstrations for each task [\cite{peng_deepmimic_2018}]. The method introduced in this chapter improves learning efficiency for training complex visuo-motor behaviours using a single walking demonstration. The contributions of this chapter are as follows:

\begin{itemize}
    \item A novel curriculum learning approach was developed for training behaviours for a dynamic biped in simulation. A simple walking trajectory consisting of a set of joint and body positions was sufficient to improve state exploration during training, resulting in behaviours for traversing a diverse set of difficult terrain types. Behaviours trained without the curriculum required far more training steps or were unable to learn a suitable policy, demonstrating improved learning efficiency with this method.
    
    
    
\end{itemize}

“Guided Curriculum Learning for Walking Over Complex Terrain” was
published and presented at the 2020 Australasian Conference on Robotics and Automation held in 
Brisbane, Australia.

\begin{figure*}[htb!]
\centering
\includegraphics[width=0.95\columnwidth]{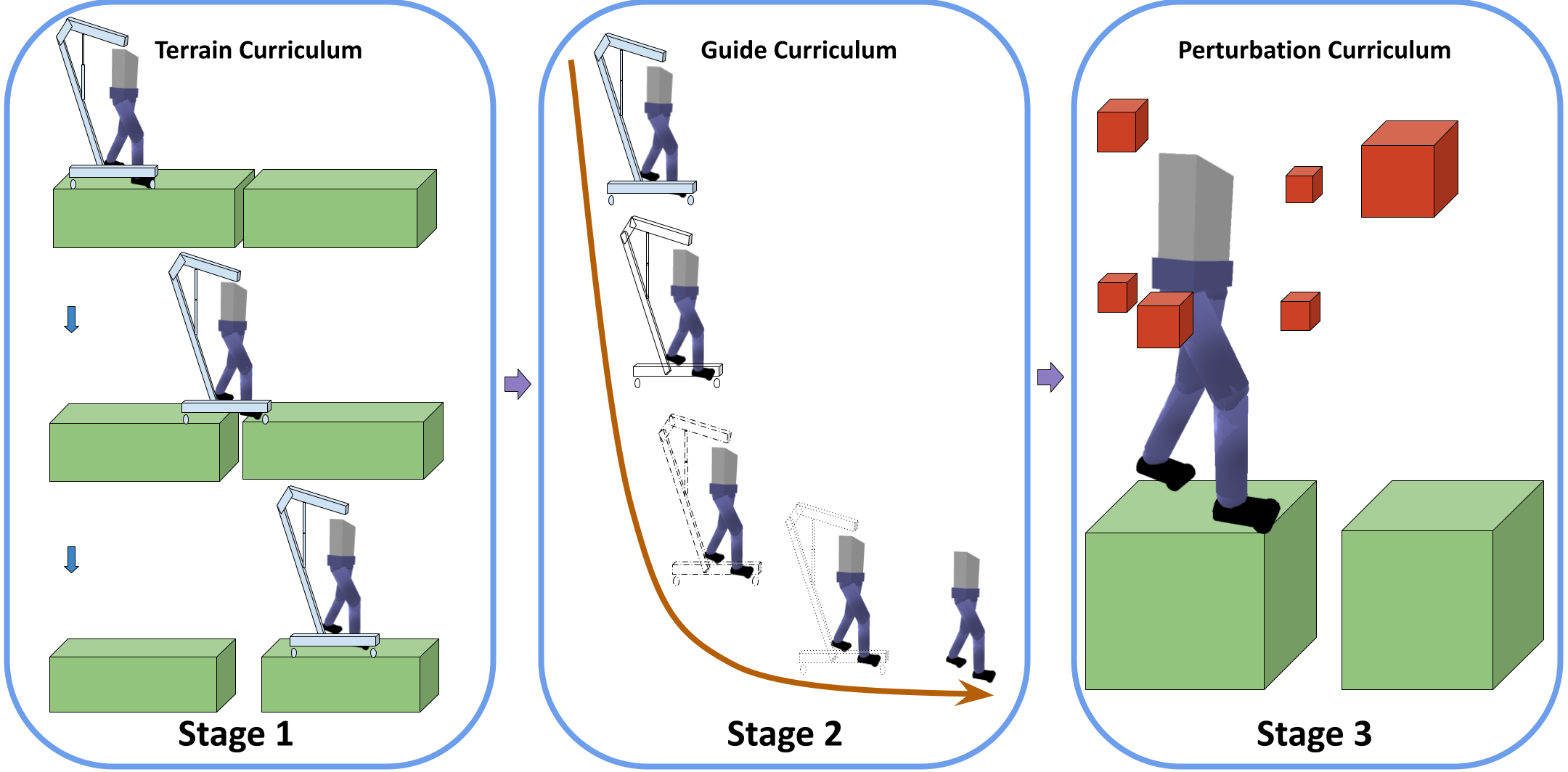}
\caption{A three stage curriculum learning method was developed for efficiently training bipedal behaviours for traversing difficult terrain types. Terrain difficulty was increased while guide forces were applied to the robot, then guide forces were reduced, and finally, perturbations increased.
}
\label{fig:cha3_fig1}
\end{figure*}

Complex behaviours that depend on perception, such as legged locomotion over challenging terrain, are difficult to design. Deep reinforcement learning (DRL) methods have been used to learn visuo-motor policies and are an alternative to labour intensive manual controller design. Training DRL policies typically require extensive interaction with the robot in the environment and commonly used exploration techniques are often not sufficient to guide the robot towards complex motions. Recent works utilise curriculum learning [\cite{yu_learning_2018}] and expert demonstrations [\cite{peng_deepmimic_2018}] to improve the learning efficiency of complex robots. This chapter explores these ideas to efficiently learn visuo-motor behaviours to enable bipedal locomotion over difficult terrain.

A three stage curriculum method was developed to improve the training efficiency of DRL policies. The curriculum learning method is shown in Figure \ref{fig:cha3_fig1}. In the first stage (\textbf{terrain curriculum}), forces are applied to the base and joints of the robot following a simple walking trajectory consisting of joint and body positions. The terrain difficult is gradually increased from an easy setting, to a more difficult level based on the success of the robot. In the second stage (\textbf{guide curriculum}), the guiding forces are gradually reduced to zero. Finally, in the third stage (\textbf{perturbation curriculum}), random perturbations with increasing magnitude are applied to the robot base to improve the robustness of the policies. The robot state, (joint positions, velocities, and end-effector contact), along with a depth image (48x48) are passed to a neural network policy that outputs the torque applied to each joint. 



\begin{figure}[h!tb]
\centering
\subfloat[Curved paths initial angle from the straight section starting at $20\degree$ and width $\SI{1.1}{\meter}$ ]{\includegraphics[width=\w cm, height=\h cm]{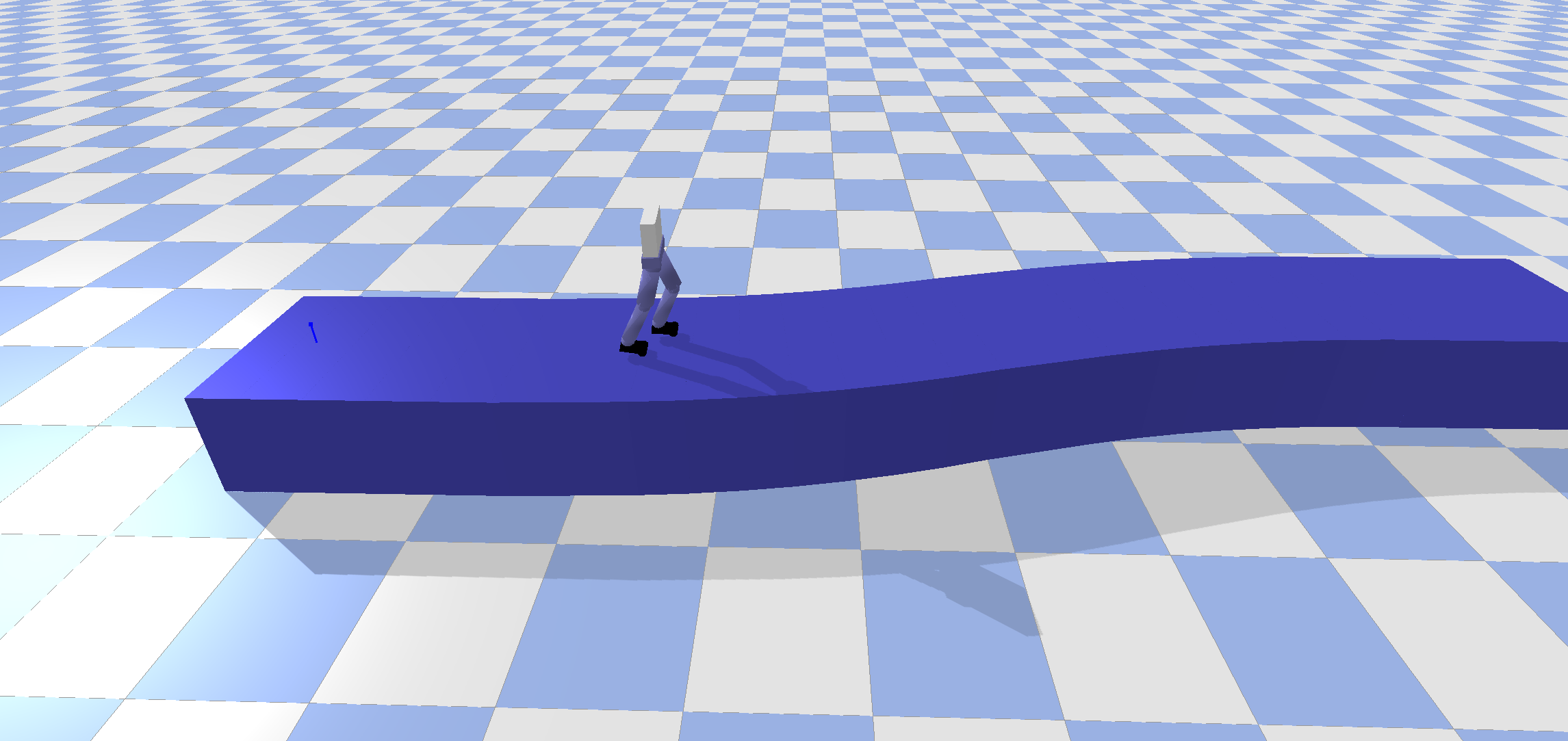}}
\hspace{0.5cm}
\subfloat[Curved path final angle from the straight section
$120\degree$ and width $\SI{0.9}{\meter}$]{\includegraphics[width=\w cm, height=\h cm]{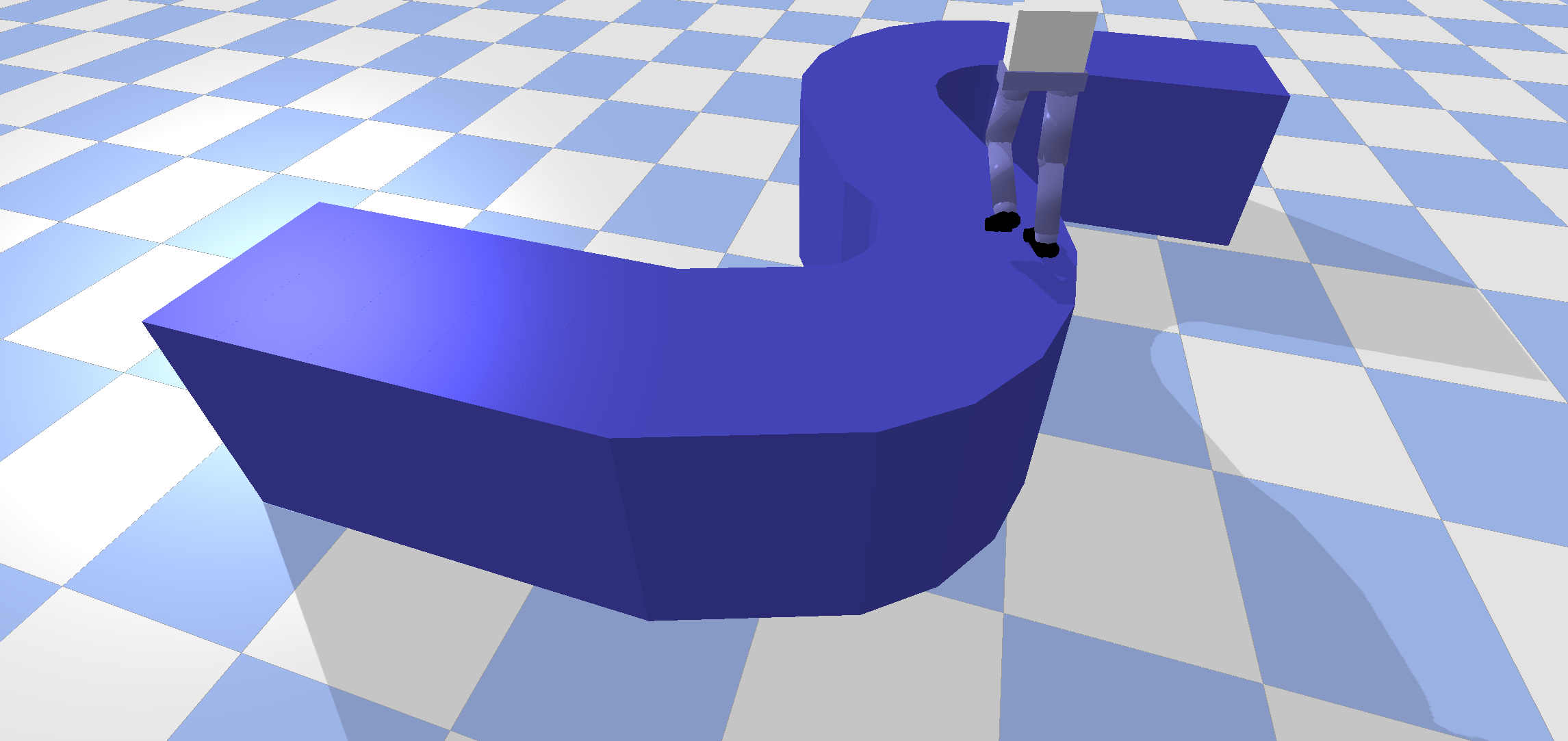}}
\hfill
\subfloat[Hurdles initial height $\SI{13}{\cm}$]{\includegraphics[width=\w cm, height=\h cm]{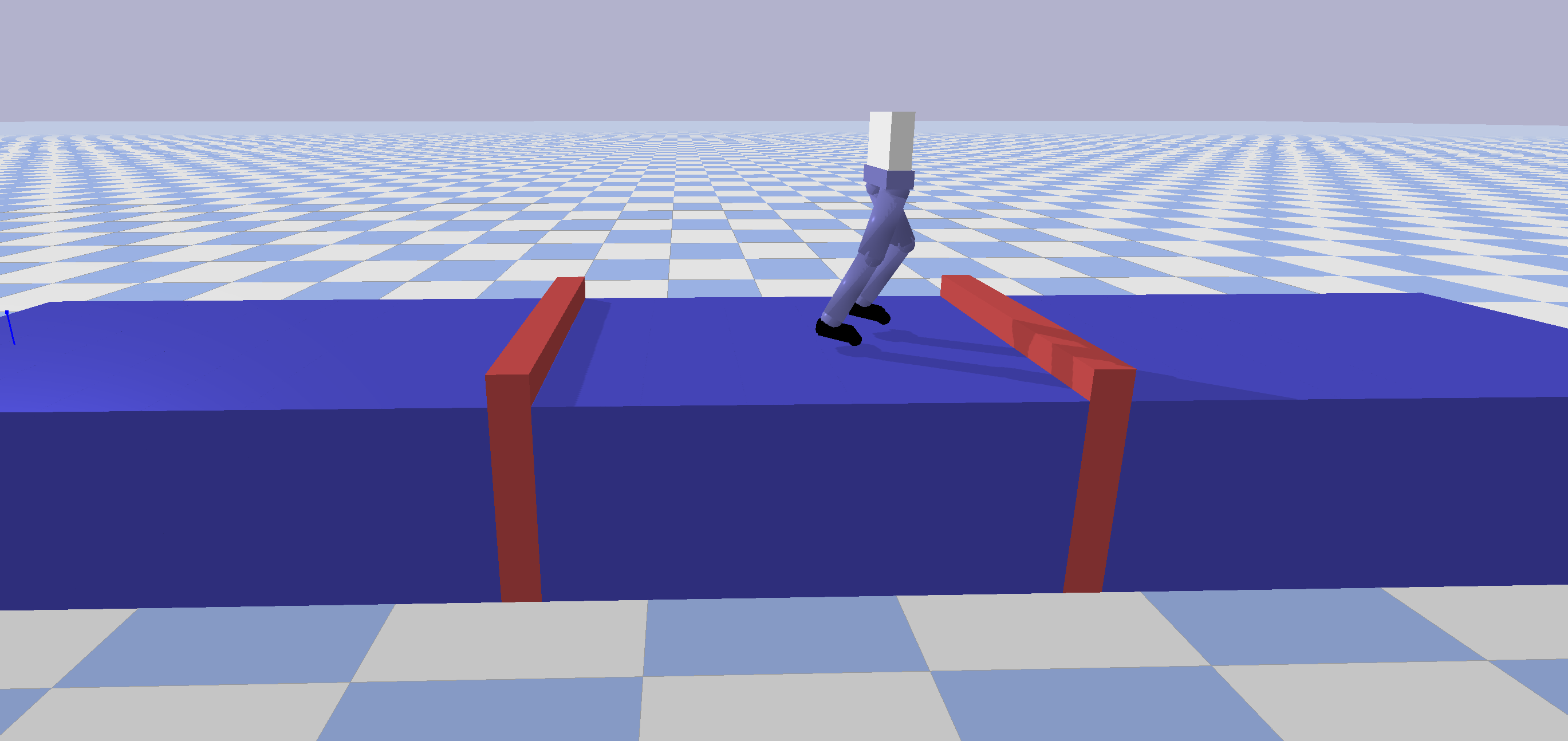}}
\hspace{0.5cm}
\subfloat[Hurdles final height $\SI{38}{\cm}$]{\includegraphics[width=\w cm, height=\h cm]{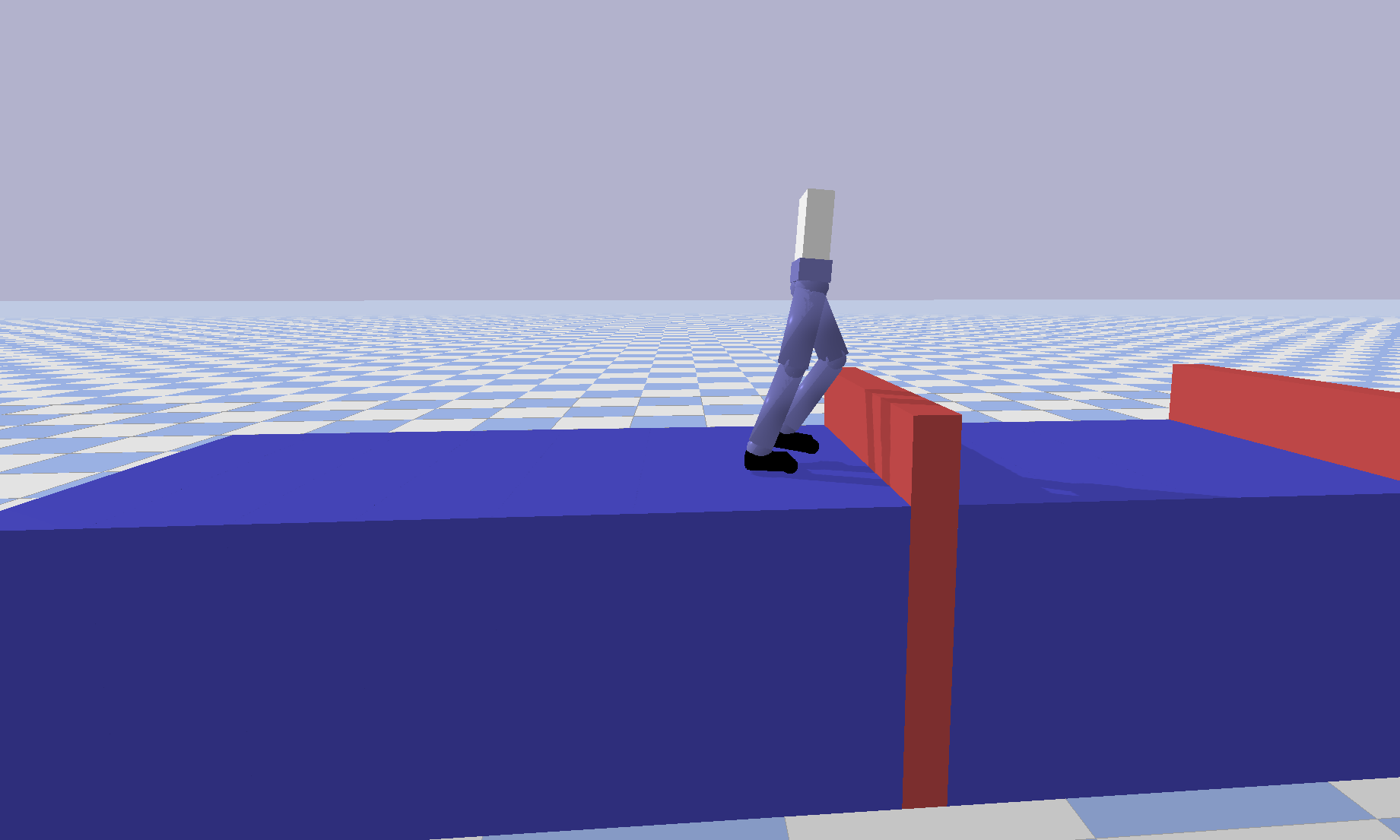}}
\hfill
\subfloat[Gaps initial length $\SI{10}{\cm}$]{\includegraphics[width=\w cm, height=\h cm]{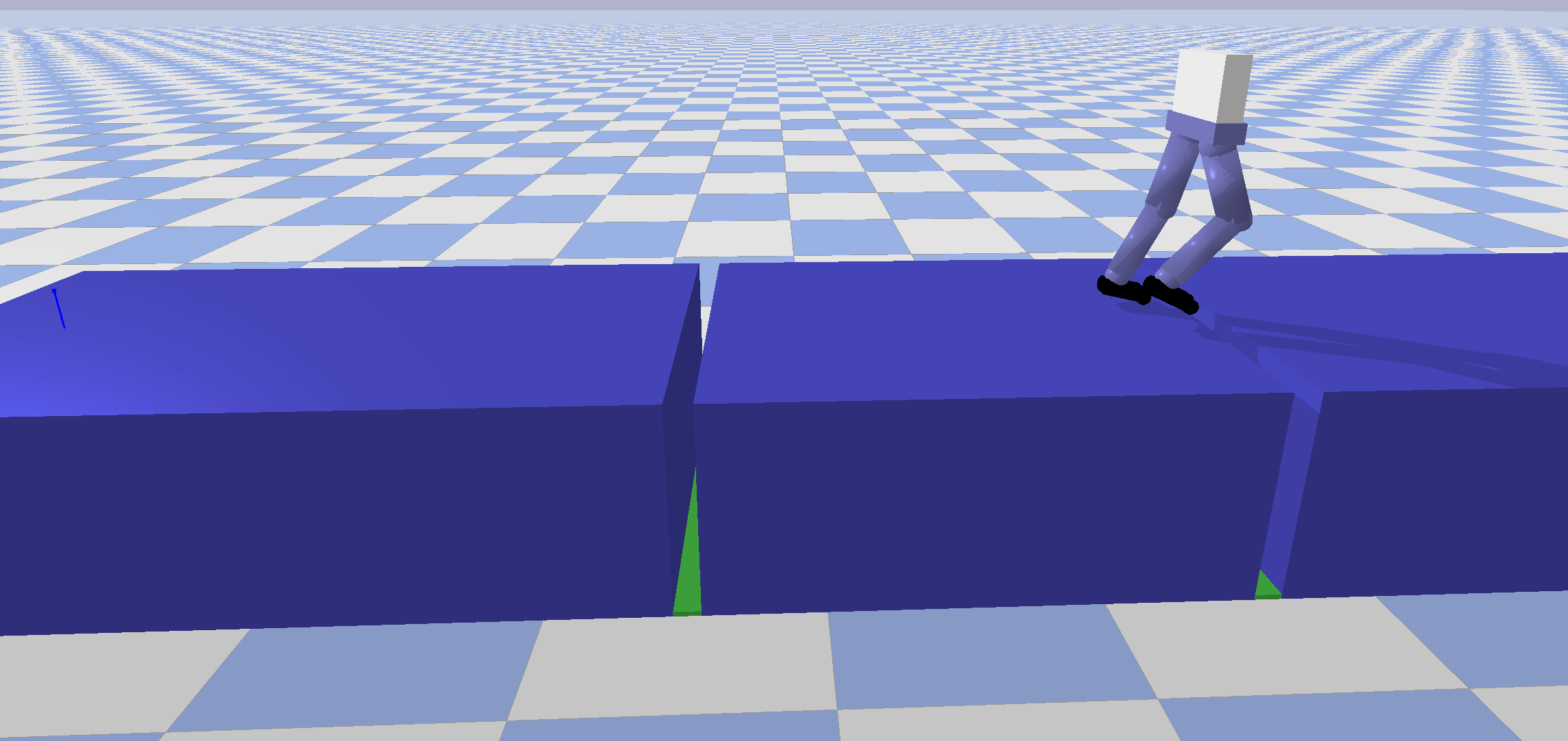}}
\hspace{0.5cm}
\subfloat[Gaps final length $\SI{100}{\cm}$]{\includegraphics[width=\w cm, height=\h cm]{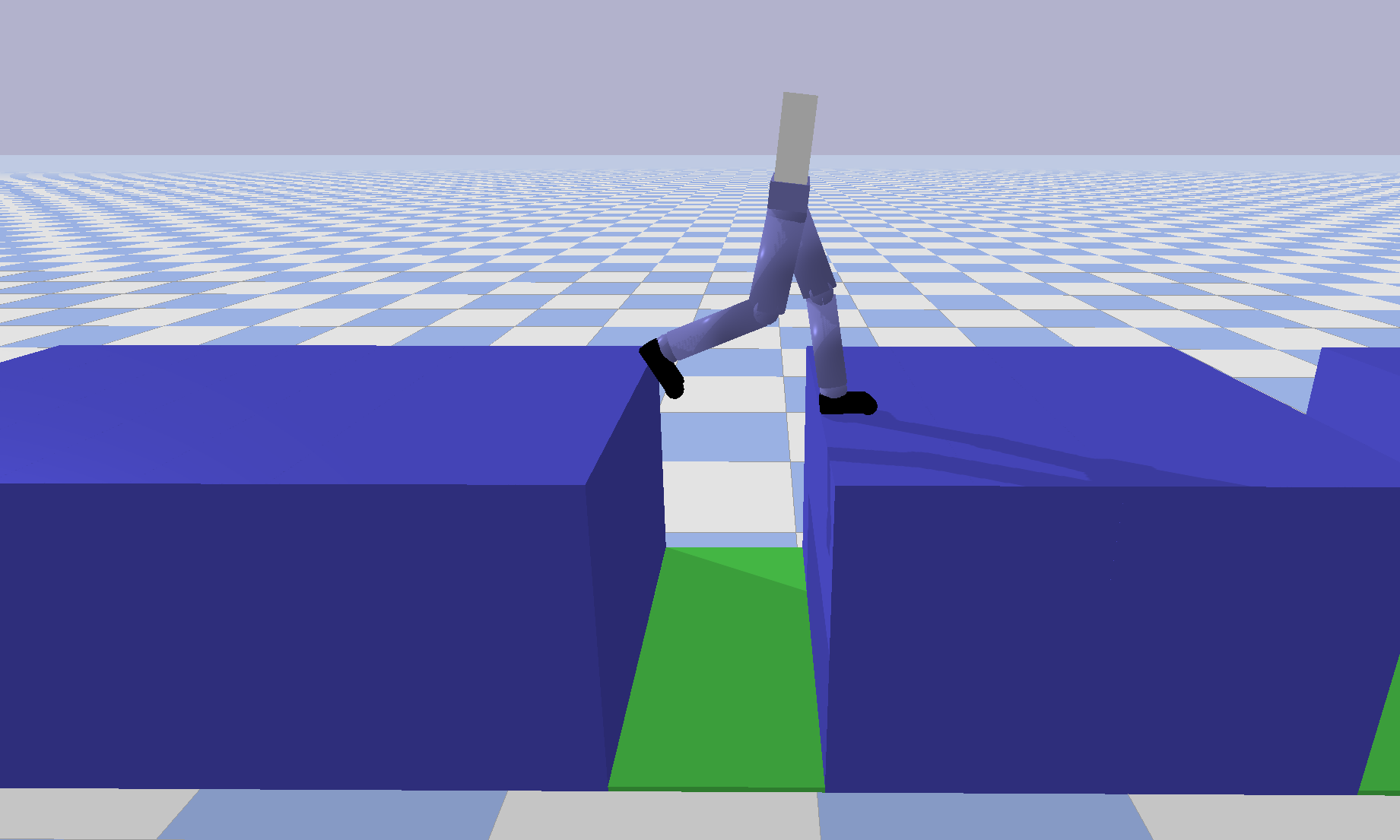}}
\hfill
\subfloat[Stairs initial height $\SI{1.7}{\cm}$]{\includegraphics[width=\w cm, height=\h cm]{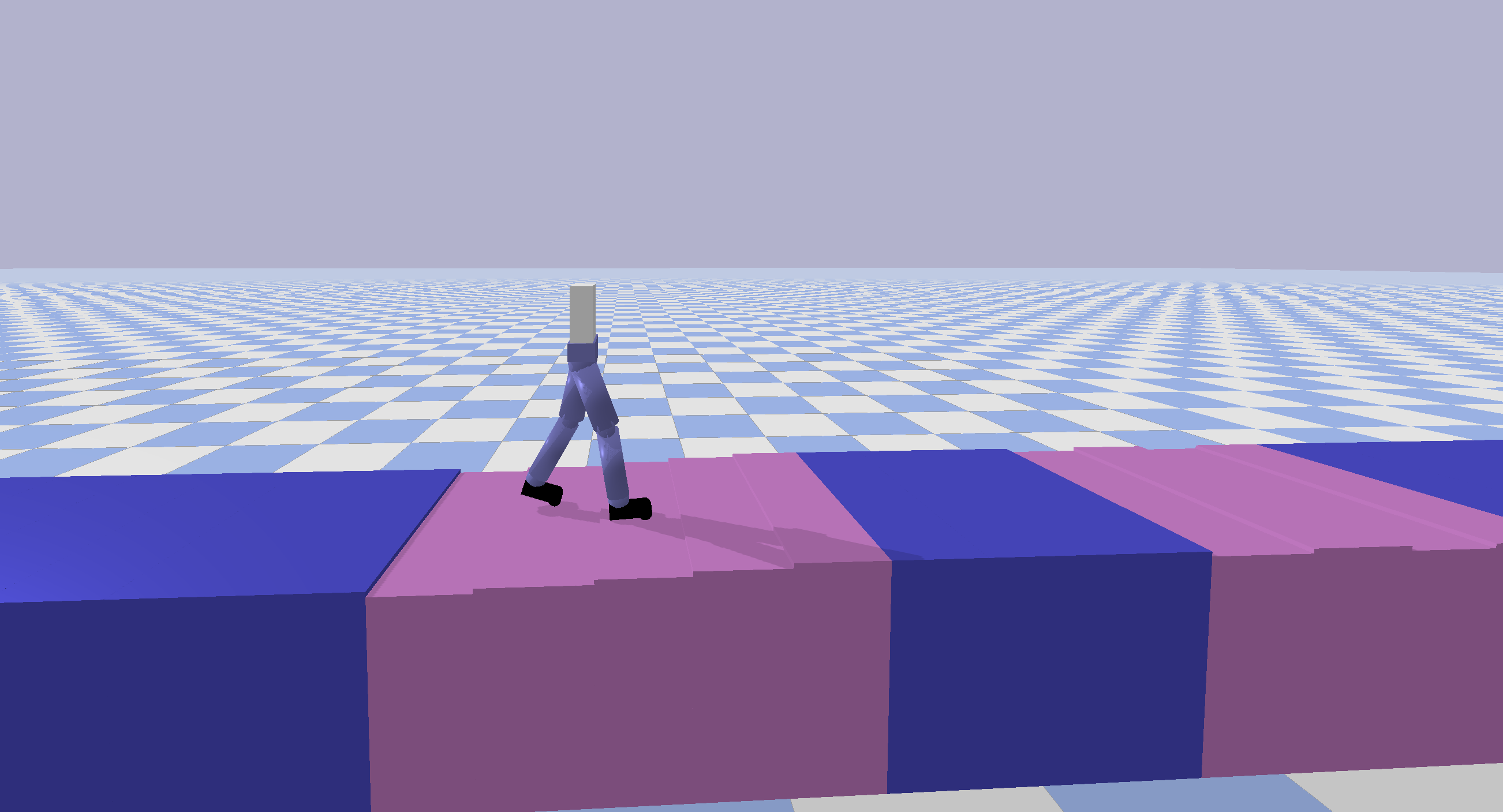}}
\hspace{0.5cm}
\subfloat[Stairs final height $\SI{17}{\cm}$]{\includegraphics[width=\w cm, height=\h cm]{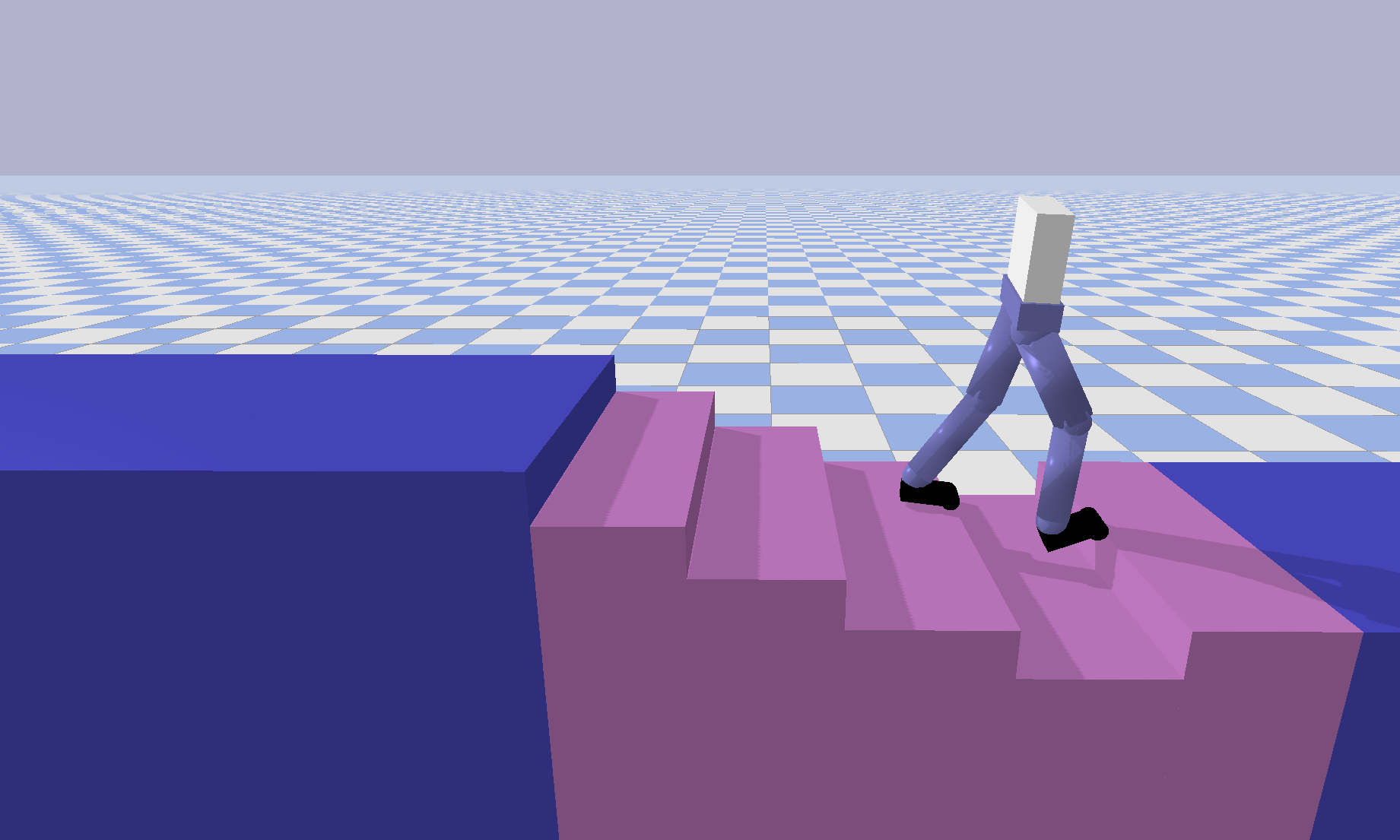}}
\hfill
\subfloat[Stepping stones initial distance apart $\SI{4}{\cm}$]{\includegraphics[width=\w cm, height=\h cm]{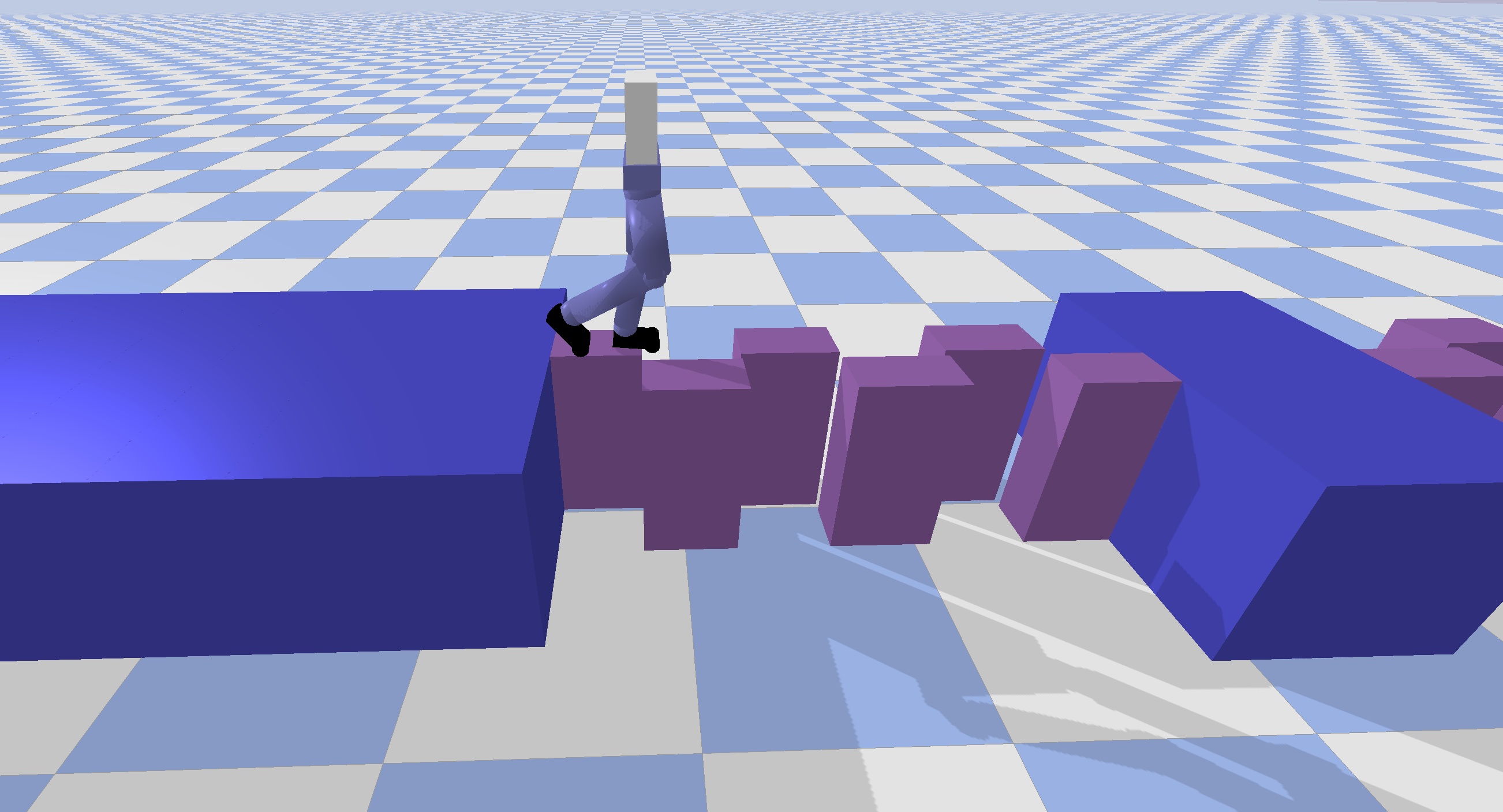}}
\hspace{0.5cm}
\subfloat[Stepping stones final distance apart $\SI{40}{\cm}$]{\includegraphics[width=\w cm, height=\h cm]{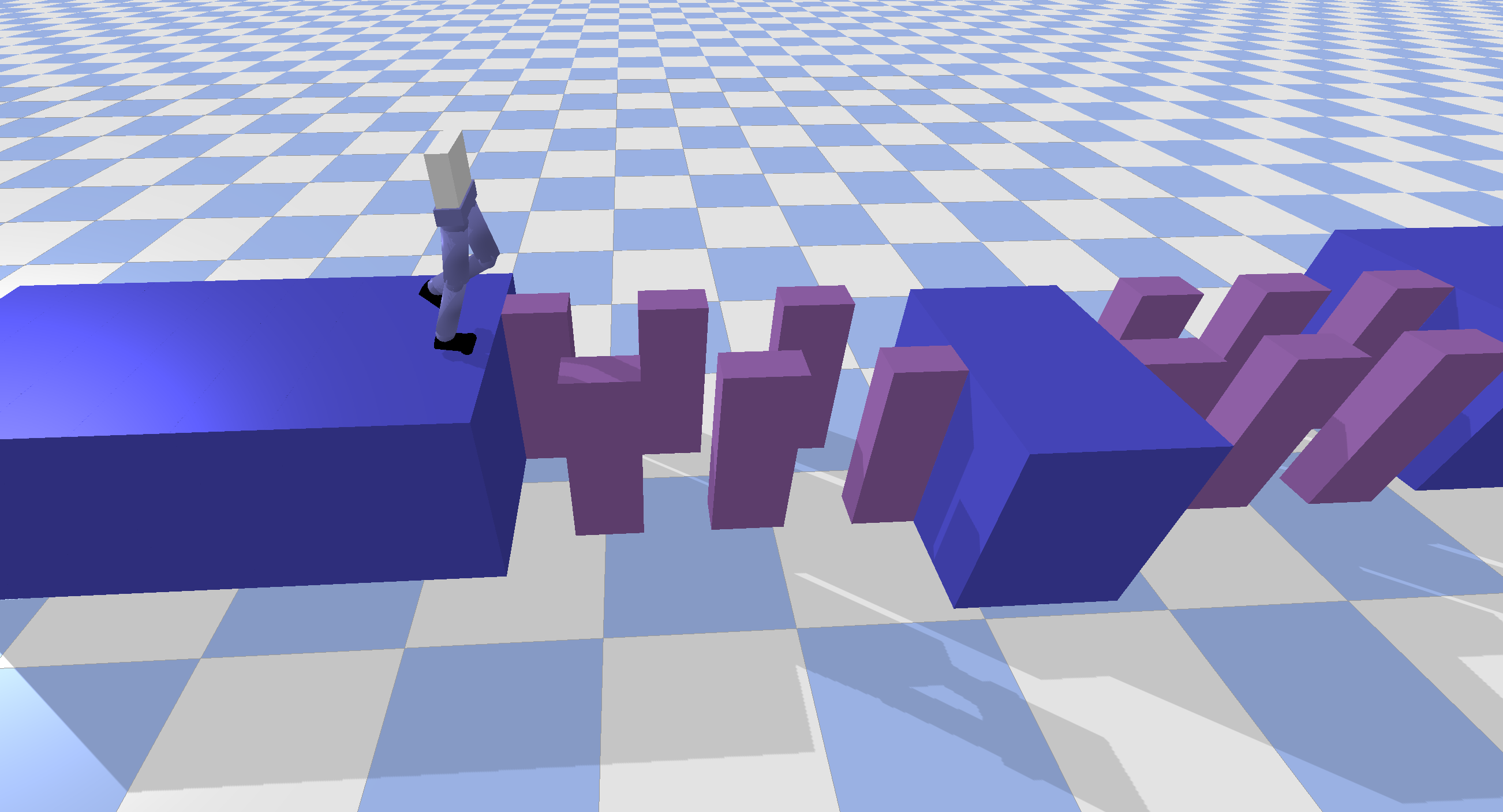}}
\caption{The initial and final terrain difficulties used in the first stage of training.}
\label{fig:cha3_fig3}
\end{figure}



\begin{table}[h!]
    \vspace{-2mm}
    \centering
    \begin{adjustbox}{max width=1.0\columnwidth}  
    \begin{tabular}{cccccc}
          & Curved & Gap & Hurdle & Stairs & Steps \\
         \hline
         No curriculum stages    & 79.4           & 12.8          & 10.3          & 12.4          & 13.4 \\
         Terrain curriculum only    & 5.5  & 5.3          & 7.1          &  11.5          & 14.2 \\
         Guide curriculum only          & 88.7           &  1.5          &  1.2          &  43.9          &   1.9 \\
         Terrain and guide curriculum     &  89.8            & 69.5           &  \textbf{79.0}          & 55.8 & 58.1   \\
         All three stages   &  \textbf{99.9}    & \textbf{72.3}  & 58.5  & \textbf{57.6}  & \textbf{60.5}  \\
    \hline
    \end{tabular}
    \end{adjustbox}
    \caption{Ablation study of the 3 curriculum stages (terrain, guide, perturbation) showing the percentage progress of the total terrain length (500 episodes of 7 artifacts in sequence).}
    \label{tab:ch3_results_ablation}
    \vspace{-5mm}
\end{table}


Simulation experiments verified the proposed learning approach, efficiently training separate behaviours for five difficult terrain types: curved paths, hurdles, gaps, stairs, and stepping stones. The terrains and their initial and final difficulties are shown in Figure \ref{fig:cha3_fig3}. Failure is indicated by the robot centre of gravity falling below a measured height. A simple hand-designed walking trajectory of joint and body positions was demonstrated to be a sufficient target to learn complex behaviours. Table \ref{tab:ch3_results_ablation} shows the results of an ablation study of the three curriculum stages. These results highlight the effect of each curriculum stage on terrain coverage by separately removing the terrain, guide, and perturbation curricula, the same as setting the hardest terrain difficulty, no guide forces, and smallest applied disturbances during training. The terrain and guide curricula provide the most improvement on terrain progress (\textbf{terrain and guide curriculum}), with improved robustness shown when perturbations were increased for all terrain types except hurdles (\textbf{all three stages}). 

Several components were identified as key to the success of this method.
\begin{itemize}
    \item Curriculum progression based on the success of the robot performed better than continuously increasing the difficulty of the scenario.
    \item Guide forces applied to the robot base and joints proved more effective than guide forces applied to the base only.
    \item The simple walking trajectory was linked to the robot during experience collection. This was done by progressing the walking trajectory as the relative swing foot of the agent made contact with the ground, improving learning, and removing the need for a phase variable required by other methods [\cite{peng_deepmimic_2018}].
    
\end{itemize}   

 
Guided curriculum learning (GCL) can be utilised for learning visuo-motor behaviours in tasks where the difficulty can be increased throughout training, and a simple trajectory can be obtained that guides the robot part of the way to the solution. For the terrains investigated, the robot quickly learned to traverse the terrain, progressing through all terrain difficulties within the first 10\% of training. Learning efficiency could be improved further by exploring hierarchical learning structures [\cite{levy_learning_2019}], and with condensed state and action representations [\cite{da_supervised_2017}]. Suggestions from this work would be to advance the target trajectory with the progression of the learning agent, increase the difficulty level with task success, and ensure the task can be solved with guide forces early in training. This method could be used for other locomotion tasks, for example jumping, or extended to grasping and other control tasks.

\clearpage

\includepdf[pages=-,pagecommand={},width=\textwidth]{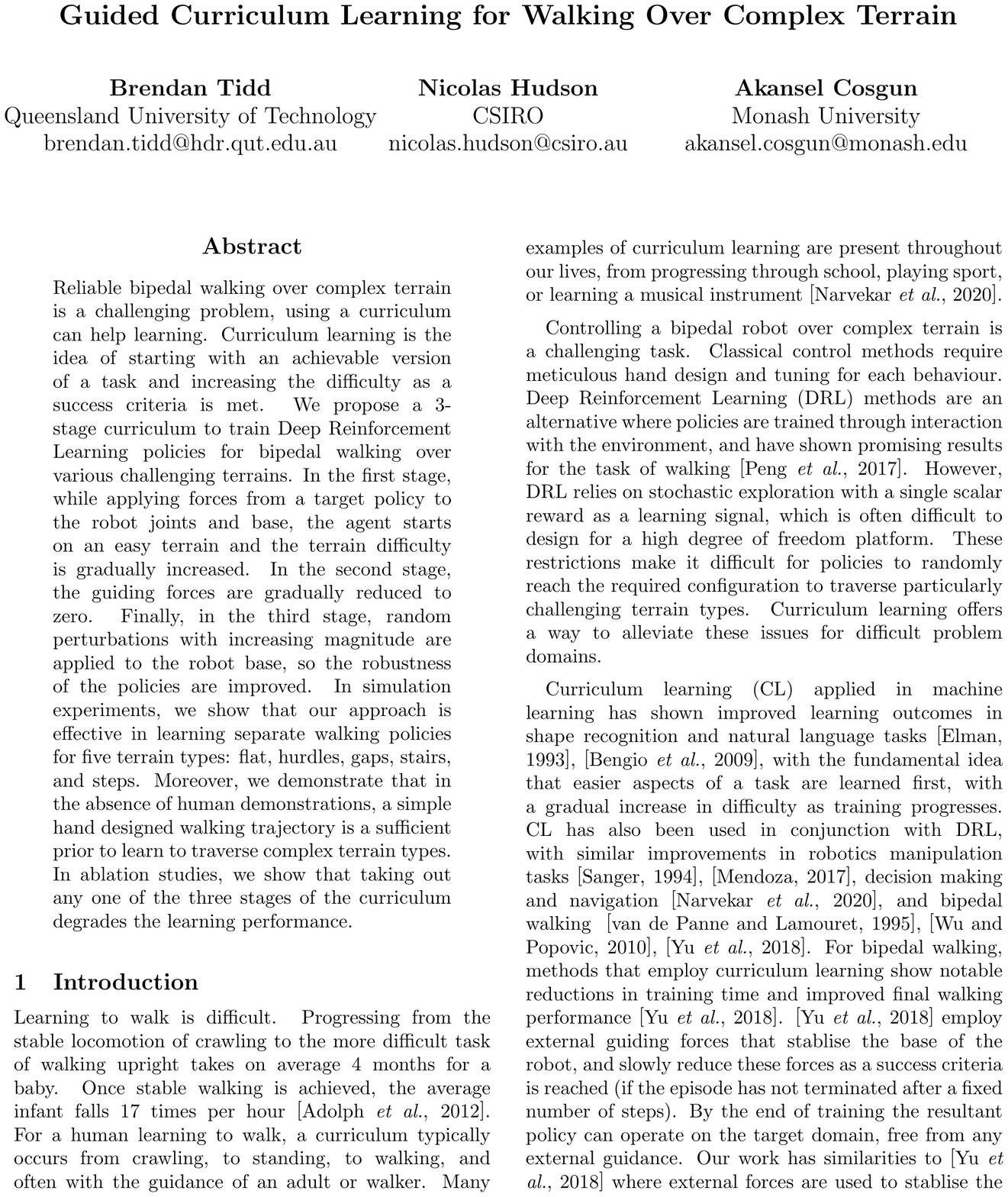}

\chapter[Passing Through Narrow Gaps]{Passing Through Narrow Gaps with Deep Reinforcement Learning}

\label{cha:ch3b}

\includepdf[pages=-,pagecommand={},width=\textwidth]{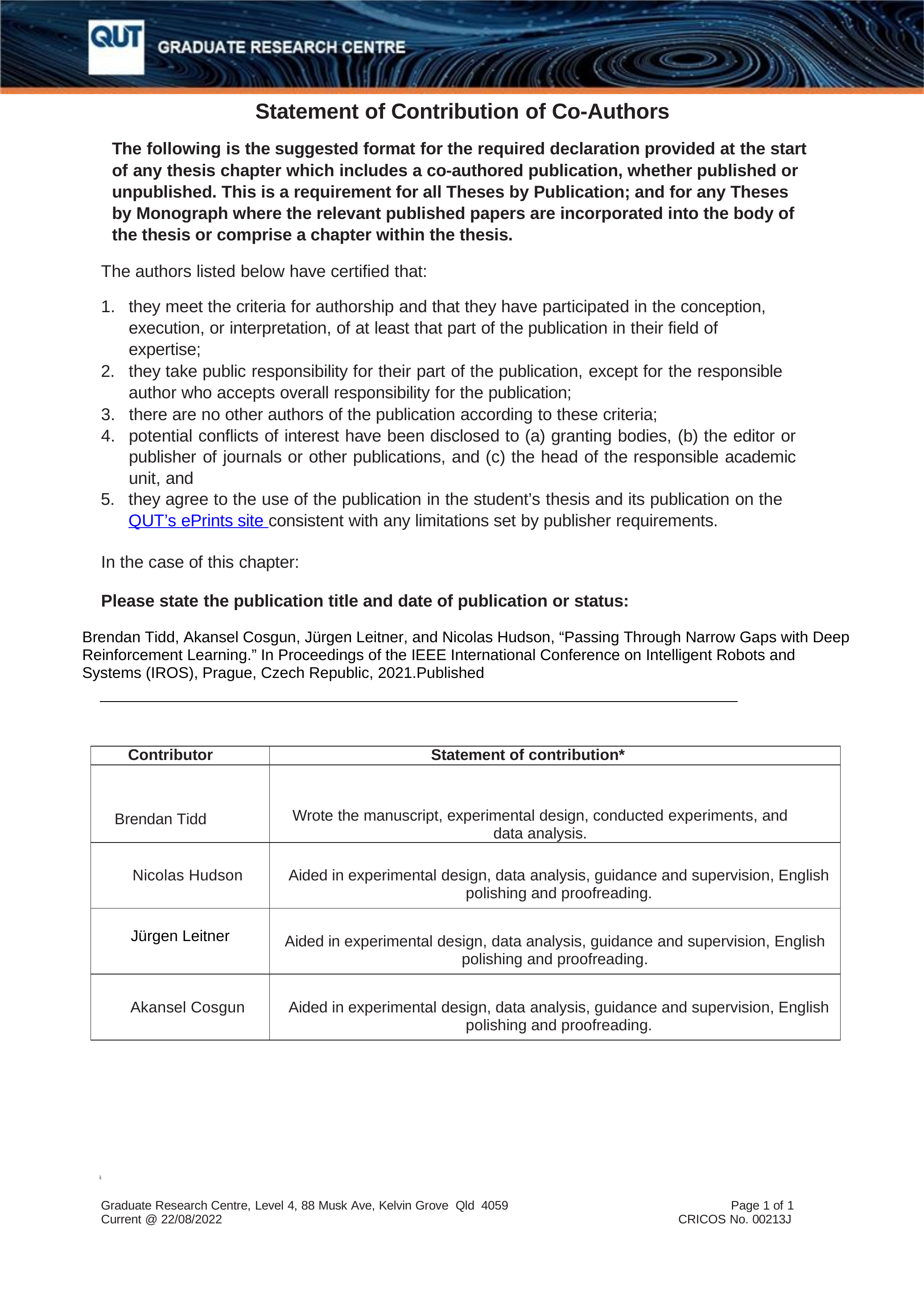}


This chapter furthers explores research question 1: \textbf{\textit{how can complex visuo-motor locomotion behaviours be learned efficiently,}} and introduces the first investigation of research question 2: \textbf{\textit{how can a safe switch state be determined to facilitate the reliable switching of behaviours?}} The challenges of developing a complex navigation behaviour are explored, including the design of a gap traversal behaviour, integration with existing behaviours, and the investigation of real-world transfer of policies. The contributions of this chapter are as follows:

\begin{itemize}
    \item A gap behaviour policy was developed to enable a large mobile robot to navigate through narrow doorways marginally wider than the robot. The gap behaviour was trained efficiently using a simple waypoint controller and reward shaping. 
    

    \item A behaviour selection policy demonstrated safe switching of behaviours by learning when to autonomously activate the gap behaviour and when to use a set of traditional controllers. The effectiveness of the these behaviours was demonstrated in simulation and a real-world scenario, with several sources of the sim-to-real disparity identified.
\end{itemize}

“Passing Through Narrow Gaps with Deep Reinforcement Learning” was published and presented at the 2021 IEEE/RSJ International Conference on Intelligent Robots and Systems held in Prague, Czech Republic.
\newpage
\newpage

\begin{figure}[htb!]
    \captionsetup[subfigure]{labelformat=empty}
    \centering 
    \subfloat[]{\includegraphics[trim=1cm 1cm 1cm 2cm, clip,width=0.38\columnwidth]{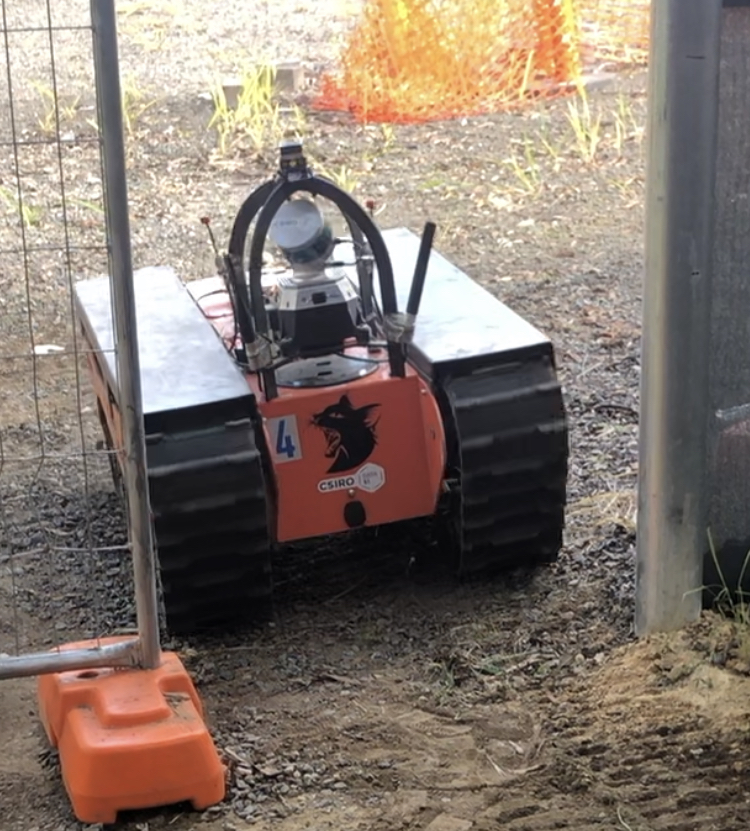}}
    \hfil
    \subfloat[]{\includegraphics[trim=5.0cm 0.0cm 0.2cm 15.3cm, clip, width=0.42\columnwidth]{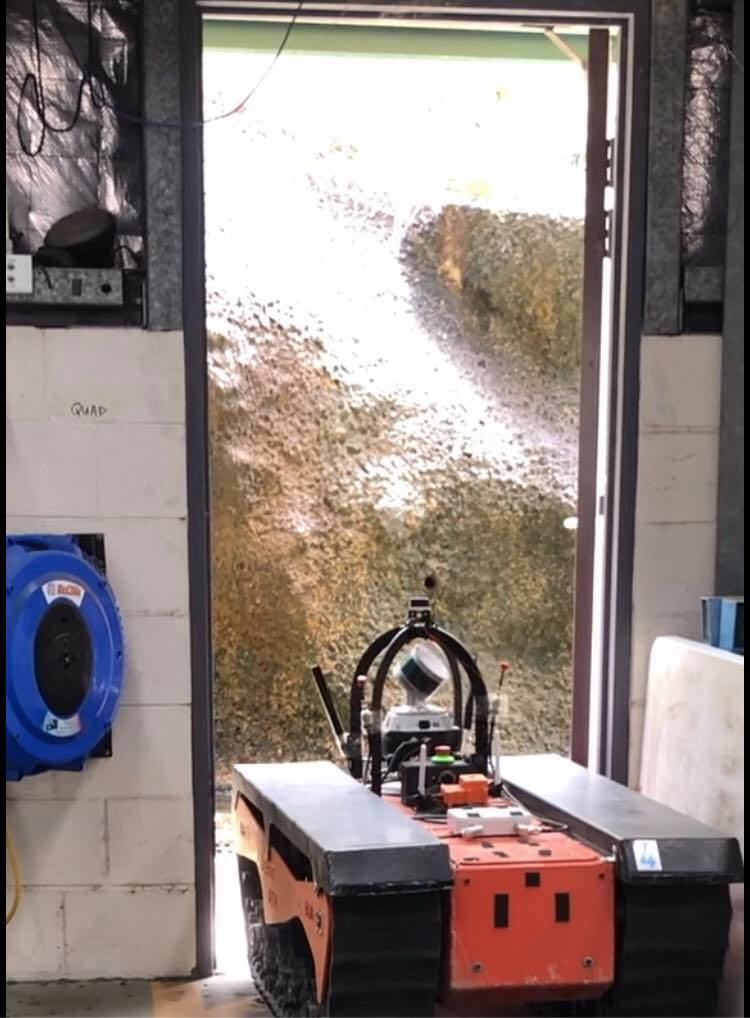}}
    \hfil
    \caption{Traversing narrow gaps with a large mobile robot is a challenging problem, particularly when  perceiving through low resolution costmaps.}
    \label{fig:cha5_fig1}
\end{figure}

The DARPA Subterranean Challenge (SubT) required teams of robots to traverse difficult and diverse underground environments. Traversing narrow gaps, for example at a tunnel entrance or a doorway (Figure \ref{fig:cha5_fig1}), was one of the challenging scenarios robots encountered. Path planning methods fail when gaps that are perceived as smaller than the robot footprint are in fact traversable, regardless of the choice of planning algorithm employed. Complex behaviours that are required in these situations, for example making contact with the environment, can be difficult to design manually. DRL is a suitable alternative to traditional control design, however, designing behaviours efficiently, and integrating learned behaviours with an existing set of controllers is challenging. These problems are investigated in this chapter.

\begin{figure}[t!]
    \centering
    \includegraphics[width=0.85\columnwidth]{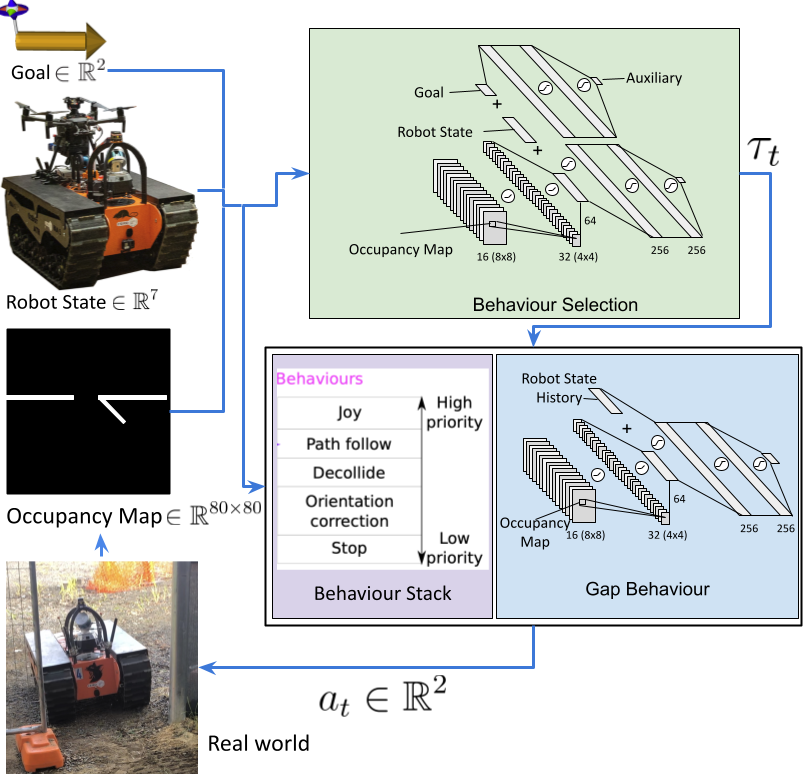}
    \caption{The behaviour selection policy determines when the gap behaviour should be active or when the robot should utilise the behaviour stack (set of existing controllers). An occupancy map and robot state are provided as input, and forward and heading velocities command the robot.}
    \label{fig:cha5_fig2}
\end{figure}

A gap behaviour policy was developed to control the robot through small gaps (only centimeters wider than the robot) using an occupancy map with a resolution of 10 cm. Efficient learning was demonstrated by training policies from a simple waypoint controller using reward shaping. Importantly, the simple controller was not able to complete the task, and was used to bootstrap learning the complex behaviour. A goal-conditioned behaviour selection policy was then trained to determine when to activate the gap behaviour. An occupancy map and the angular velocity of the robot was provided to each policy, and forward and heading velocity were used to to command the robot. A goal location was provided to the behaviour selection policy that determined where the robot should move to. If the goal location was through a small gap the gap behaviour was activated, otherwise a behaviour stack (set of existing behaviours [\cite{hines_virtual_2021}]) was used. The policies trained in simulation were demonstrated on the real platform. Figure \ref{fig:cha5_fig2} shows the pipeline that was applied in the real-world scenario.

The robot is able to move through traversable gaps perceived as smaller than the robot base. Traditional planning approaches require higher resolution perception information to solve this problem, whereas the learned approach determines the most likely position of the true edge of the gap by the filling and unfilling of cells in the occupancy map, and through interaction with the environment (making contact with the walls). In simulation experiments, a success rate of 93\% was achieved when the gap behaviour was activated manually by an operator, and 63\% with autonomous activation using the behaviour selection policy (Table \ref{tab:ch5_sim_results}). In real-world experiments, a drop in performance was observed, with a success rate of 73\% with manual activation, and 40\% with autonomous behaviour selection (Table \ref{tab:ch5_real_results}). In both the simulated and real-world environments, methods that rely on planning a path with the low resolution costmap while considering the kinematic modelling of the robot were unable to traverse the gap [\cite{hudson_heterogeneous_2021}].

While the feasibility of this method was demonstrated in simulation, the difference in performance between simulated and real-world scenarios highlights the difficulty of direct sim-to-real transfer for complex DRL policies. Policy training was performed with ground truth perception, however, a decrease in performance was shown when evaluated in the same simulation with a SLAM system for occupancy map generation. This indicates a discrepancy between the ground truth and generated occupancy maps. A common failure case occurred in real-world experiments where the robot would catch the doorway causing the tracks to slide on the ground. This behaviour was not experienced during training, inclusion of this scenario at training time could reduce the occurrence of this failure case. Factors such as imperfect perception on the real agent, inaccuracies in simulation dynamics, and out of distribution examples seen during deployment contributed to the drop in performance in the real world. 

\begin{table}[h!]
    \centering
    \begin{tabular}{cccccc}
                                      & Success \% & Time(s) & Operator Actions \\
       \hline
        Behaviour Stack      & 0.0            &  50.2 &    1  \\
        Manual Gap Behaviour Selection        & 93.3          &  29.7 &    3  \\
        Auto Gap Behaviour Selection  & 63.3          &  45.3 &    1 \\
    \hline
    \end{tabular}
    \caption{Simulation results for passing through a narrow gap (30 trials).}
    \label{tab:ch5_sim_results}
\end{table}

\begin{table}[h!]
    \centering
    \begin{tabular}{cccccc}
                                      & Success \% & Time(s) & Operator Actions \\
       \hline
        Behaviour Stack              & 0.0           &  44.6 &    1  \\
        Manual Gap Behaviour Selection               & 73.3          &  24.1 &    3  \\
        Auto Gap Behaviour Selection         & 40.0          &  30.9 &    1 \\
    \hline
    \end{tabular}
    \caption{Real robot results for passing through a narrow gap (15 trials).}
    \label{tab:ch5_real_results}
\end{table}

The methods developed in this chapter can be applied to other tasks where learning efficiency can be improved by providing a simple trajectory to guide learning. Learning when to activate the task behaviour requires a baseline controller for moving the robot towards the task, and the switch states to be defined during training. For example, in the finals of the SubT Challenge, many robots from Team CSIRO Data61 were immobilised as a result of train tracks, where turning on the tracks resulted in damage to the robot (tracked) or caused the robot to slip and fall (legged). A waypoint could be provided during training to guide the robot away from dangerous maneuvers over train tracks. Sim-to-real transfer requires the simulation and real world perception and robot dynamics to closely match, and many scenarios with variation are required during training to ensure the real distribution of terrains are experienced.


\clearpage 

\includepdf[pages=-,pagecommand={},width=\textwidth]{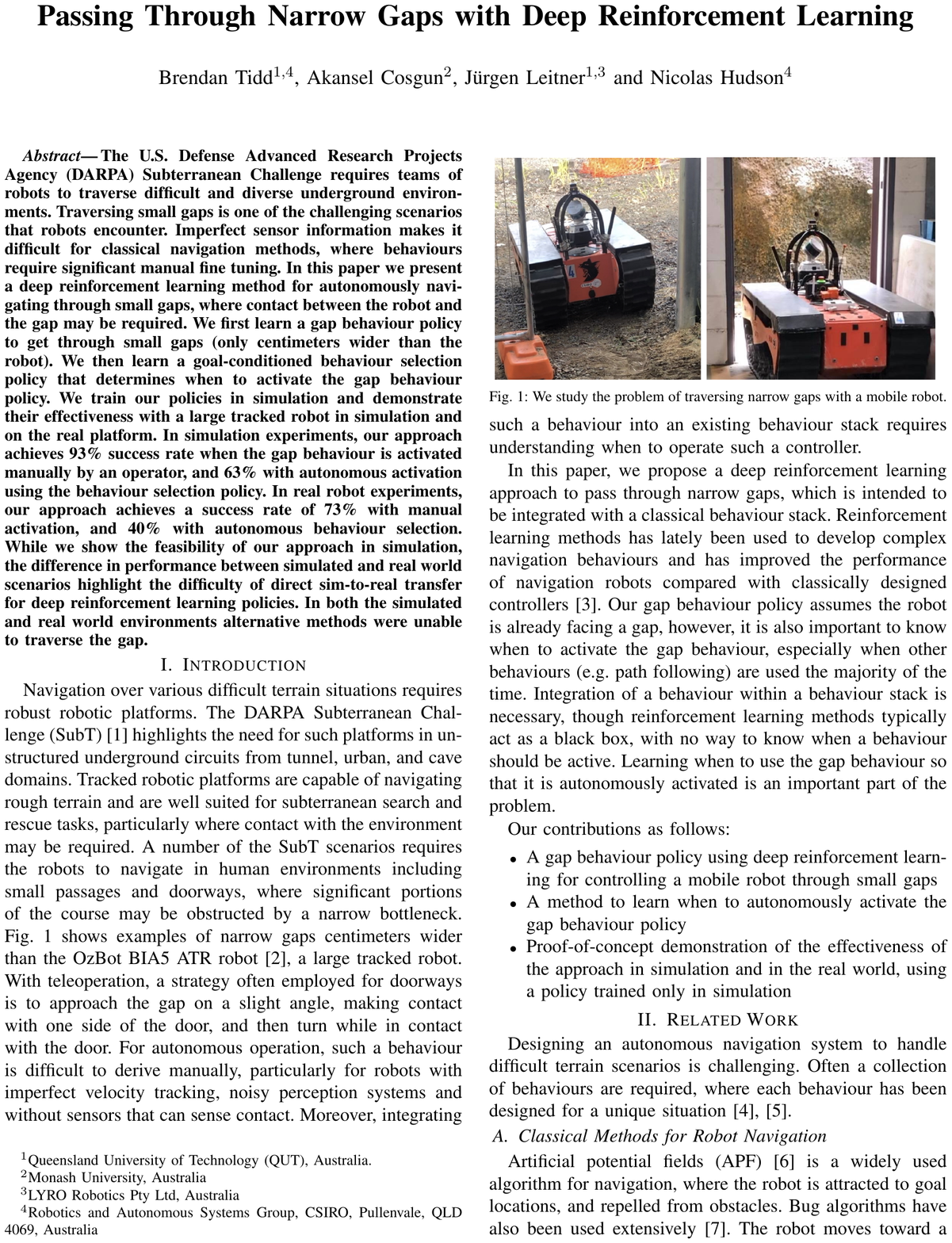}

\chapter[Learning When to Switch]{Learning When to Switch: Composing Controllers to Traverse a Sequence of Terrain Artifacts}

\label{cha:ch4}

\includepdf[pages=-,pagecommand={},width=\textwidth]{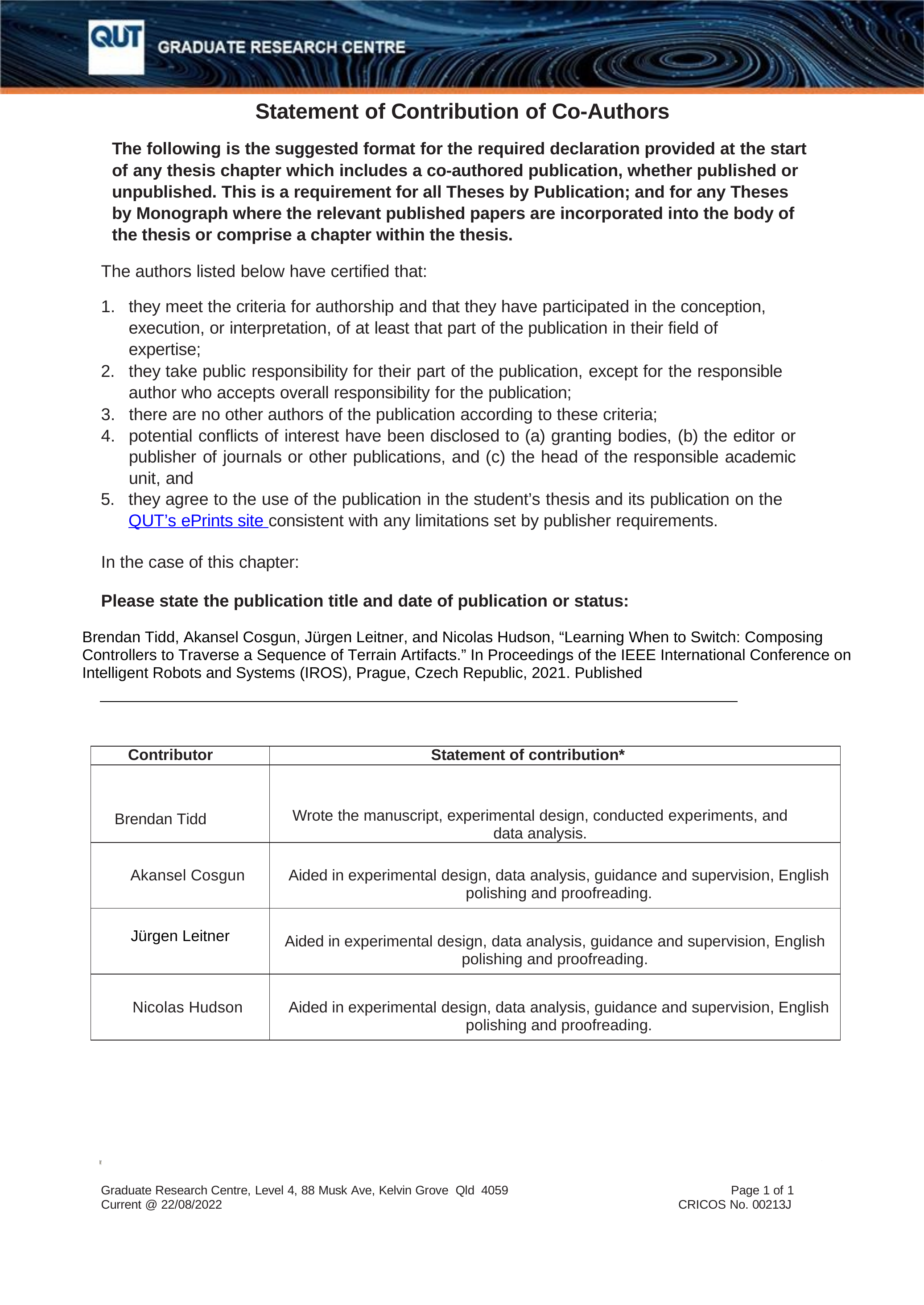}



Chapter \ref{cha:ch3a} discussed methods for efficiently learning separate locomotion behaviours for a simulated biped. This chapter considers combining behaviours to traverse a sequence of obstacles, continuing the investigation of research question 2: \textbf{\textit{how can a safe switch state be determined to facilitate the reliable switching of sequential behaviours?}} When composing complex controllers, it is important to understand which controller is appropriate for the upcoming terrain conditions, and when the selected controller can be safely activated. This chapter focuses on the latter problem, controller selection is provided by an oracle terrain detector. Understanding region of attraction (RoA) overlap can be difficult for complex controllers. This work estimates states within this overlap by learning the likelihood of success for switching between a set of learned behaviours. Typically, hierarchical methods that use control primitives require extensive retraining as new behaviours become available. In this work, retraining was minimised using supervised learning for estimating when to switch. The contributions of this chapter, from \cite{tidd_learning_2021} are as follows:

\begin{itemize}

    \item The effect of state overlap on safe policy switching was investigated by training diverse policies from common initial conditions. For biped experiments, the requirement of each policy to stand the robot in place for a short duration provided sufficient RoA overlap for reliable behaviour switching.
    \item A policy switching network was designed to estimate the likelihood of success in a given state, improving the reliability of switching behaviours, and enabling the robot to traverse a sequence of obstacles. 
\end{itemize}

“Learning When to Switch: Composing Controllers to Traverse a Sequence of Terrain Artifacts” was
published and presented at the 2021 IEEE/RSJ International Conference on Intelligent Robots and Systems held in Prague, Czech Republic. 

Dynamic platforms such as bipeds rely on controllers that consider the stability of the robot during operation. Switching from one behaviour to the next must occur when stability is ensured by the receiving controller. RoA overlap between two behaviours refers to a set of states, from which a designated behaviour will result if either controller is active. To safely switch, sequential behaviours must have a RoA overlap. The aim of this chapter is to traverse a sequence of obstacles (stairs, gaps, and hurdles), using separate behaviours. To determine when the robot was in a suitable switch state, a novel switch estimator policy was trained to predict the likelihood of successfully traversing the upcoming terrain from the current state. Additionally, the criticality of overlapping states was investigated by comparing the effect of initial conditions on the reliability of switching.  


\begin{figure}[t!]
\centering
\includegraphics[width=0.75\columnwidth]{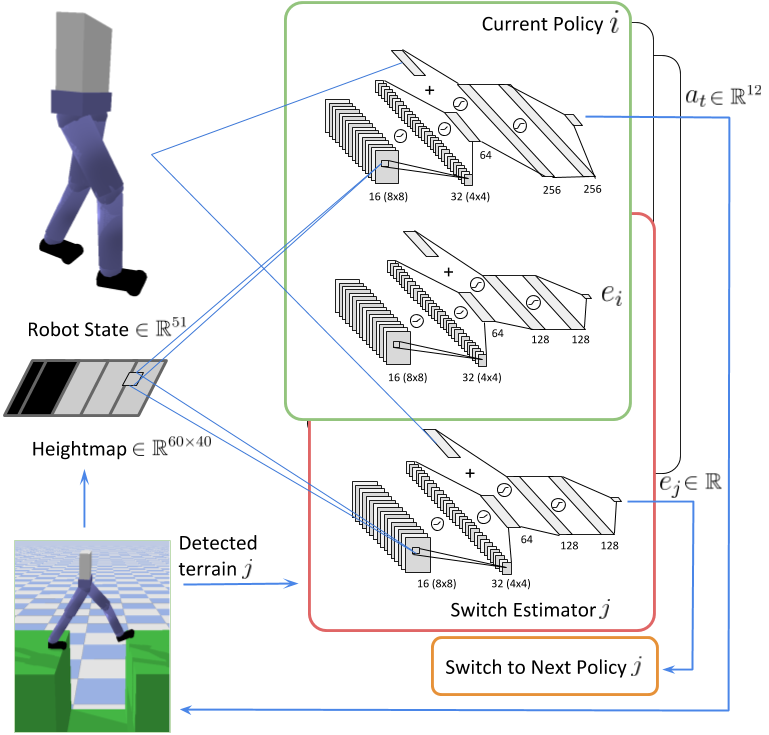}
\caption{A switch estimator improved the reliability of changing behaviours after a new terrain was detected. Estimate $e_j$ is the likelihood of success for behaviour $j$ if the switch occurred in the current state, given by the robot state and heightmap. This figure shows the transition from behaviour $i$ to behaviour $j$}
\label{fig:cha4_fig2}
\end{figure}

\begin{figure}[t!]
\centering
\includegraphics[width=0.9\columnwidth]{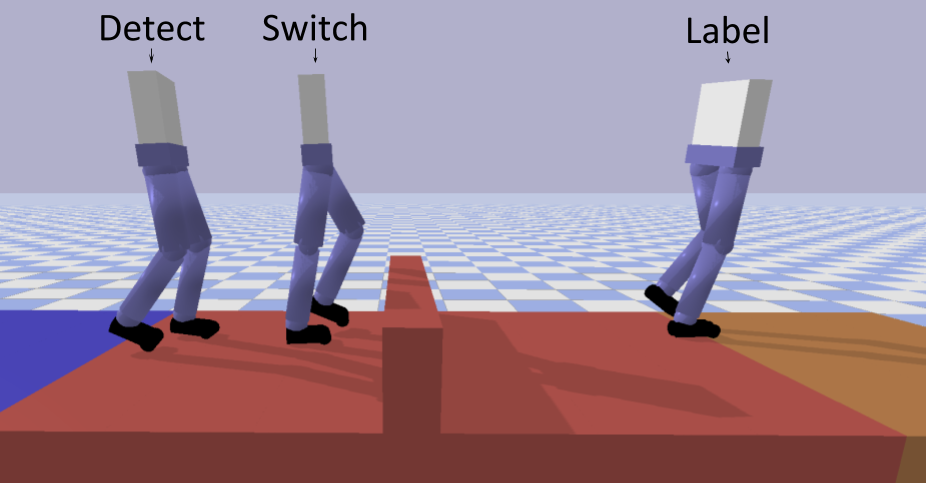}
\hfill
\caption{Each switch estimator was trained with data collected from the respective terrain.}
\label{fig:cha4_fig3}
\end{figure}

The pipeline for switching behaviours is outlined in Figure \ref{fig:cha4_fig2}. Each terrain type has a policy (trained with curriculum learning from Chapter \ref{cha:ch3a}), and a switch estimator introduced in this chapter. To ensure RoA overlap, each policy was trained to first stand the robot in place for a short duration before commencing the specific behaviour. When a new terrain was identified by a terrain oracle, the associated switch estimator determined the likelihood of success for switching to the appropriate behaviour for the upcoming terrain. A switch estimator was trained for each terrain type, using supervised learning, by collecting switch samples from each respective terrain type. During data collection the robot was controlled by a policy different to the terrain behaviour, and switching occurred at a random interval after the terrain was detected. The state the switch occurred and the label corresponding to the success of the terrain traversal was recorded and used to train the switch estimator policy (Figure \ref{fig:cha4_fig3}).

The results presented in Table \ref{tab:ch4a_test_results} show the total distance covered on a sequence of randomly selected obstacles, and the success rate (\% success) of passing the final terrain in the sequence. Several methods were compared, including switching a \textbf{random} distance from the terrain, \textbf{on detection} of the terrain, according to a \textbf{lookup table}, and when the centre of mass (CoM) of the robot was over the support of the stance foot (\textbf{CoM over feet}). The switch estimator method performed the most reliably on the sequence of obstacles. To verify the effect of RoA overlap, behaviours were trained by commencing each episode with random joint positions, instead of standing in place for a short duration. While the behaviours initialised from random joint positions performed well on individual obstacles, results show very poor performance when behaviour switching was required. This demonstrates that training policies from a standing initial configuration provided sufficient RoA overlap for the explored policies, and predicting when to switch using switch estimator policies improved the reliability of switching of behaviours.

\begin{table}[htb!]
    \centering

    \begin{tabular}{ccc}
         Switch Method & \% Total Dist. & \% Success \\   
         \hline
         Random  & 42.7 & 10.1 \\
         On detection  &   75.6 & 60.1 \\
         Lookup table &   76.3  & 59.0 \\
         CoM over feet  & 79.1 & 66.8 \\
         Switch Estimator & \textbf{82.4} & \textbf{71.4}\\   
         Switch Estimator no overlap& 17.5 & 0.7 \\   

    \end{tabular}

    \caption{Success rate and average distance travelled of the total terrain length of various switching methods traversing a sequence of obstacles.}
    \label{tab:ch4a_test_results}
\end{table}

The ideas introduced in this chapter can be applied to scenarios where sequential controllers are developed that have a RoA overlap, and the overlap or reliable switch conditions are difficult to define. This chapter has demonstrated learning when to switch with a dynamic, underactuated biped, with torque controlled actuation and many DoF. For robots of various complexity, learning switch estimators can improve the reliability of switching between visuo-motor behaviours.



\clearpage 

\includepdf[pages=-,pagecommand={},width=\textwidth]{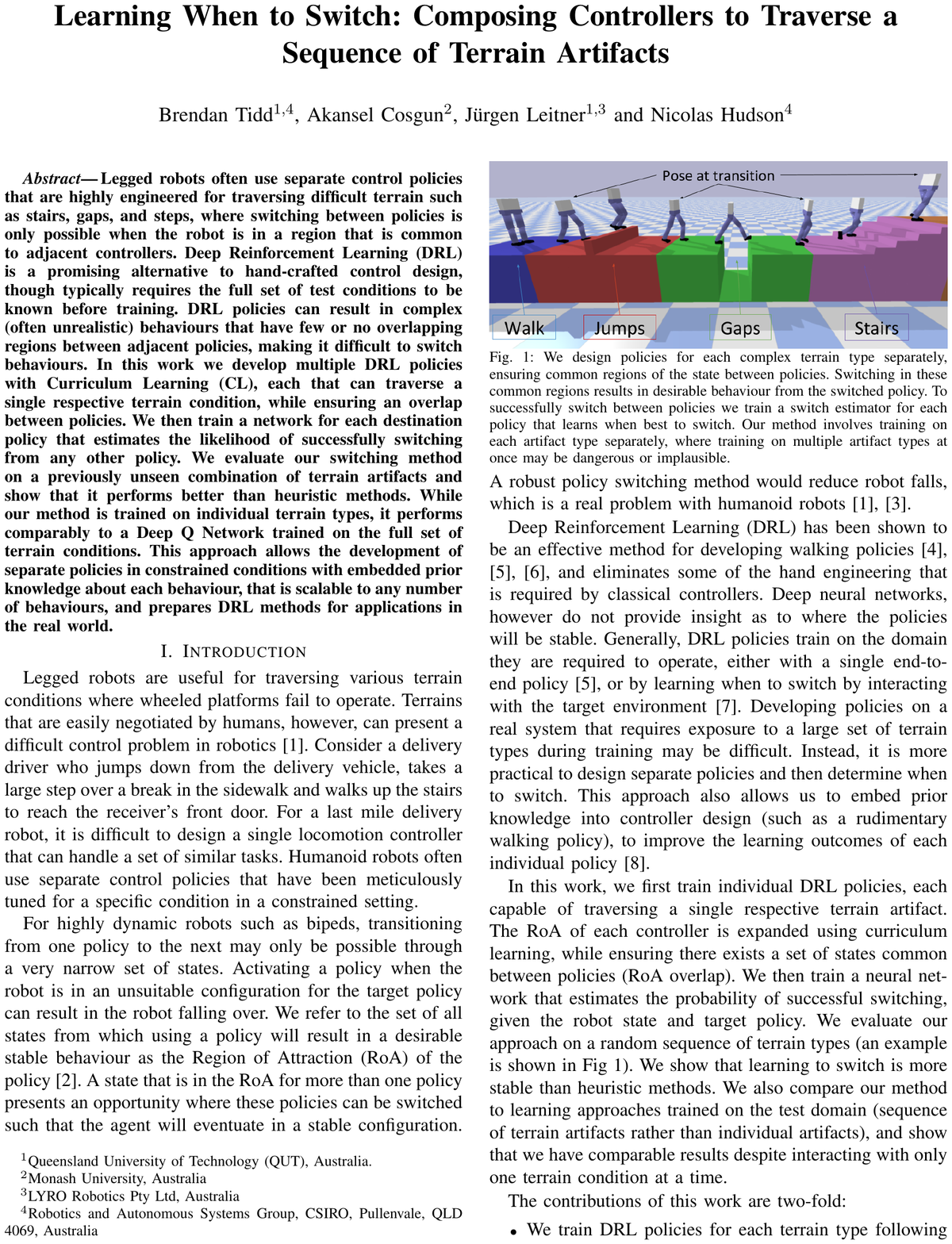}

\chapter[Learning Setup Policies]{Learning Setup Policies: Reliable Transition Between Locomotion Behaviours}

\label{cha:ch5}

\includepdf[pages=-,pagecommand={},width=\textwidth]{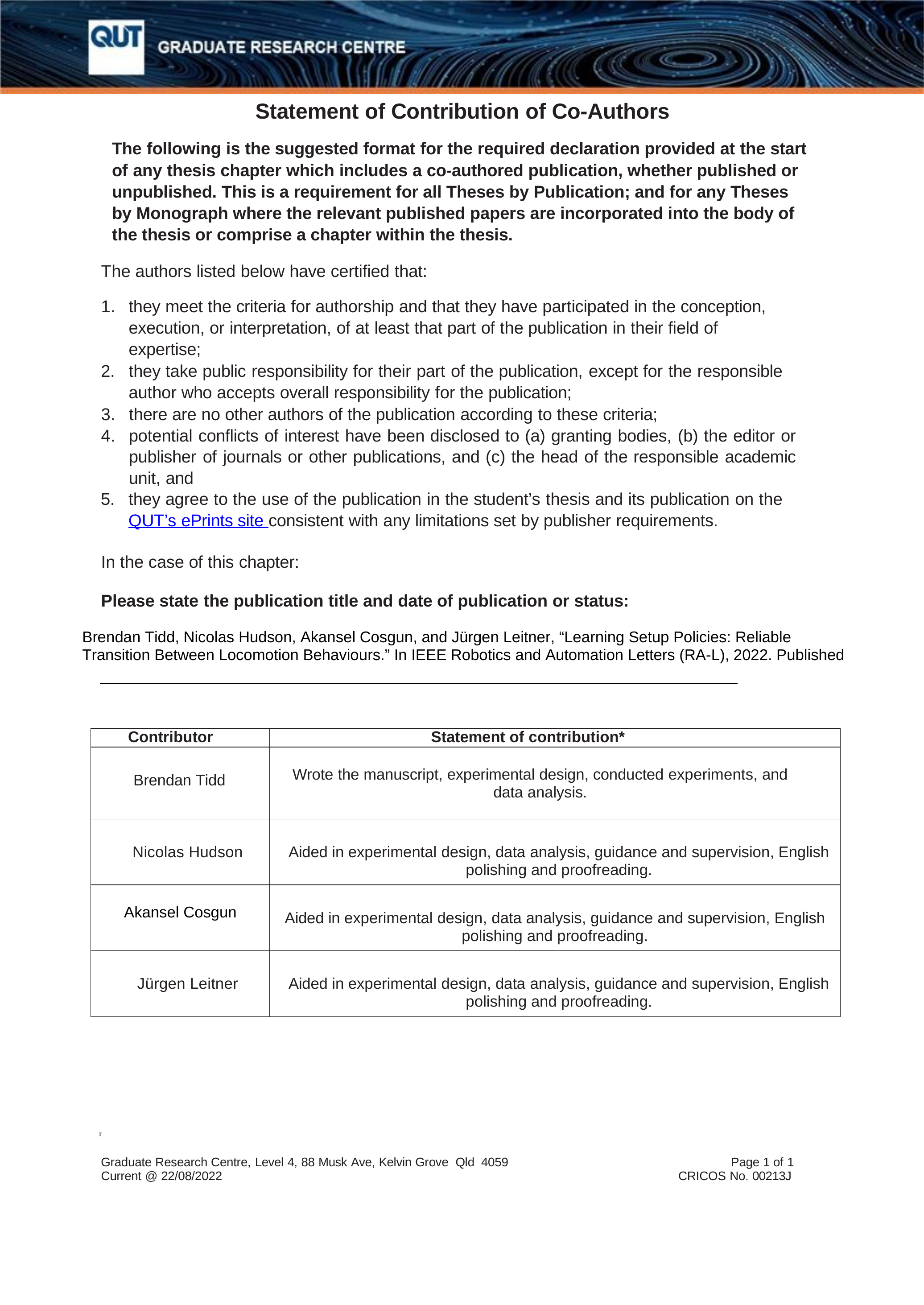}
%
This chapter continues to explore research question 2: \textbf{\textit{how can a safe switch state be determined to facilitate the reliable switching of behaviours,}} and presents the investigation of research question 3: \textbf{\textit{how can an agent prepare for an upcoming behaviour such that safe switching can occur?}} This chapter considers the scenario where a reliable switch state does not exist between behaviours. Setup policies were developed for moving a robot into the region of attraction (RoA) of the behaviour required to traverse the terrain. The presented method was shown to outperform the best existing method for learning transition behaviours [\cite{lee_composing_2019}]. The contributions of this chapter are listed below. 
\begin{itemize}
    
    \item Setup policies were developed that demonstrated an improved success rate for transitioning between difficult behaviours, particularly for controllers without a dependable (RoA) overlap. A switch mechanism was learned in conjunction with the setup policy to improve the reliability of behaviour switching. 
    
    \item A reward signal was designed to train setup policies that guided the robot towards the RoA of a target behaviour. By weighting the target policy value function by the advantage, the reward minimised overestimation bias and resulted in better learning outcomes compared to several alternative reward functions.
\end{itemize}

“Learning Setup Policies: Reliable Transition Between Locomotion Behaviours” was published in IEEE Robotics and Automation Letters (RA-L). 

\newpage

\begin{figure}[htb!]
\centering
\includegraphics[width=0.85\columnwidth]{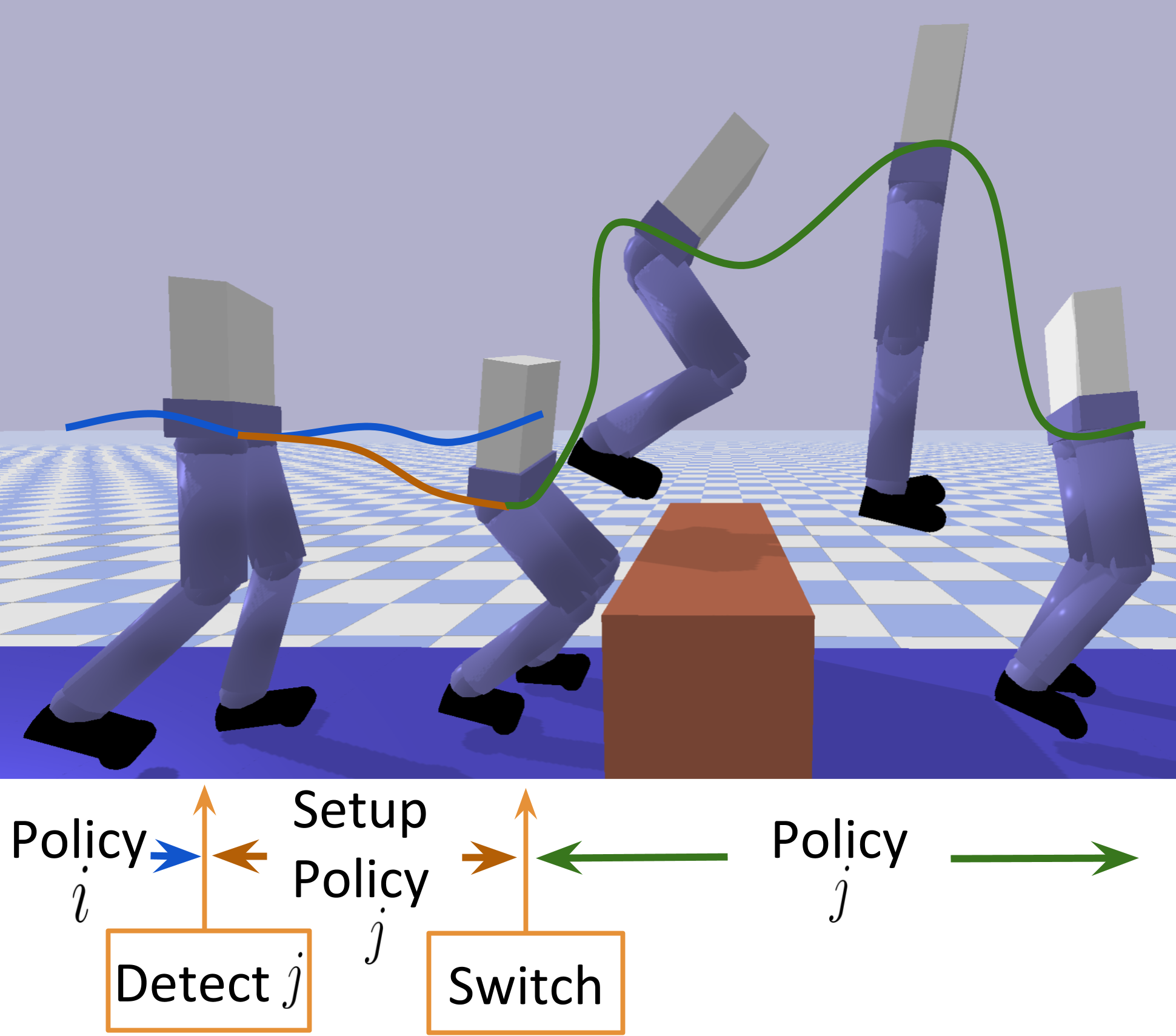}
\hfill
\caption{Setup policies enable the transition from the trajectory of the robot controlled by policy $i$ to the trajectory of the robot controlled by target policy $j$.}
\label{fig:cha4_setup_fig1}
\end{figure}

Chapter \ref{cha:ch4} investigated a method for determining a switch state to improve the reliability of behaviour switching, however, a switch state may not exist between behaviours that are very different from one another. Dynamic behaviours with a narrow region of attraction (RoA) are particularly sensitive to instability from behaviour switching. For example, periodic behaviours, such as walking, and proximity of the robot to an obstacle may prevent the timely activation of the next behaviour. Transitioning between behaviours considering these challenges is investigated in this chapter.

\begin{figure}[tb!]
\centering
\subfloat[]{\includegraphics[width=0.8\columnwidth]{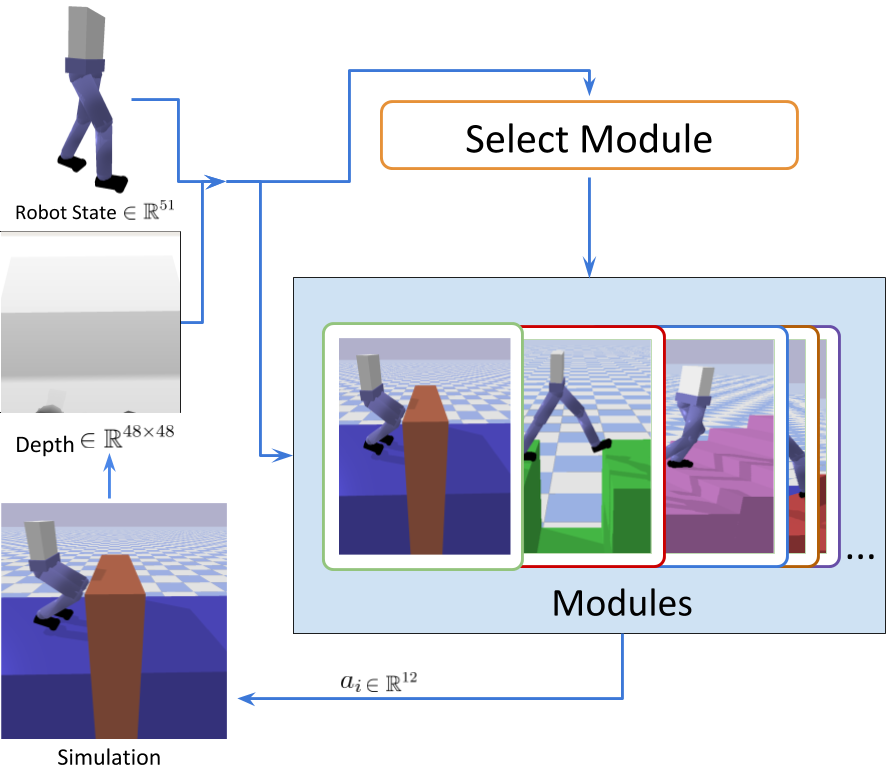}}
\hfill
\subfloat[]{\includegraphics[width=0.85\columnwidth]{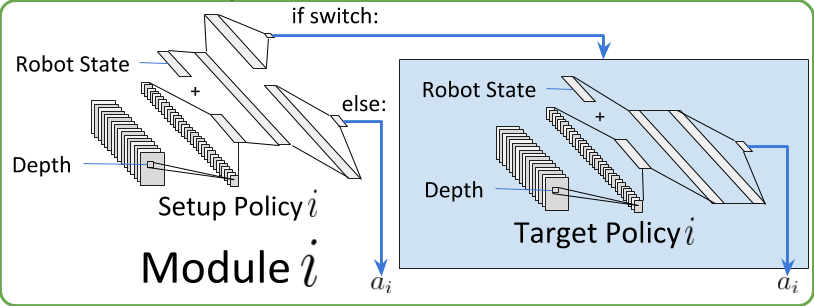}}
\caption{a) A collection of modules was used for traversing a sequence of terrain types, where complex obstacles require specialist policies. b) Each module (target policy and setup policy) was trained on a single terrain type, where setup policies prepare the robot for the target policy.}
\label{fig:cha4_setup_fig2}
\end{figure}


Figure \ref{fig:cha4_setup_fig1} highlights the primary problem investigated in this chapter. The trajectory induced by activating one behaviour does not intersect the trajectory from the behaviour needed to traverse the obstacle. Setup policies were initialised from a simple walking policy to ensure the robot can initially move towards the terrain, and trained to transition from one behaviour to the next while also learning when to switch. Figure \ref{fig:cha4_setup_fig2}a) shows the pipeline from robot state and perception input to joint torque, using any number of modules. Figure \ref{fig:cha4_setup_fig2}b) introduces the separate modules containing both the setup policy (including a switch bit for determining when to switch), and the target policy. Research question 2 is further investigated through the design of a switch mechanism trained simultaneously with the transition policy. Research question 3 was explored by learning how to transition to a target behaviour improving the safety of controller switching, with pronounced benefit in cases where behaviours do not have a reliable intersection.


\begin{figure}[tb!]
\centering
\includegraphics[width=0.85\columnwidth]{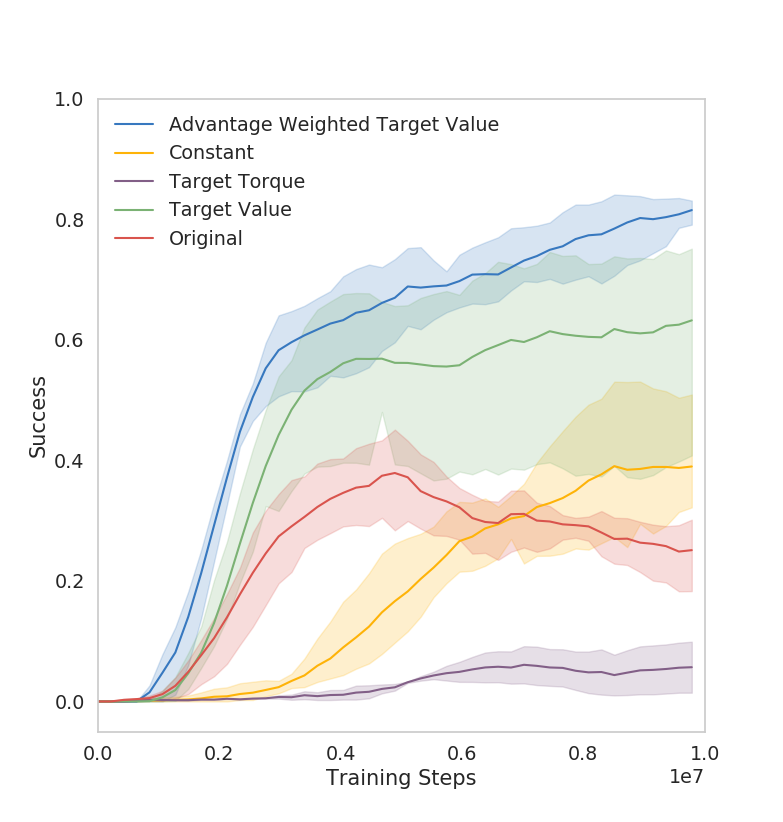}
\hfill
\caption{Several options for reward function were investigated. This figure shows the evolution of the success rate with training (from three random seeds).}
\label{fig:ch4b_results}
\end{figure}

Setup policies were trained to maximise a novel reward function, \textbf{Advantage Weighted Target Value}, that scales the value function output of the target policy by the advantage observed in each step of the environment, results are shown in Figure \ref{fig:ch4b_results}. Temporal difference (TD) error was used to estimate the advantage at each timestep, providing an confidence measurement for reducing the overestimation bias that is common with value functions for states that did not appear in training [\cite{haarnoja_learning_2019}]. Compared with several alternative reward functions, \textbf{Advantage Weighted Target Value} was shown to achieve the highest success rate, with a lower variance than using the target policy value function directly.

\newpage

Results show that setup policies perform more reliably that several alternative methods for transitioning from a walking policy to a complex jump behaviour. Comparative methods include the switch method from Chapter \ref{cha:ch4} and the best method from the literature that applies a proximity prediction network [\cite{lee_composing_2019}]. Table \ref{tab:ch4b_results_compare} shows that setup policies improve the success rate for jumping over a large obstacle. This method improved the success rate for traversing a sequence of terrains compared switching without setup policies (Table \ref{tab:ch4b_multi_results}). Additionally, no polices require retraining as new behaviours are introduced.

\begin{table}[h!]
    \centering
    \begin{tabular}{cccccc}
                                      & Success\% & Distance\% \\
       \hline
        Learn When to Switch          & 0.0           & 44.2  \\
        Proximity Prediction         & 51.3           & 82.4 \\
        Setup Policy (Ours)         & \textbf{82.2} & \textbf{96.1}\\
    \hline
    \end{tabular}
    \caption{Success and average distance covered when switching from a walking policy to a jump policy on a single terrain sample.}
    \label{tab:ch4b_results_compare}
\end{table}


\begin{table}[h!]
    \centering
    \begin{tabular}{cccccc}
                                      & Success \% & Distance \% \\
       \hline
        Without Setup Policies        & 1.9      & 36.3  \\
        With Setup Policies       & \textbf{71.2}      & \textbf{80.2}  \\
    \hline
    \end{tabular}
    \caption{Success rate and average distance travelled of the total terrain length from 1000 episodes of all 5 terrain types (stairs, gaps, jumps, hurdles, stepping stones), randomly shuffled each episode.}
    \label{tab:ch4b_multi_results}
\end{table}

Setup polices improve the reliability of transitioning between sequential behaviours, this method provides the greatest advantage for scenarios where two behaviours do not have dependable RoA overlap. The specific focus of this work is transitioning from a walking policy to a target behaviour, allowing for modularity and scalability with minimal retraining as new behaviours are added. However, robust transition from a practical number of initial behaviours should be considered for this method to be applied more generally. The ideas from this chapter can be applied as a confidence metric for RoA expansion for increasing the robustness of behaviours for safe reinforcement learning, as an expressive reward for behaviour cloning in teacher-student learning, and to blend controllers for combining several composite behaviours. 


\clearpage 

\includepdf[pages=-,pagecommand={},width=\textwidth]{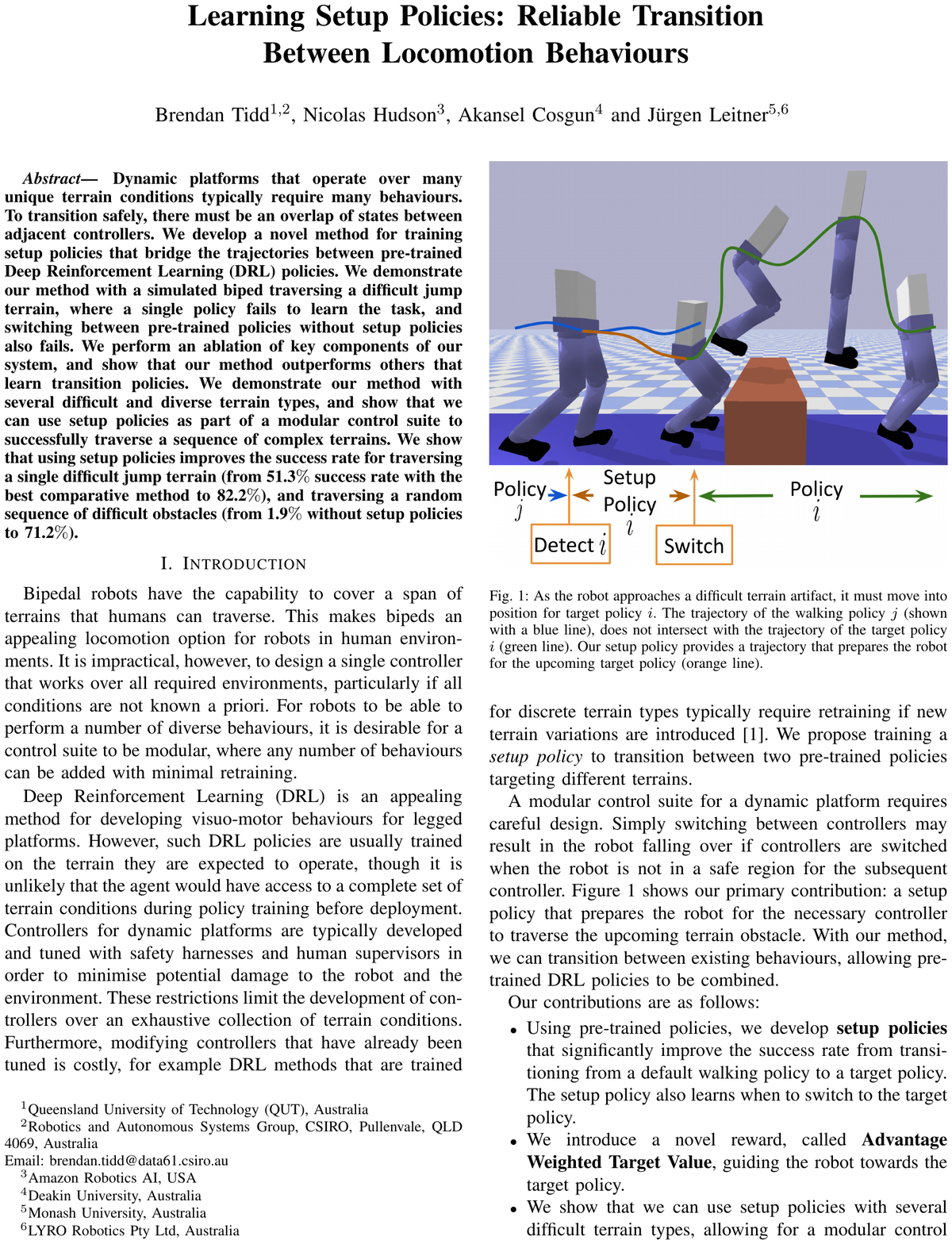}


\chapter[Conclusion]{Conclusion}
\label{cha:conclusion}

This thesis explored the development of complex visuo-motor behaviours for robot locomotion traversing difficult terrain. Real-world challenges prevent the development of single monolithic controllers for diverse terrain conditions that are encountered by mobile robots. The challenges of learning and combining separate controllers was investigated throughout this work.

Deep reinforcement learning (DRL) approaches are an alternative to traditional methods for designing complex visuo-motor policies, however, extensive interaction with the environment is typically required. Furthermore, learned policies behave as a black box where integration with other behaviours is challenging. This thesis investigated methods for improving the sample efficiency of training behaviours, and for integrating these policies to create modular behaviour libraries. This final chapter summarises the progress made while investigating the posited research questions and includes a discussion with avenues for future work.


\section[Summary]{Summary}
\label{cha:summary}

    \textbf{Research Question 1: \textit{How can complex visuo-motor locomotion behaviours be learned efficiently?}}
        
        Learning to traverse complex terrain with a dynamic biped is challenging. Chapter \ref{cha:ch3a} introduced a three-stage curriculum learning approach for traversing a diverse set of terrain types with a biped in a simulation environment. This work showed that a single walking trajectory of body and joint positions was sufficient to improve learning outcomes for several diverse and complex behaviours, reducing the dependence on expensive expert demonstrations. The simple trajectory was provided as guide forces during the \textbf{guide curriculum}, that was active while obstacle difficulty was slowly increased (\textbf{terrain curriculum}). Guide forces were then decreased, and finally a \textbf{perturbation curriculum} further improved the robustness of the policies. Each of these stages contributed to the efficient learning of robust behaviours for traversing curved paths, stairs, stepping stones, hurdles, and gaps. 
        
        
        Planning through narrow passageways is challenging with low resolution costmaps, complex maneuvers are required when traversable gaps are perceived as smaller than the robot footprint. Chapter \ref{cha:ch3b} introduced a behaviour for negotiating narrow gaps with a large tracked platform. A reward was introduced from key placement of waypoints to enable the efficient learning of a gap behaviour. This work was developed in simulation using ground truth perception and applied directly to a robot in a real-world scenario. The resulting success rate was 93\% in simulation, and 73\% on the real platform, where traditional path planning methods that consider the kinematics of the robot were unable to complete the task [\cite{hudson_heterogeneous_2021}]. Factors such as imperfect perception on the real agent, inaccuracies in simulation dynamics, and out of distribution examples seen during deployment were major challenges identified in the sim-to-real transfer.
    
    Both of these contributions demonstrate the efficient learning of visuo-motor policies through curriculum learning and reward design. The methods developed from this investigation resulted in complex behaviours and the traversal of difficult terrains where alternative methods were shown to be inefficient, or unable to complete the tasks. This completes the investigation of research question 1. 
    
    \textbf{Research Question 2: \textit{How can a safe switch state be determined to facilitate the reliable switching of behaviours?}}
        
        Autonomous behaviour selection reduces operator load in stressful multi-agent scenarios. A behaviour selection module was trained in Chapter \ref{cha:ch3b} that learned to switch to a gap behaviour, enabling a tracked robot to pass through a small doorway from a single operator command. The resulting method achieved a success rate of 63\% in simulation and 40\% in real experiments.
        
        Reliable behaviour switching must occur when the robot is in the region of attraction (RoA) of the upcoming behaviour. Chapter \ref{cha:ch4} introduced a switch estimator policy that was trained with supervised learning to predict the likelihood of success for switching to a given policy in the current state. Using policies developed in Chapter \ref{cha:ch3a} for the simulated biped, switch estimator policies improved the success rate for traversing a sequence of difficult terrain to 71.4\%, from 66.8\% using the next best switch method. The effect of state overlap on safe policy switching was also investigated. Initiating diverse policies from a common configuration during training provided sufficient RoA overlap for reliable switching, where policies trained from a random initial configuration performed poorly (success rate of 0.7\%). 
        
        Chapter \ref{cha:ch5} trained a switch mechanism concurrently with a transition policy. A key element was an extended reward that provided the policy with experience after the transition occurred.
        
    Understanding when to switch is essential for composing behaviours. Each of these contributions determined safe states for the reliable switching of behaviours, completing the investigation of research question 2.

    \textbf{Research Question 3: \textit{How can an agent prepare for an upcoming behaviour such that safe switching can occur?}}

        The reliable transition between controllers is a critical issue for behaviour-based mobility, however, controllers that produce different behaviours may not have a dependable RoA overlap. Setup policies developed in Chapter \ref{cha:ch5} improved the reliability of transitioning from a walking behaviour to a difficult jump behaviour, achieving an 82\% success rate (compared to 51.3\% for the next best transition method). Setup policies were trained by scaling the value function of the target policy by the temporal difference (TD) error. This method resulted in an improved success rate for traversing a sequence of difficult obstacles where transitioning between several complex behaviours was required.
        
        
    Behaviours that do not have a clear RoA overlap are unreliable when switching is required. Setup policies were developed to prepare the robot for the upcoming behaviour such that safe switching was possible, completing the investigation of research question 3. 
 
\section[Discussion]{Discussion}
\label{cha:discussion}

Discussion and ideas for future work are presented in this final section.

\begin{enumerate}[label=\textbf{\roman*},leftmargin=*]

\item \textbf{Adaptive curriculum learning}
\\
Curriculum learning was used in Chapter \ref{cha:ch3a} for developing complex visuo-motor policies. While a manually tuned curriculum worked well for the investigated robots and terrain conditions, an autonomous process could allow for quicker learning and better generalisation. Recently, works that have applied curriculum learning to traverse difficult terrain with legged robots have developed adaptive [\cite{xie_allsteps_2020}] or game-inspired curriculum [\cite{rudin_learning_2021}], and teacher-student learning [\cite{miki_learning_2022}]. Applying these ideas with guided curriculum learning (GCL) (Chapter \ref{cha:ch3a}) could improve learning outcomes and robustness for a greater range of challenging behaviours. 
    
\item \textbf{Behaviour granularity}
\\
The behaviours developed throughout this thesis were designed for traversing specific terrain types, however, single policies have been shown to cross a variety of terrain conditions [\cite{heess_emergence_2017}, \cite{miki_learning_2022}]. In Chapter \ref{cha:ch4} it was demonstrated that an end-to-end policy was able to traverse gaps, hurdles, and stairs, however, as the terrains became more difficult, single policies were unable to learn the challenging maneuvers (Chapter \ref{cha:ch5}). While a single policy can be trained to perform a variety of tasks, as task diversity increases, behaviours become difficult to train and simple trajectories are no longer enough to guide learning. Furthermore, as composite behaviours emerge, for example grasping an object while walking, deciding how to separate behaviours becomes an exciting challenge. Behaviour switching introduces a risk, therefore single policies should be trained for many tasks where possible, however, as task complexity grows there will always be a need to transition and switch between behaviours. Exploring when to have a single controller performing multiple functions and how fine-grained separate behaviours should be to perform complex tasks are interesting design decisions as behaviour libraries scale.


\item \textbf{Perception granularity}
\\
This thesis explored the development of visuo-motor policies using a depth image or occupancy map as perception input, however, these modalities limit the potential for learning complex behaviours. For example, passing through narrow gaps is difficult with low resolution occupancy maps (Chapter \ref{cha:ch3b}), robots must learn to make contact with the environment to wiggle through. Furthermore, cost and occupancy maps require post-processing, adding to computational complexity. DRL controllers have the unique ability to utilise expressive perception capabilities, such as images, point cloud, or Surfels (points with shape coding [\cite{pfister_surfels_2000}]), improving fine-grained control in challenging environments. Other advantages include the extraction of semantic information from RGB images, for example to enable legged platforms to walk through grass. Considering more expressive perception sensors for learning complex visuo-motor behaviours would enable more graceful control for traversing difficult terrain.


\item \textbf{Sim-to-real transfer}
\\
Improving the efficiency of learning complex visuo-motor behaviours has not negated the need for training in simulation. Transferring policies developed in simulation directly to real agents remains challenging. The simulation environment used to train policies must closely match the real world, Chapter \ref{cha:ch3b} highlighted several sources of disparity, including perception discrepancy, scenarios not experienced during training, and differences in dynamics properties. Adaptions to training could improve transfer, including adding noise to perception inputs, expanding the simulation samples to minimise out of distribution examples, modelling the real-world acceleration limits of the robot in simulation, and dynamics randomisation (for example friction and robot inertia properties). While these strategies can improve transfer, as robot and environmental complexity increases the simulation properties differ further from the real world. Data collected from real world interactions can be used to train models that refine the simulation and improve transfer. \cite{hwangbo_learning_2019} train an actuator model that captures difficult to model properties such as latency and response of a series elastic actuator. Developing robust models that extract valuable information from real-world data for use in training improves the reliability of sim-to-real transfer and unlocks learning methods for broader practical applications.

\item \textbf{Training with confidence}
\\
It has been demonstrated in this work that RoA overlap was important for switching between behaviours. Methods for switching were investigated with the assumption these controllers were not amenable, however, the ability to increase the RoA of these policies could greatly improve their robustness, and the reliability of behaviour switching. Metrics for estimating where a policy is confident, such as temporal difference (TD) error as explored in Chapter \ref{cha:ch5}, and ensemble learning [\cite{lakshminarayanan_simple_2017}], can be used for RoA expansion, teacher-student learning, and safe reinforcement learning by guiding the robot to meaningful states where the robot is not yet confident [\cite{berkenkamp_safe_2017}]. In addition, these ideas could be applied to controller blending for combining multiple behaviours for performing complex composite tasks, for example object manipulation while walking. 

\item \textbf{Robust Behaviour Transitioning}
\\
In Chapter~\ref{cha:ch5} setup policies were introduced that safely guide a robot from one behaviour to another. To apply this idea more broadly, transition policies should consider a range of initial behaviours during training. As the capabilities of separate behaviours are expanded [\cite{2022-TOG-ASE}], it is not yet clear where these transitions will occur, nor the extent of overlap with other behaviours necessary to encompass the practical use of the robot. Future work in this space will explore generalised transition behaviours that exist at the boundary of highly complex behaviours. Understanding the transition boundaries of high-dimensional dynamic systems to safely switch between adjacent behaviours remains a challenging problem. 

\item \textbf{Terrain detection}
\\
This thesis considered the problem of switching between behaviours, however, in chapter \ref{cha:ch4} and \ref{cha:ch5} a terrain oracle was used for terrain detection. Switching between behaviours and training setup policies relied on ground truth knowledge of the terrain type. A terrain detector would be required when these strategies are deployed on real systems. To continue the theme of this thesis, a terrain detector would need to be adaptable to new terrains as behaviours become available. Possible solutions include offline data storage for retraining a classifier as required, and developing a detector using open set classification techniques [\cite{miller_class_2021}].

\end{enumerate}

\renewcommand{\bibname}{References}
\bibliographystyle{apalike}      
\bibliography{./References}       
\addcontentsline{toc}{chapter}{\bibname}    

\cleardoublepage
\end{document}